\pdfoutput=1
\documentclass[10pt]{article}

\PassOptionsToPackage{sort&compress}{natbib}
\usepackage{thuidm}
\setcitestyle{numbers,super,open={},close={}}
\usepackage[utf8]{inputenc}
\usepackage{CJKutf8}
\usepackage{nicefrac}
\usepackage{multicol}
\usepackage{multirow}
\usepackage{tabularx}
\usepackage{adjustbox}
\usepackage{colortbl}
\usepackage{amsfonts}
\usepackage{eucal}          %
\usepackage{mathtools}
\usepackage{dsfont}
\usepackage{pifont}
\usepackage{wrapfig}
\usepackage[ruled,vlined]{algorithm2e}
\usepackage{etoc}
\usepackage[section]{placeins}

\graphicspath{{figures/}}

\definecolor{Gray}{gray}{0.9}
\definecolor{babypink}{rgb}{0.96, 0.76, 0.76}
\definecolor{champagne}{rgb}{0.97, 0.91, 0.81}

\SetAlCapFnt{\sffamily\bfseries\small\color{thupurpledark}}
\SetAlCapNameFnt{\small}

\makeatletter
\newcommand{\thuidm@renewthm}[3]{%
  \expandafter\let\csname #1\endcsname\relax
  \expandafter\let\csname end#1\endcsname\relax
  \expandafter\let\csname c@#1\endcsname\relax
  \theoremstyle{#3}\newtheorem{#1}{#2}}
\thuidm@renewthm{theorem}{Theorem}{thuidm}
\thuidm@renewthm{lemma}{Lemma}{thuidm}
\thuidm@renewthm{proposition}{Proposition}{thuidm}
\thuidm@renewthm{corollary}{Corollary}{thuidm}
\thuidm@renewthm{definition}{Definition}{thuidmplain}
\thuidm@renewthm{remark}{Remark}{thuidmplain}
\thuidm@renewthm{assumption}{Assumption}{thuidmplain}
\theoremstyle{thuidmplain}

\newcommand{\thuidm@venue}{Accepted to \textit{Nature Communications}}
\newcommand{\thuidm@statusdate}{\thuidm@metaline{Status:}{\thuidm@venue}%
  \thuidm@metaline{Code:}{\href{https://github.com/thu-rllab/MPTS}{github.com/thu-rllab/MPTS}}%
  \thuidm@metaline{Date:}}
\patchcmd{\maketitle}{\thuidm@metaline{Date:}}{\thuidm@statusdate}{}{\PackageWarning{main}{could not add status line}}
\patchcmd{\maketitle}{\thuidm@metaline{Date:}}{\thuidm@statusdate}{}{\PackageWarning{main}{could not add status line}}
\makeatother

\title{Model Predictive Task Sampling for\\Efficient and Robust Adaptation}

\author{%
  Qi Cheems Wang (\begin{CJK}{UTF8}{gbsn}王琦\end{CJK})\aff{1,*},
  Zehao Xiao (\begin{CJK}{UTF8}{gbsn}肖泽昊\end{CJK})\aff{2,*},
  Yixiu Mao (\begin{CJK}{UTF8}{gbsn}毛逸休\end{CJK})\aff{1,*},
  Yun Qu (\begin{CJK}{UTF8}{gbsn}曲云\end{CJK})\aff{1,*},\\
  Jiayi Shen (\begin{CJK}{UTF8}{gbsn}沈佳怡\end{CJK})\aff{2},
  Yiqin Lv (\begin{CJK}{UTF8}{gbsn}吕怡琴\end{CJK})\aff{1},
  Xiangyang Ji (\begin{CJK}{UTF8}{gbsn}季向阳\end{CJK})\aff{1,$\dagger$}%
}

\affiliation{%
  \aff{1}Department of Automation, Tsinghua University,\quad
  \aff{2}Informatics Institute, University of Amsterdam\\
  \aff{*}Equal contribution,\quad \aff{$\dagger$}Corresponding author%
}

\datestamp{October 2026}
\correspondence{Xiangyang Ji at \href{mailto:xyji@tsinghua.edu.cn}{xyji@tsinghua.edu.cn}}
\labwebsite[www.thuidm.com]{https://www.thuidm.com/}

\hypersetup{
  pdftitle={Model Predictive Task Sampling for Efficient and Robust Adaptation},
  pdfauthor={Qi Cheems Wang, Zehao Xiao, Yixiu Mao, Yun Qu, Jiayi Shen, Yiqin Lv, Xiangyang Ji},
  pdfsubject={Accepted to Nature Communications}}

\begin{document}

\begin{abstract}
Foundation models and generalist policies rely on adaptation learning, including meta-training, domain randomization, and supervised finetuning.
Challenging task prioritized sampling is essential for adaptation robustness under distribution shifts. 
However, ranking task difficulty requires exhaustive evaluation, which is prohibitively expensive whether through agent-environment interactions in meta reinforcement learning and domain randomization, or repeated forward passes in large model finetuning. 
We present Model Predictive Task Sampling (MPTS), a framework that bridges task space and adaptation risk distributions for active task selection. 
MPTS employs a lightweight generative model to predict task-specific adaptation risk, amortizing costly evaluation while provably ranking task difficulties. 
It can seamlessly integrate into zero-shot, few-shot, and supervised finetuning settings. 
We conduct extensive experiments in pattern recognition using foundation models and sequential decision-making. 
Empirically, MPTS significantly enhances adaptation robustness for tail risk or out-of-distribution tasks compared to superior methods, and improves learning efficiency over subset selection methods such as $\text{CVaR}_{\alpha}$.
\end{abstract}

\maketitle

\etocdepthtag.toc{mainpart}

\section{Introduction}

Generalization across diverse scenarios remains a central challenge in artificial general intelligence.
The rise of generative AI offers a promising solution, driving the development of foundation models \citep{radford2021learning,achiam2023gpt,kirillov2023segment}. 
Unlike traditional task-specific models, which might fail in new tasks, foundation models enable fast deployment across diverse scenarios without learning from scratch. 
Their rapid problem-solving stems from widely adopted adaptation learning paradigms, including pretraining, meta-learning, and supervised finetuning (SFT).

\begin{figure}[!htbp]
\begin{center}
\centerline{\includegraphics[width=0.98\textwidth]{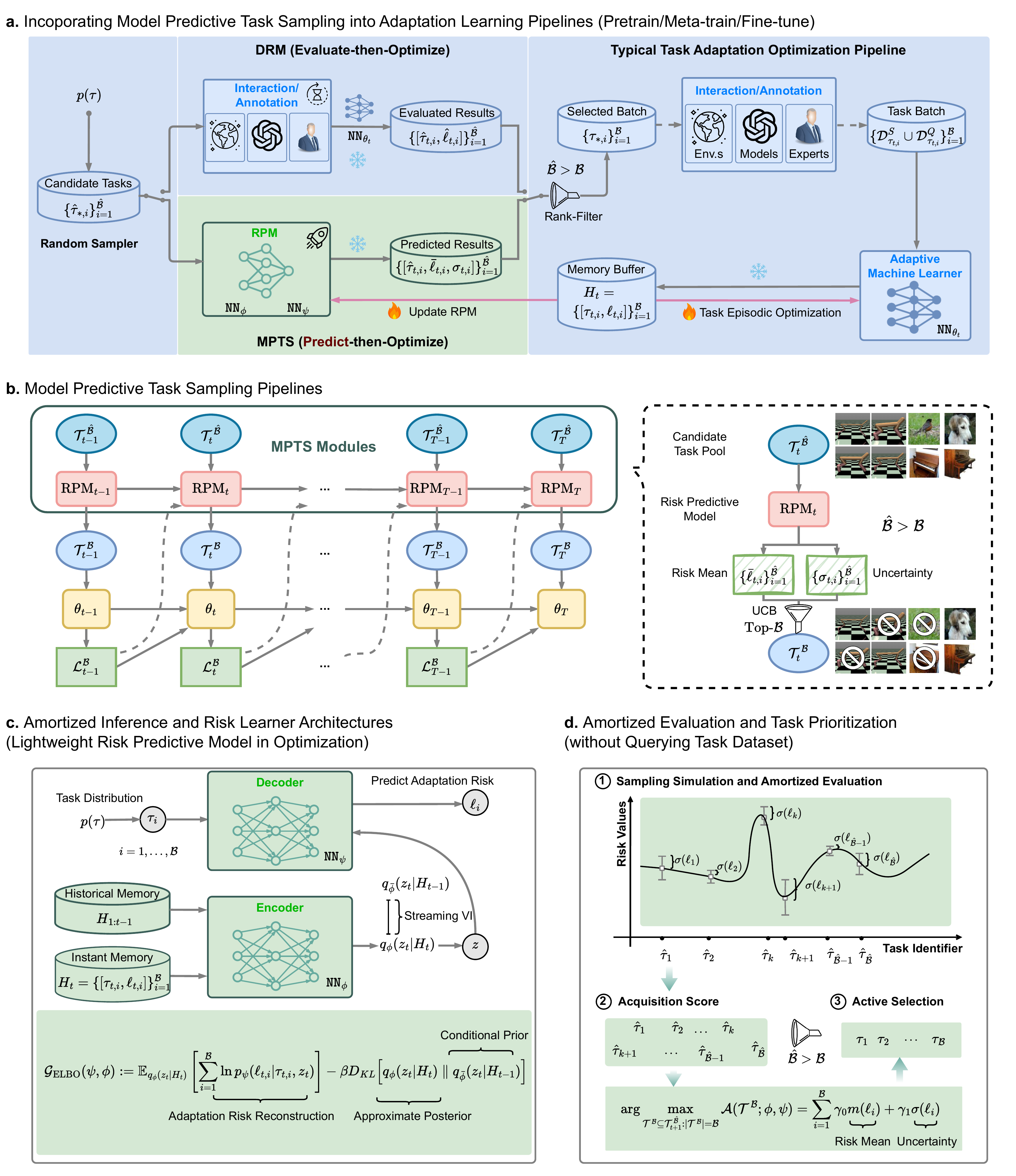}}
\caption{\textbf{Framework of MPTS in Adaptation Learning.}
\textbf{a.}
The left shows standard random sampling for generating candidate tasks.
The middle in blue denotes costly evaluation of $\hat{\mathcal{B}}$ tasks (e.g., agent–environment interaction or foundation model forward pass) in DRM to select Top-$\mathcal{B}$ worst ones.
The middle in green depicts MPTS, which predicts task difficulty via a lightweight generative model, avoiding expensive evaluation.
The right illustrates the standard optimization pipeline in Meta-Learning, DR, or SFT (Snow: frozen models; Fire: updated models).
\textbf{b.}
MPTS samples candidate tasks, ranks their difficulty via the risk predictive model's predictions for subset selection, and updates the learner, approximating $\text{CVaR}_{\alpha}$'s Monte Carlo optimization in a predictive manner. 
The gathered risk signals further update RPMs online.
\textbf{c.}
The RPM utilizes the risk history $H_{1:t}$ to train under a streaming VI framework.
\textbf{d.}
The RPM simulates adaptation outcomes $p(\ell\vert\bm\tau,H_{1:t};\bm\theta_{t})$ for $\hat{\mathcal{B}}$ candidate identifiers, computes acquisition scores, and selects the Top-$\mathcal{B}$ identifiers for the $(t+1)$-th iteration.
}
\label{fig_overall_mpts}
\end{center}
\end{figure}

\begin{figure}[!htbp]
\begin{center}
\centerline{\includegraphics[width=1.0\textwidth]{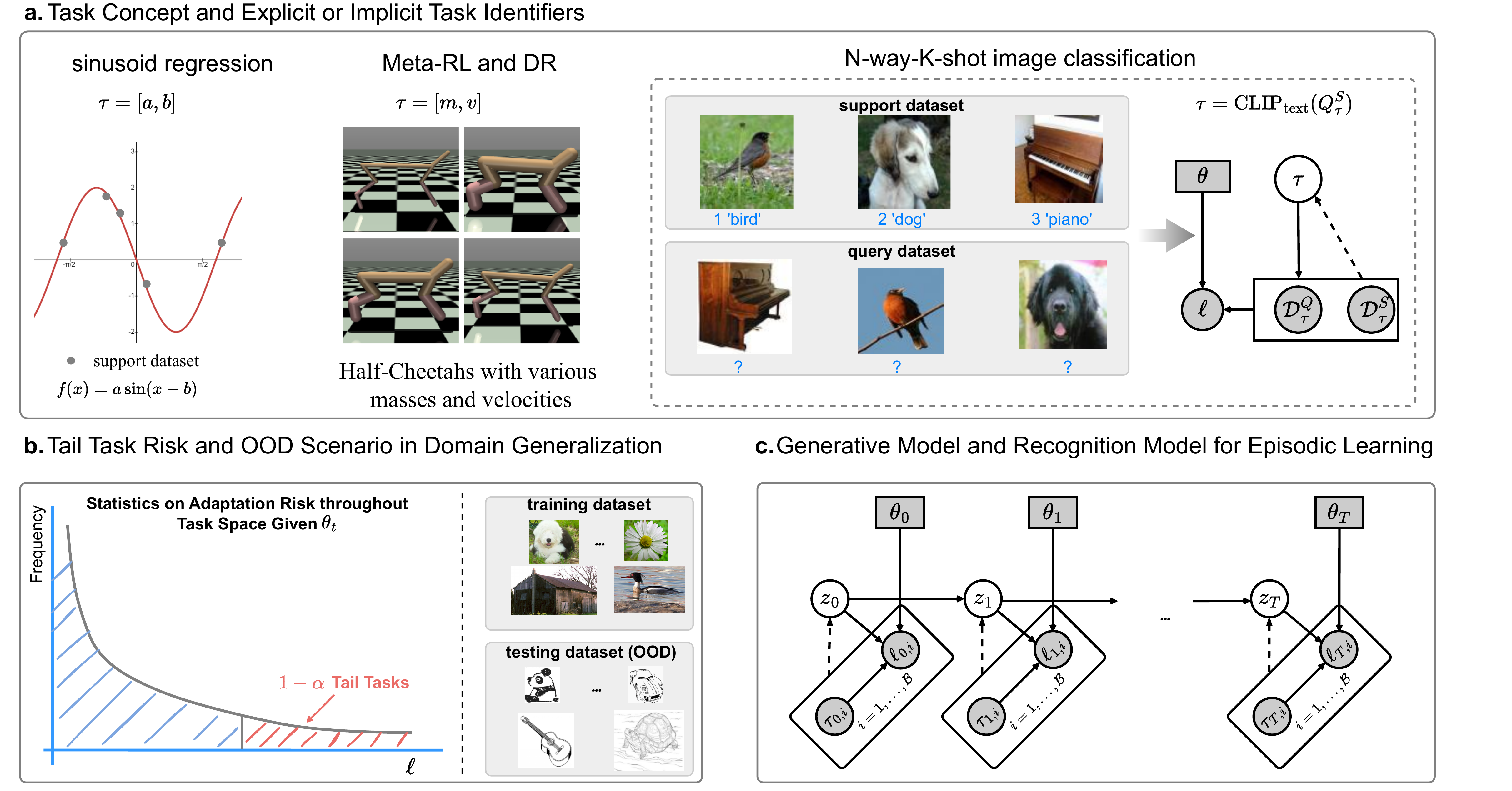}}
\caption{\textbf{Fundamental Concepts: Task Identifiers, Episodic Learning and Probabilistic Graphical Models.}
\textbf{a.} The task distribution is uniform and defined over meaningful identifiers $\bm\tau$.
For example, the amplitude and the phase $[a,b]$ specifies a sinusoid curve to complete with \texttt{K-shot} observed data points. 
Robots like Half-Cheetahs are trained to accomplish different locomotion tasks with varying masses and velocities.
Some multimodal pattern recognition tasks' identifiers are implicit but can be described from a reference model, e.g., text encoders in CLIP \citep{radford2021learning}.
\textbf{b.} The tail task generalization corresponds to $\text{CVaR}_{\alpha}$, i.e., the integral of tail task risk values in red. 
In OOD generalization, this work prompt-tunes CLIP on ImageNet \citep{russakovsky2015imagenet} to test on ImageNet-S \citep{wang2019learning}. 
\textbf{c.} Here, the generative model includes grey units as observed variables and white ones as unobservable.
The solid directed lines describe the generative model \citep{tomczak2024deep}.
We use the dash-directed lines to indicate the recognition model and approximate inference within autoencoding variational Bayes \citep{kingma2013auto}.
}
\label{fig_basic_setup}
\end{center}
\end{figure}

These paradigms train machine learners over a task distribution, consolidating past experience into problem-solving priors to handle unseen but related tasks in zero-shot or few-shot settings \citep{radford2021learning,alayrac2022flamingo}. 
Each iteration samples a task batch, e.g., from a uniform distribution, and then executes the learning-to-adapt step (see Fig. \ref{fig_overall_mpts}). 
Large language models (LLMs), for instance, treat episodic corpus datasets as tasks and perform in-context learning for adaptation \citep{brown2020language}.
Similarly, in obtaining generalist robotic policies, decision-making environments, such as Markov decision processes (MDPs), are randomized for robots to perform policy optimization. 
These task distributions are typically determined by identifiers; e.g., in Fig. \ref{fig_basic_setup}a, varying physics parameters configure different MDPs as tasks for domain randomization (DR) \citep{akkaya2019solving} and meta reinforcement learning (Meta-RL) \citep{duan2016rl}.

Distribution shifts \citep{koh2021wilds,ajay2022distributionally} are prevalent in real-world scenarios, making task adaptation robustness at test time increasingly critical \citep{sun2024evaluating,zhou2024larger}. 
In this context, task sampling strategies play a pivotal role, yet uniform sampling often causes catastrophic failures in risk-sensitive scenarios due to the under-sampling of critical tasks.
Two real-world applications highlight this case:
(i) Tail tasks. In developing autonomous-driving systems, traffic accidents are rare in training datasets but disproportionately important for testing robustness \citep{rempe2022generating}.
(ii) OOD tasks. Robots trained in controlled environments struggle in unstructured real-world settings, e.g., leading to errors in navigation and object manipulation.
To improve robustness, challenging task prioritized sampling \citep{rempe2022generating,dubey2024llama,wang2024simple} has gained traction, where assessing task difficulty is central to robust optimization. 
These methods \citep{sorscher2022beyond,evans2023bad,wang2024simple,dubey2024llama,greenberg2024train,evans2024data} evaluate, rank, and prioritize difficult tasks for iterative optimization (see Fig. \ref{fig_overall_mpts}a).
However, precisely evaluating tasks via losses, human annotations, or gradients incurs high computational costs. 
For instance, in LLM alignment, task evaluation through SFT requires extensive forward passes, while preference optimization consumes millions of expert annotations \citep{ouyang2022training}. 
Similarly, in DR and Meta-RL, agents must interact with numerous MDPs to collect post-adaptation episodes and compute returns.
These challenges uncover the urgent need for more efficient learning strategies when enhancing robust adaptation, particularly when deploying foundation models or when environment interactions are costly.

\subsection{Optimization histories to steer efficient task selection}

Motivated by the above pressing demands, we dive into robust active task sampling, a paradigm that has the potential to eliminate unnecessary costs associated with task construction, intensive annotations, or computational overhead during the evaluation of a machine learner's adaptation to specific tasks. 
In scenarios involving zero-shot learning, few-shot learning, or SFT, we aim to develop a task sampling strategy that requires fewer sample and computation resources but retains more deployment benefits such as adaptation robustness in pattern recognition with foundation models and risk-averse sequential decision-making.

Note that our brain is energy-efficient and simulates the outcome of decision-making in unencountered scenarios from accumulated experience, without actual trials. 
This capability arises from mechanisms like implicit information gating and active task selection \citep{wang2018prefrontal,zheng2024rapid,friedman2022role}. Inspired by this, we propose a model-based optimization approach for adaptive learning, dynamically adjusting task sampling strategies using predicted outcomes as feedback.
This work explores the design of risk predictive models (RPMs) for robust task sampling based on two key insights:
(i) Adaptation risk is probably predictable in episodic learning, providing a basis for task difficulty ranking and selection;
(ii) Generative modeling of adaptation risk captures risk distributions throughout the task space with quantified uncertainty, aligning optimization with robustness principles.

To this end, we introduce Model Predictive Task Sampling (MPTS), a framework for risk-aware task selection. 
As shown in Fig. \ref{fig_overall_mpts}a-b, MPTS leverages historical risk to train a lightweight RPM, which forecasts adaptation risks across the task space to guide the task sampler and optimize the adaptive machine learner.
This way amortizes expensive task evaluation for ranking their difficulty to select the worst subset (see the comparison in the middle block of Fig. \ref{fig_overall_mpts}a). 
The RPM in Fig. \ref{fig_overall_mpts}c adopts a variational autoencoder (VAE) \citep{kingma2013auto} structure, generating adaptation risk estimates via posterior inference \citep{stephan2017stochastic}. 
Finally, the acquisition function in Fig. \ref{fig_overall_mpts}d integrates worst-case performance and predictive uncertainty into the rule of subset selection.

MPTS also draws inspiration from active inference \citep{friston2016active}, which operates through a loop of perception, action, and learning to minimize uncertainty about the planning environment. 
Here, subset selection from the task batch can be viewed as online planning to derive a robust machine learner.
Technically, MPTS specifies or infers identifiers from the task distribution (see examples in Fig. \ref{fig_basic_setup}a) to establish mappings between identifiers and adaptation risk. 
It employs streaming variational inference (VI) \citep{broderick2013streaming,nguyen2017variational} for the RPM's training. 
Furthermore, by simulating adaptation outcomes in a larger identifier pool for subset selection, MPTS balances exploration (uncertainty minimization) and exploitation (worst-case robustness) across the task space.
MPTS is also theoretically grounded, where the optimization proceeds in the direction of robustness enhancement while assuring convergence.
Importantly, this work examines the predictability of online task difficulty with generative models.  
In primary, our proposed MPTS enjoys several benefits in practice:
(1) The optimization pipeline of MPTS can advance the machine learner's adaptation robustness under severe task distribution shifts, such as tail risk or OOD task scenarios;
(2) Constructing the lightweight RPM to amortize expensive task evaluation, MPTS can diminish computational overhead, avoid unnecessary annotations, and promote efficient exploration in the task space;
(3) Learning from risk histories, MPTS serves as a plug-and-play module to rank the task difficulties in optimization and allows seamlessly integration into robust zero-shot or few-shot learning and SFT.

This work evaluates MPTS across few-shot regression, image classification with foundation models, Meta-RL, robotic DR, and prompt-tuning foundation models. 
Empirical results demonstrate MPTS's robust adaptation performance across diverse scenarios. 
Compared to SoTA robust adaptation methods, MPTS significantly reduces computational overhead, memory usage, and environment interactions while, in some cases, accelerating learning.

\subsection{Adaptation risk and robustness concept}

We represent a task sample by $\tau\sim p(\tau)$, with $\mathcal{T}$ denoting the task space.
Each task $\tau$ within the distribution is specified by an identifier \citep{kaddour2020probabilistic}, a real-valued vector $\bm\tau$, as illustrated in Fig. \ref{fig_basic_setup}a.
The task-specific risk function $\ell:\mathcal{D}_{\tau}^{S}\cup \mathcal{D}_{\tau}^{Q}\times\Theta\mapsto\mathbb{R}$ evaluates the adaptation performance of a machine learner $\bm\theta$ on $\tau$.
For example, in regression, the support dataset $\mathcal{D}_{\tau}^{S}=\{[\bm x_i,\bm y_i]\}_{i=1}^{K}$ enables rapid adaptation to obtain the model $p_{\bm\theta}(\bm y\vert\mathcal{D}_{\tau}^{S},\bm x)$; while the query dataset $\mathcal{D}_{\tau}^{Q}=\{[\bm x_i,\bm y_i]\}_{i=K+1}^{K+N}$ is used for post-adaptation evaluation as risk $\ell=-\frac{1}{N}\sum_{i=K+1}^{K+N}\ln p_{\bm\theta}(\bm y_i\vert\mathcal{D}_{\tau}^{S},\bm x_i)$.

If $|\mathcal{D}_{\tau}^{S}|=\emptyset$, $\ell$ measures zero-shot adaptation; otherwise, it reflects few-shot adaptation risk.
In SFT, each sample $(\bm x,\bm y)\in\mathcal{D}_{\text{SFT}}$ is treated as a task.
The episodic task batch history is defined as $\hat{H}_{t}=\{\bm\theta_{t},\left\{\left(\bm\tau_{t,i},\mathcal{D}_{\tau_{t,i}},\ell_
{t,i}\right)\right\}_{i=1}^{\mathcal{B}}\}$, where $\mathcal{B}$ is the task batch size and
$\bm\theta_{t}$ represents the machine learner's parameter in $t$-th iteration.
The tuple set $\left\{\left(\bm\tau_{t,i},\mathcal{D}_{\tau_{t,i}},\ell_{t,i}\right)\right\}_{i=1}^{\mathcal{B}}$ includes the sampled task identifier batch $\{\bm\tau_{t,i}\}_{i=1}^{\mathcal{B}}$, the support and query dataset $\{\mathcal{D}_{\tau_{t,i}}:=\mathcal{D}_{\tau_{t,i}}^{S}\cup \mathcal{D}_{\tau_{t,i}}^{Q}\}_{i=1}^{\mathcal{B}}$, and the evaluated adaptation risk $\{\ell_{t,i}\}_{i=1}^{\mathcal{B}}$.
For simplicity, the risk history is expressed as $H_{t}=\{[\bm\tau_{t,i},\ell_{t,i}]\}_{i=1}^{\mathcal{B}}$, which depends on $\bm\theta_t$.

The learning setup optimizes the machine learner within $p(\tau)$.
Our analysis is interested in the adaptation risk in the task space as illustrated in Fig. \ref{fig_overall_mpts}d.
Such a perspective emphasizes the interplay between the task identifier $\bm\tau$, the task-specific dataset $\mathcal{D}_{\tau}^{S}\cup\mathcal{D}_{\tau}^{Q}$ and the adaptation risk function $\ell$ conditioned on $\bm\theta$.
Regarding adaptation performance, we mainly examine zero-shot learning, few-shot learning, and SFT scenarios.

In terms of zero-shot adaptation, we evaluate $\ell$ on the query dataset $\mathcal{D}_{\tau}^{Q}$ conditioned on the machine learner $\bm\theta$ during training, i.e., $\ell(\mathcal{D}_{\tau}^{Q};\bm\theta)$.
With robotic DR \citep{akkaya2019solving} as an example, $\ell(\mathcal{D}_{\tau}^{Q};\bm\theta)$ denotes the negative return of trajectories collected under the policy $\bm\theta$ in MDP $\tau$.
This setup is without support information.
In terms of few-shot adaptation, the form of $\ell$ is specific to meta-learning methods.
For instance, MAML \citep{finn2017model} implements a bi-level optimization framework.
In this case, $\ell(\mathcal{D}_{\tau}^{Q},\mathcal{D}_{\tau}^{S};\bm\theta)$ is written as $\ell(\mathcal{D}_{\tau}^{Q};\bm\theta_{\text{meta}}-\eta\nabla_{\bm\theta}\ell(\mathcal{D}_{\tau}^{S};\bm\theta))$, where $\bm\theta_{\text{meta}}$ denotes the meta initialization, and the inside-bracket term corresponds to finetuning $\bm\theta_{\text{meta}}$ tailored to $\tau$ with $\eta$ the learning rate.
In terms of supervised finetuning, the machine learner adapts to a specific downstream task using a labeled dataset. 
Formally, given a SFT dataset $\mathcal{D}_{\text{SFT}}$ and a test dataset $\mathcal{D}^{Q}$, the objective is 
$\ell(\mathcal{D}^{Q};\bm\theta(\mathcal{D}_{\text{SFT}},\bm\theta_{\text{init}}))$, 
where $\bm\theta_{\text{init}}$ is the pretrained initialization and $\bm\theta$ denotes the finetuned parameter after adaptation. 
Unlike meta-learning, which optimizes over task distributions, SFT relies solely on the labeled samples from a specific task.

\begin{definition}[Conditional Value-at-Risk, CVaR\citep{rockafellar2000optimization}]\label{def_cvar}
Given the machine learner parameter $\bm\theta$, we denote the task specific random variable by $\ell_i:=\ell(\mathcal{D}_{\tau_i}^{Q},\mathcal{D}_{\tau_i}^{S};\bm\theta)$.
Throughout the task space $\mathcal{T}$, let the cumulative risk distribution and the quantile of risk values respectively be $F(\ell)$ and $\ell^{\alpha}=\min_{\ell}\{\ell\vert F(\ell)\geq\alpha\}$.
Then the CVaR at $\alpha$-robustness level can be estimated as:
\begin{equation}
    \begin{split}
    \text{CVaR}_{\alpha}[\ell(\mathcal{T};\bm\theta)]=
    \int \ell dF^{\alpha}(\ell;\bm\theta),
    \end{split}
\end{equation}
where we define the normalized cumulative distribution of task risk values by:
\begin{equation}
    \begin{split}
        F^{\alpha}(\ell;\bm\theta)=
        \begin{cases}
        0, & l<\ell^{\alpha}\\
        \frac{F(\ell;\bm\theta)-\alpha}{1-\alpha}, & l\geq\ell^{\alpha}.
        \end{cases}
    \end{split}
\end{equation}
And this induce the tail risk task density function denoted by $p_{\alpha}(\tau;\bm\theta)$.
\end{definition}

As illustrated in Fig. \ref{fig_basic_setup}b, the robustness metric $\text{CVaR}_{\alpha}$ in Definition \ref{def_cvar} is commonly used for measuring the expected risk in the worse scenarios, i.e., $1-\alpha$ proportional tailed cases, with $\alpha\in(0,1)$ a specific confidence level.
The induced normalized distribution can be interpreted as a form of distributional shift. 
Beyond this notion, other robustness concepts exist, such as resilience to label noise \citep{yao2021meta,he2024robust}, with representative approaches including ExcessMTL \citep{he2024robust} that explicitly address such scenarios.

\section{Results}

This section reports primary findings in robust adaptation and analyzes the effect of MPTS. 
Prior to elaborating on the experimental setups, we outline the \textit{predict-then-optimize} workflow underpinning MPTS.
In other words, we propose a plausible solution to robust adaptation: a computation- and annotation-efficient framework, as formalized in Definition \ref{definition_mpts}.

\begin{definition}[Model Predictive Task Sampling]\label{definition_mpts}
To amortize the expensive online task evaluation, MPTS reuses the risk history $H_{1:t} = \{\{(\bm\tau_{i,j}, \ell_{i,j})\}_{j=1}^{\mathcal{B}}\}_{i=1}^t$ to construct the risk predictive model $p_{\bm\psi}(\ell\vert\bm\tau_{i},H_{1:t};\bm\theta_{t})$ that guides the active selection of the task subset $\mathcal{T}^{\mathcal{B}}$ from a larger task pool $\hat{\mathcal{B}}$ for $(t+1)$-th optimization step:
$$\mathcal{T}_{t+1}^{\mathcal{B}}=\arg\max_{\mathcal{T}^{\mathcal{B}}\subseteq\mathcal{T}^{\hat{\mathcal{B}}}_{t+1}:|\mathcal{T}^{\mathcal{B}}|=\mathcal{B}}\mathcal{A}(\mathcal{T}^{\mathcal{B}},\bar{\mathcal{L}}^{\mathcal{B}}),$$
where $\bar{\mathcal{L}}^{\mathcal{B}}=\{\bar{\ell}_{t+1,i}\}_{i=1}^{\hat{\mathcal{B}}}$ denotes the predicted risk quantity of a larger task pool to roughly score task difficulties without exact evaluation at $(t+1)$-th iteration.
And $\mathcal{A}(\cdot)$ specifies certain subset acquisition rules, e.g., prioritizing tasks with worse adaptation losses for the purpose of robust optimization. 
\end{definition}

\subsection{Risk prediction with theoretical guarantee and MPTS workflows}

First, we characterize the optimization pipeline for a typical family of robust adaptation methods, i.e., the Monte Carlo estimate for $\text{CVaR}_{\alpha}$ minimization \citep{rockafellar2000optimization}:
\begin{equation}
    \begin{split}
        \cdots\xlongrightarrow{\text{update}}\bm\theta_{t-1}\xlongrightarrow{\text{evaluate}}\{[\hat{\bm\tau}_{t-1,i},\hat{\ell}_{t-1,i}]\}_{i=1}^{\hat{\mathcal{B}}}\xlongrightarrow{\text{Top-$\mathcal{B}$}}H_{t-1}:=\{[\bm\tau_{t-1,i},\ell_{t-1,i}]\}_{i=1}^{\mathcal{B}}\xlongrightarrow{\text{update}}\bm\theta_{t}\xlongrightarrow{\text{evaluate}}\{[\hat{\bm\tau}_{t,i},\hat{\ell}_{t,i}]\}_{i=1}^{\hat{\mathcal{B}}}\xlongrightarrow{\text{Top-$\mathcal{B}$}}\cdots,
    \end{split}
\end{equation}
which picks up the tail tasks to optimize in each iteration.
Existing works to prioritize challenging tasks over iterations \citep{rempe2022generating,dubey2024llama,wang2024simple,evans2024data} take the above steps yet suffer from: (i) learning efficiency issues, such as the need for extensive evaluation of the machine learner across numerous tasks for subset selection, and (ii) restricted batch sizes for evaluating or exploring the task space due to sample or memory constraints.
Notably, nearly all of these approaches fail to leverage the optimization outcomes $H_{1:t}$.

Let us predict what to optimize from the cumulated risk episodes.
MPTS differs from prior works and reuses $H_{1:t}$ to train the RPM.
Coupling the identifier $\bm\tau$ and adaptation risk $\ell(\mathcal{D}_{\tau}^{Q},\mathcal{D}_{\tau}^{S};\bm\theta)$ forms a streaming database to online learn.
In Section \ref{method_sec}, Theorem \ref{theorem_prob_score} will provide a provable basis for ranking tasks from predicted outcomes, suggesting stable ranking relation of task difficulties under perturbations in $\bm\theta$, e.g., a gradient update with a small learning rate. 
Thus, the candidate tasks $\mathcal{T}_{t+1}^{\hat{\mathcal{B}}}$ at $\bm\theta_{t}$ probabilistically preserve their relative difficulty rank at $\bm\theta_{t+1}$.
Theorem \ref{theorem_converge} further validates the convergence of the optimization pipeline.
Moreover, learning adaptation risk provides a risk distribution over the task space.

MPTS approximates $\text{CVaR}_{\alpha}$ optimization with efficiency and exploration benefits in an amortized manner.
Note that the standard $\text{CVaR}_{\alpha}$ evaluates a large set of candidate tasks $\hat{\mathcal{B}}=\frac{\mathcal{B}}{1-\alpha}$ and selects the $\mathcal{B}$ worst-performing ones to optimize. 
This results in a waste of $\frac{\alpha}{1-\alpha}\mathcal{B}$ task evaluations, which are expensive in either loss computation with large models or agent-environment interactions in RL.
MPTS proposes to improve this by replacing the expensive evaluation of $\hat{\mathcal{B}}$ tasks with a predictive inference step using the RPM.
Learning $p(\ell\vert\bm\tau, H_{1:t};\bm\theta_{t})$ enables efficient evaluation across infinite tasks with minimal computation, expanding the pseudo batch size $\hat{\mathcal{B}}$ for subset selection and fostering exploration. 
As shown in Lemma \ref{lemma_misrank_acute_angle}, MPTS introduces external approximation bias into the gradient.
For clarity, we treat MPTS as a risk minimization framework under specific acquisition criteria. 
As shown in Fig. \ref{fig_overall_mpts} and Fig. \ref{fig_basic_setup}c, its core workflow involves training the RPM $p(\ell\vert\bm\tau,H_{1:t};\bm\theta_{t})$, evaluating task-specific adaptation risk via posterior inference, and screening task subsets using the upper confidence bound (UCB) principle \citep{auer2002finite} for $(t+1)$-th optimization. 
These operations are formalized in Eq. (\ref{eq_mpts_workflows}), where the Monte Carlo estimate of the RPM yields the mean $m(\ell)$ and standard deviation $\sigma(\ell)$ of adaptation risk, while the acquisition function $\mathcal{A}(\cdot)$ quantifies total task subset risk.

\begin{subequations}\label{eq_mpts_workflows}
\begin{align}
    \text{Approximate Post-Adaptation Risk}:\ \max_{\bm\psi\in\bm\Psi}\mathcal{L}_{\text{ML}}(\bm\psi):=\ln p_{\bm\psi}(H_{t}\vert H_{1:t-1})\\
        \text{Amortized Task Evaluation}:\ p_{\bm\psi}(\ell\vert\bm\tau_{i},H_{1:t};\bm\theta_{t})\rightarrow\{m(\ell_i),\sigma(\ell_i)\}_{i=1}^{\hat{\mathcal{B}}}\\
        \text{Active Task Subset Selection}:\
        \mathcal{T}_{t+1}^{\mathcal{B}}=\arg\max_{\mathcal{T}^{\mathcal{B}}\subseteq\mathcal{T}^{\hat{\mathcal{B}}}_{t+1}:|\mathcal{T}^{\mathcal{B}}|=\mathcal{B}}\mathcal{A}(\mathcal{T}^{\mathcal{B}};\bm\phi,\bm\psi)  
\end{align}
\end{subequations}

Approximating optimization outcome relies on streaming VI \citep{broderick2013streaming,nguyen2017variational}, with the RPM a lightweight model.
Selecting a portion of challenging tasks to optimize, MPTS can be viewed as a biased surrogate of $\text{CVaR}_{1-\mathcal{B}/\hat{\mathcal{B}}}$ minimization while circumventing extra computations, annotations, or environment interactions.
This design not only enhances learning efficiency but also aligns with the overarching goals of robust adaptation.
Repeating the boxed steps of MPTS until convergence brings a robust adaptive machine learner, and the implementation details are attached in Section \ref{method_sec} and Algorithm \ref{alg_mpts}.    

\begin{algorithm}[H]
\SetAlgoLined
\SetKwInOut{Input}{Input}
\SetKwInOut{Output}{Output}
\Input{Task distribution $p(\tau)$;
 Task batch size $\mathcal{B}$;
 Candidate batch size $\hat{\mathcal{B}}$;
 Latest updated $\{\bm\psi,\bm\phi\}$;
 Latest history $H_{t-1}$;
 Iteration number $K$;
 Learning rate $\lambda_{2}$.}
\Output{Selected identifier batch $\{\bm\tau_{t,i}\}_{i=1}^{\mathcal{B}}$.}

\tcp{\textbf{Train the Risk Predictive Model via Stochastic Gradient Variational Bayes}}
\For{$i=1$ \KwTo $K$}
{   
Perform gradient updates given $H_{t-1}$:

\quad $\bm\phi\leftarrow\bm\phi+\lambda_{2}\nabla_{\bm\phi}\mathcal{G}_{\text{ELBO}}(\bm\psi,\bm\phi)$ in Eq. (\ref{eq_approx_elbo}b);

\quad 
$\bm\psi\leftarrow\bm\psi+\lambda_{2}\nabla_{\bm\psi}\mathcal{G}_{\text{ELBO}}(\bm\psi,\bm\phi)$ in Eq. (\ref{eq_approx_elbo}b);

}

\tcp{\textbf{Simulate Zero-shot, Few-shot Adaptation and SFT Results with Trained RPMs}}
Randomly sample $\{\bm\hat{\bm\tau}_{t,i}\}_{i=1}^{\hat{\mathcal{B}}}$ from $p(\tau)$;

Run amortized evaluation on candidate tasks $\{\delta_{i}:=\gamma_{0}m(\ell_i)+\gamma_{1}\sigma(\ell_i)\}_{i=1}^{\hat{\mathcal{B}}}$ in Eq. (\ref{eq_acq});

\tcp{\textbf{Active Subset Selection from Predicted Results}}
Rank $\{\delta_{i}\}_{i=1}^{\hat{\mathcal{B}}}$ and screen Top-$\mathcal{B}$ values;

Return the screened identifier subset $\{\bm\tau_{t,i}\}_{i=1}^{\mathcal{B}}$.

\caption{Model Predictive Task Sampling}
\label{alg_mpts}
\end{algorithm}

\subsection{Robust adaptation benchmarks and baselines.}

The experimental design considers the benchmark typicality and the practical challenges. 
Downstream tasks span pattern recognition and sequential decision-making, with certain experiments involving multimodal foundation models.
These experiments mainly examine few-shot adaptation and include (1) \texttt{K-shot} sinusoid regression \citep{finn2017model}, (2) \texttt{N-way K-shot} image classification \citep{liu2020adaptive,gondal2024domain} with CLIP \citep{radford2021learning} models, and (3) Meta-RL \citep{finn2017model}.
Additionally, MPTS validates scenarios like (4) robotic DR \citep{mehta2020active} for zero-shot adaptation and (5) SFT CLIP models towards image classification.

This study primarily compares MPTS with risk minimization principles and focuses on robustness improvement. 
While these methods, including MPTS, are agnostic to zero-shot, few-shot learning, or finetuning techniques, we adopt SoTA backbones for experiments.
For sinusoid regression and Meta-RL, we use MAML \citep{finn2017model} as default. 
As CLIP has strong zero-shot performance, we extend it with MaPLe \citep{khattak2023maple} for \texttt{N-way K-shot} image classification. 
For robotic DR in Ergo-Reacher and Lunar-Lander \citep{mehta2020active}, we employ TD3 \citep{fujimoto2018addressing} due to its stability. 
In SFT, we again use MaPLe for prompt-tuning in image classification.

Baselines include Empirical Risk Minimization (ERM) \citep{vapnik1998statistical}, Distributionally Robust Risk Minimization (DRM) \citep{rajeswaran2022epopt,evans2023bad,greenberg2024train,wang2024simple,lv2024theoretical}, and Group Distributionally Robust Risk Minimization (GDRM) \citep{sagawa2019distributionally,xie2024doremi,hejnaremix}.
Meanwhile, for certain scenarios, we further include some robust optimization methods such as Excess Multi-Task Learning (EMTL) \citep{he2024robust}, Difficulty-Aware Task Sampler (DATS) \citep{toloubidokhti2023dats}, Online Hard Task Mining Sampler (OHTM) \citep{kumar2023effect}, and Task Difficulty Prioritized Sampler (TDPS) \citep{wang2024towards} for comparison.
MPTS and EMTL both aim to prioritize challenging tasks based on their statistics, but they differ in their approaches: MPTS uses a subset selection method, while EMTL employs a reweighting strategy.
Accordingly, adaptation robustness is evaluated via $\text{CVaR}_{\alpha}$ across validation/testing tasks with $\alpha=\{0.9, 0.7, 0.5\}$, including some OOD results. 
We also compare computational cost, memory usage, and sample efficiency.
For fairness, all baselines share the same task batch $\mathcal{B}$ in optimization, excluding pruned easier tasks. 
ERM, GDRM, EMTL, DATS, TDPS, and OHTM use batch size $\mathcal{B}$, while DRM samples $\hat{\mathcal{B}}=2\mathcal{B}$, filtering half for stable optimization. 
See Appendix~\ref{sec_supp_exp} for details.

\subsection{Demonstration of the MPTS's role in \texttt{K-shot} sinusoid regression}

In \texttt{K-shot} sinusoid regression \citep{finn2017model}, the function family $\{f(x)=a_{i}\sin(x-b_{i})\vert (a_i,b_i)\in[0.1,5.0]\times[0.0,\pi]\}$ is specified by the identifier $\bm\tau=[a,b]$.
This serves as a toy case to illustrate MPTS and the role of the RPM.

The RPM allows for roughly scoring the task difficulty over iterations.
In MPTS, for the screened subset at $(t+1)$-th iteration, we track the predicted Top-$\mathcal{B}$ risk values $\{\bar{\ell}_{t+1,i}:\approx\mathbb{E}_{q_{\bm\phi}(\bm z_{t}\vert H_{t})}[p_{\bm\psi}(\ell\vert\bm\tau_{t+1,i},H_{1:t})]\}_{i=1}^{\mathcal{B}}$ and corresponding exact evaluations $\{\ell_{t+1,i}\}_{i=1}^{\mathcal{B}}$ from $\bm\theta_{t+1}$ to compute the Pearson correlation coefficient (PCC) $\rho_{\bar{\ell},\ell}:=\frac{\sum_{i=1}^{\mathcal{B}}(\bar{\ell}_{t+1,i}-\text{Mean}[\{\bar{\ell}_{t+1,.}\}])(\ell_{t+1,i}-\text{Mean}[\{\ell_{t+1,.}\}])}{\sqrt{\sum_{i=1}^{\mathcal{B}}(\bar{\ell}_{t+1,i}-\text{Mean}[\{\bar{\ell}_{t+1,.}\}])^{2}}\sqrt{\sum_{i=1}^{\mathcal{B}}(\ell_{t+1,i}-\text{Mean}[\{\ell_{t+1,.}\}])^{2}}}$.
For continuous risk values, PCC reasonably quantifies the correlation effect of ranking in a batch.
The RPM amortizes the exact evaluation $\ell(\mathcal{D}_{\tau}^{Q},\mathcal{D}_{\tau}^{S};\bm\theta_{t})$ $\forall\bm\tau\in\mathcal{T}\ \text{and}\ \bm\theta_{t}\in\bm\Theta$ using risk histories, roughly scoring adaptation difficulty for next iteration.
Only if the RPM approximately ranks tasks, MPTS can trust amortized evaluations for worst subset selection.

\begin{figure}[!htbp]
\begin{center}
\centerline{\includegraphics[width=1.0\textwidth]{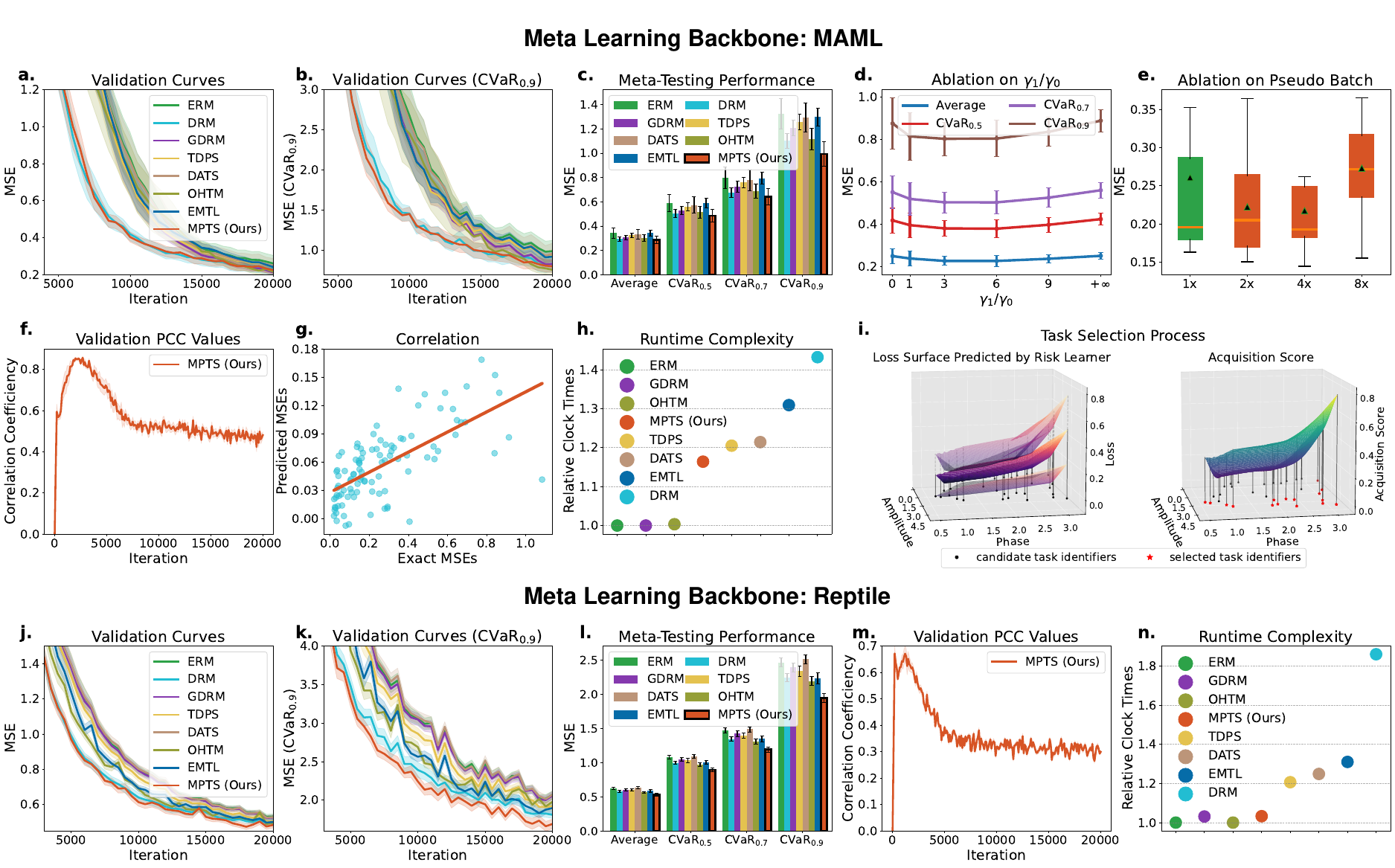}}
\caption{\textbf{\texttt{K-shot} Sinusoid Regression Results (10 Runs).}
Note that \textbf{a}-\textbf{i} are results with MAML as the backbone, while \textbf{j}-\textbf{n} reports the results with Reptile as the backbone.
\textbf{a.} Shown are curves of averaged MSEs with standard error of the mean (SEM) on the validation task set during meta-training for all methods.
\textbf{b.} Curves illustrate the $\text{CVaR}_{0.9}$ MSEs with SEM on the validation task set during the meta-training process.
\textbf{c.} The meta-trained machine learners are tested on a fixed task set, reporting the average MSEs and CVaR values with SEM.
\textbf{d.} Displayed are meta-testing results with SEM using MPTS machine learners trained by various $\gamma_1/\ \gamma_0$ ratios.
\textbf{e.} Displayed are meta-testing results using MPTS machine learners trained in various pseudo batch sizes, i.e., $\hat{\mathcal{B}}=\{1\mathcal{B},2\mathcal{B},4\mathcal{B},8\mathcal{B}\}$.
\textbf{f.} The PCC values with SEM are tracked during meta-training.
\textbf{g.} At a specific iteration, the statistical correlation between predicted and exact adaptation risk values of the task batch is visualized with overall $\rho_{\bar{\ell},\ell}=0.669$.
\textbf{h.} The required relative run-time is computed for all methods during meta-training with ERM as the anchor.
\textbf{i.} At some meta-training time-step, we visualize the subset selection from the pseudo batch under the RPM.
\textbf{j.} We illustrate the curves of averaged MSEs with SEM on the validation task set during Reptile meta-training for all relevant baselines.
\textbf{k.} We track the corresponding $\text{CVaR}_{0.9}$ MSEs with SEM on the validation task set throughout the Reptile meta-training.
\textbf{l.} Reported are the tested average MSEs and CVaR values with SEM using the meta-trained machine learners on a fixed task set.
\textbf{m.} The PCC values with SEM are tracked for MPTS during Reptile meta-training.
\textbf{n.} We compute the relative run-time for all methods during meta-training with ERM as the anchor.
}
\label{fig_meta_sin}
\end{center}
\end{figure}

As shown in Fig. \ref{fig_meta_sin}f, $\rho_{\bar{\ell},\ell}$ remains between 0.4–0.8 across iterations, validating the reliability of the RPM in predicting adaptation difficulty. 
However, PCC declines over time, a trend also observed across experiments, likely due to model $\bm\theta_t$'s convergence. 
This reduces task diversity, negatively affecting the RPM's training after local task space overoptimization.
Fig. \ref{fig_meta_sin}g shows the statistical correlation between predicted and exact adaptation risk at a specific iteration. 
Scattered points demonstrate strong overall alignment, despite varying value scales between iterations. 
Notably, difficult tasks with high mean square errors (MSEs) are well identified and clustered around or above the correlation slope along the $x$-axis.
MPTS can also accelerates the learning process and improves comprehensive adaptation performance under active sampling.
In Fig. \ref{fig_meta_sin}a-b, MPTS converges faster in average and $\text{CVaR}_{0.9}$ MSEs and reaches performance comparable to other baselines with less iterations, e.g., 15K steps, due to its uncertainty-guided worst-case acquisition. 
DRM processes 2$\mathcal{B}$ tasks to filter half per iteration, raising 0.5× computational overhead over ERM (see Fig. \ref{fig_meta_sin}h). 
In contrast, MPTS incurs only 0.14× runtime increase, a marginal overhead.
To illustrate active task sampling, Fig. \ref{fig_meta_sin}i visualizes predicted risk values over the task space. 
Selected tasks favor regions with high deviations, clustering in high-risk areas.

In meta-testing, Fig. \ref{fig_meta_sin}c shows that MPTS, OHTM, and DRM achieve the lowest average and $\text{CVaR}_{\alpha}$ MSEs, with their advantage over other baselines increasing at $\alpha$.
EMTL, DATS and TDPS require more computational resources to achieve slight gains in robustness metrics.
Prior work \citep{lv2024theoretical} confirms that DRM's efficiency is sacrificed for robustness, relying on intensive task evaluation. 
Using MAML, gradient-based inner-loop adaptation further increases overhead, whereas MPTS bypasses it via probabilistic prediction, reducing computational cost.

Meanwhile, the appropriate hyper-parameter setup can secure performance and efficiency.
We first analyze the acquisition function $\mathcal{A}(\mathcal{T}^{\mathcal{B}};\bm\phi,\bm\psi)$ by varying trade-off parameters $\{\gamma_0,\gamma_1\}$ in Fig. \ref{fig_overall_mpts}d and Eq. (\ref{eq_acq}). 
Meta-testing machine learners trained with $\frac{\gamma_1}{\gamma_0}=\{1.0,3.0,6.0,9.0\}$, $\gamma_0=0.0$ and $\gamma_1=0.0$ shows that higher uncertainty weights lower average MSEs (Fig. \ref{fig_meta_sin}d). 
However, removing worst-case considerations ($\gamma_0=0.0$) weakens performance.
We further examine the impact of pseudo batch size $\hat{\mathcal{B}}$ in Fig. \ref{fig_meta_sin}e. 
Increasing $\hat{\mathcal{B}}$ reduces average MSEs, but excessively large values (e.g., $\hat{\mathcal{B}}=8\times\mathcal{B}$) degrade performance. 
This occurs because an enlarged identifier set under worst-case selection might over-optimize local task regions, hindering global generalization.
Thus, MPTS configuration follows two principles:
(i) $\hat{\mathcal{B}}$ should be moderate to encourage exploration while preventing excessive local optimization.
(ii) Since adaptation robustness is the priority, we consistently set $\gamma_0\in\mathbb{R}_{+}$ as the default in all experiments.

In addition, MPTS is plug-and-play and agnostic to the backbone.
We further conduct experiments using another SoTA meta-optimizer Reptile \citep{nichol2018first} as the backbone. 
From Fig. \ref{fig_meta_sin}j–k, the convergence trends of most algorithms are similar to those observed under MAML in Fig. \ref{fig_meta_sin}a–b. 
OHTM lags slightly behind MPTS, and the overall MSEs are higher. 
This observation is corroborated by the meta-testing results in Fig. \ref{fig_meta_sin}l, where the secondary method OHTM outperforms the remaining baselines with a lower computational cost. 
MPTS continues to exhibit a positive correlation between predicted and exactly evaluated adaptation results (Fig. \ref{fig_meta_sin}m). 
The order of computational complexity across all methods remains the same as under MAML, though the relative scale in Fig. \ref{fig_meta_sin}n varies slightly.

\subsection{Few-Shot adaptation benefits from MPTS in robustness and learning efficiency}

In terms of few-shot image classification, we perform \texttt{5-way 1-shot} image classification using MaPLe, with six meta-training datasets from ImageNet-CG \citep{hendrycks2019benchmarking}, ImageNet-CI \citep{hendrycks2019benchmarking}, ImageNet-CS \citep{hendrycks2019benchmarking}, ImageNet-A \citep{hendrycks2021natural}, ImageNet-S \citep{wang2019learning} and ImageNet-R \citep{hendrycks2021many}.
The efficiency row in Table \ref{table_few_shot_test} compares computational time and memory usage across methods during meta-training.
The overhead from optimizing RPMs in MPTS is negligible, whereas DRM incurs 1.3$\times$ computational time and 1.6$\times$ memory usage relative to ERM.
DATS and TDPS also significantly suffer from more computational complexity issues due to the additional gradient computations required to derive task-specific weights.
Similarly, EMTL also takes 1.7$\times$ runtime overhead and 1.91$\times$ memory consumption for calculating task importance.
As a memory retrieval strategy, OHTM occupies 2.54$\times$ memory to prioritize difficult tasks during optimization.
In Table \ref{table_few_shot_test}, MPTS consistently surpasses risk-aware baselines and achieves significant performance gains, i.e., 0.36-2.32\% accuracy increase, over ERM across all benchmark datasets.
The improvement is particularly prominent under the stringent $\text{CVaR}_{0.9}$ metric, e.g., +2.32\% on ImageNet-CS and +2.3\% on ImageNet-S compared to ERM. 
In contrast, DRM provides moderate improvements after sacrificing more computations during meta-training.
EMTL serves as the runner-up on several benchmarks, such as ImageNet-CG and ImageNet-A.
There is no evident runner-up.
OHTM, DATS, and TDPS, suffer from diminishing average and CVaR accuracies on ImageNet-CI while mostly beating ERM in other cases.
Overall, this benchmark result demonstrates the comprehensive merits of prioritizing challenging tasks, such as optimizing worst-case performance, which surprisingly improves the average results.
This contradicts the following conclusion in Meta-RL yet is consistent with the consequence of prioritizing complex tasks during LLM pretraining \citep{dubey2024llama}. 

\begingroup
\setlength{\tabcolsep}{8pt}
\begin{small}
\begin{table}[!htbp]
   \begin{center}
    \caption{\textbf{\texttt{5-Way 1-Shot} Meta-Testing Classification Results on Various Datasets and Efficiency Comparison (10 Runs).}
    We report testing $\text{CVaR}_{0.9}$, $\text{CVaR}_{0.7}$, $\text{CVaR}_{0.5}$ and average accuracies with corresponding SEMs evaluated by the meta-trained machine learner on ID and OOD datasets. 
    The \textbf{best results} are in bold with the \underline{runner-up} underlined, and MPTS's performance gains over ERM $\Delta\uparrow$ are marked in blue.
    With experiments on ImageNet-A as an example, we report the memory cost and runtime relative to ERM during meta-training.}
    \label{table_few_shot_test}
    \resizebox{1.0\columnwidth}{!}{
    \begin{tabular}{|c|c|c|c|c|c|c|c|c|cc|}
      \toprule %
      Dataset & Metrics & ERM & DRM & GDRM & OHTM & DATS & TDPS & EMTL & MPTS (Ours) & $\Delta\uparrow$ \\
      \toprule %
      ImageNet-CG \citep{hendrycks2019benchmarking}
      & $\text{CVaR}_{0.9}$ & 76.87$\pm${\scriptsize 0.31} & 77.47$\pm${\scriptsize 0.24} & 77.56$\pm${\scriptsize 0.19} & 76.52$\pm${\scriptsize 0.14} & 77.55$\pm${\scriptsize 0.21} & 77.28$\pm${\scriptsize 0.17} & \underline{77.59$\pm${\scriptsize 0.25}} & \textbf{78.16$\pm${\scriptsize 0.25}} & +1.29 \\
      & $\text{CVaR}_{0.7}$ & 82.04$\pm${\scriptsize 0.21} & 82.47$\pm${\scriptsize 0.20} & \underline{82.62$\pm${\scriptsize 0.20}} & 81.34$\pm${\scriptsize 0.17} & 82.31$\pm${\scriptsize 0.18} & 82.33$\pm${\scriptsize 0.24} & 82.56$\pm${\scriptsize 0.14} & \textbf{82.95$\pm${\scriptsize 0.26}} & +0.91 \\
      & $\text{CVaR}_{0.5}$ & 84.62$\pm${\scriptsize 0.24} & 84.99$\pm${\scriptsize 0.17} & 85.16$\pm${\scriptsize 0.22} & 84.06$\pm${\scriptsize 0.29} & 85.19$\pm${\scriptsize 0.14} & \underline{85.29$\pm${\scriptsize 0.23}} & 85.14$\pm${\scriptsize 0.17} &  \textbf{85.33$\pm${\scriptsize 0.26}} & +0.71\\
      & Avg & 89.22$\pm${\scriptsize 0.22} & 89.54$\pm${\scriptsize 0.16} & 89.49$\pm${\scriptsize 0.18} & 88.59$\pm${\scriptsize 0.17} & 89.45$\pm${\scriptsize 0.12} & \underline{89.69$\pm${\scriptsize 0.11}} & 89.34$\pm${\scriptsize 0.21} & \textbf{89.86$\pm${\scriptsize 0.14}} & +0.64 \\
      \midrule
      ImageNet-CI \citep{hendrycks2019benchmarking} 
      & $\text{CVaR}_{0.9}$ & 80.40$\pm${\scriptsize 0.18} & \underline{80.49$\pm${\scriptsize 0.22}} & 80.26$\pm${\scriptsize 0.19} & 77.08$\pm${\scriptsize 0.18} & 78.86$\pm${\scriptsize 0.15} & 78.07$\pm${\scriptsize 0.19} & 80.41$\pm${\scriptsize 0.16} & \textbf{80.94$\pm${\scriptsize 0.19}} & +0.54 \\
      & $\text{CVaR}_{0.7}$ & 84.98$\pm${\scriptsize 0.17} & \underline{85.05$\pm${\scriptsize 0.19}} & 84.90$\pm${\scriptsize 0.22} & 81.60$\pm${\scriptsize 0.11} & 83.66$\pm${\scriptsize 0.22} & 83.16$\pm${\scriptsize 0.11} & 85.04$\pm${\scriptsize 0.18} & \textbf{85.70$\pm${\scriptsize 0.24}} & +0.72\\
      & $\text{CVaR}_{0.5}$ & 87.01$\pm${\scriptsize 0.20} & \underline{87.47$\pm${\scriptsize 0.26}} & 87.04$\pm${\scriptsize 0.18} & 84.14$\pm${\scriptsize 0.22} & 86.08$\pm${\scriptsize 0.16} & 85.75$\pm${\scriptsize 0.13} & 87.13$\pm${\scriptsize 0.15} & \textbf{87.74$\pm${\scriptsize 0.15}} & +0.73 \\
      & Avg & 91.20$\pm${\scriptsize 0.23} & \underline{91.40$\pm${\scriptsize 0.18}} & 91.21$\pm${\scriptsize 0.17} & 88.81$\pm${\scriptsize 0.23} & 90.97$\pm${\scriptsize 0.13} & 90.67$\pm${\scriptsize 0.18} & 91.26$\pm${\scriptsize 0.10} & \textbf{91.56$\pm${\scriptsize 0.12}} & +0.36 \\
      \midrule
      ImageNet-CS \citep{hendrycks2019benchmarking} 
      & $\text{CVaR}_{0.9}$ & 76.15$\pm${\scriptsize 0.20} & 77.78$\pm${\scriptsize 0.24} & 76.25$\pm${\scriptsize 0.15} & \underline{77.84$\pm${\scriptsize 0.23}} & 77.77$\pm${\scriptsize 0.16} & 77.47$\pm${\scriptsize 0.16} & 77.42$\pm${\scriptsize 0.19} & \textbf{78.47$\pm${\scriptsize 0.23}} & +2.32  \\
      & $\text{CVaR}_{0.7}$ & 81.55$\pm${\scriptsize 0.19} & 82.64$\pm${\scriptsize 0.16} & 81.42$\pm${\scriptsize 0.17} & 82.59$\pm${\scriptsize 0.20} & \underline{82.81$\pm${\scriptsize 0.19}} & 82.31$\pm${\scriptsize 0.17} & 81.60$\pm${\scriptsize 0.20} & \textbf{83.13$\pm${\scriptsize 0.16}} & +1.58 \\
      & $\text{CVaR}_{0.5}$ & 84.44$\pm${\scriptsize 0.19} & 85.16$\pm${\scriptsize 0.25} & 84.19$\pm${\scriptsize 0.15} & \underline{85.23$\pm${\scriptsize 0.18}} & 84.67$\pm${\scriptsize 0.20} & 84.84$\pm${\scriptsize 0.13} & 84.90$\pm${\scriptsize 0.14} & \textbf{85.87$\pm${\scriptsize 0.22}} & +1.43 \\
      & Avg & 89.30$\pm${\scriptsize 0.20} & 89.80$\pm${\scriptsize 0.21} & 89.11$\pm${\scriptsize 0.17} & \underline{89.87$\pm${\scriptsize 0.25}} & 89.60$\pm${\scriptsize 0.16} & 89.82$\pm${\scriptsize 0.20} & 89.42$\pm${\scriptsize 0.22} & \textbf{90.28$\pm${\scriptsize 0.11}} & +0.98 \\
      \midrule
      ImageNet-A \citep{hendrycks2021natural} 
      & $\text{CVaR}_{0.9}$ & 76.71$\pm${\scriptsize 0.22} & 77.36$\pm${\scriptsize 0.26} & 77.45$\pm${\scriptsize 0.22} & 76.81$\pm${\scriptsize 0.18} & \underline{77.63$\pm${\scriptsize 0.21}} & 77.12$\pm${\scriptsize 0.20} & 77.54$\pm${\scriptsize 0.16} & \textbf{77.96$\pm${\scriptsize 0.11}} & +1.25 \\
      & $\text{CVaR}_{0.7}$ & 82.03$\pm${\scriptsize 0.25} & 82.52$\pm${\scriptsize 0.27} & 82.37$\pm${\scriptsize 0.20} & 82.72$\pm${\scriptsize 0.10} & 82.69$\pm${\scriptsize 0.20} & 82.04$\pm${\scriptsize 0.23} & \underline{82.87$\pm${\scriptsize 0.20}} & \textbf{83.43$\pm${\scriptsize 0.18}} & +1.40 \\
      & $\text{CVaR}_{0.5}$ & 84.50$\pm${\scriptsize 0.22} & 85.15$\pm${\scriptsize 0.19} & 85.34$\pm${\scriptsize 0.25} & 85.20$\pm${\scriptsize 0.23} & 85.77$\pm${\scriptsize 0.19} & 84.92$\pm${\scriptsize 0.20} & \underline{85.80$\pm${\scriptsize 0.14}} & \textbf{86.74$\pm${\scriptsize 0.24}} & +2.24 \\
      & Avg & 89.38$\pm${\scriptsize 0.21} & 90.31$\pm${\scriptsize 0.17} & 90.29$\pm${\scriptsize 0.17} & \underline{90.68$\pm${\scriptsize 0.20}} & 90.58$\pm${\scriptsize 0.18} & 90.03$\pm${\scriptsize 0.23} & 90.21$\pm${\scriptsize 0.15} & \textbf{91.10$\pm${\scriptsize 0.21}} & +1.72 \\
      \midrule
      ImageNet-S \citep{wang2019learning} 
      & $\text{CVaR}_{0.9}$ & 82.58$\pm${\scriptsize 0.21} & 83.87$\pm${\scriptsize 0.10} & 83.18$\pm${\scriptsize 0.22} & 82.23$\pm${\scriptsize 0.18} & 83.60$\pm${\scriptsize 0.19} & \underline{83.93$\pm${\scriptsize 0.15}} & 83.51$\pm${\scriptsize 0.17} & \textbf{84.88$\pm${\scriptsize 0.14}} & +2.30 \\
      & $\text{CVaR}_{0.7}$ & 87.23$\pm${\scriptsize 0.23} & \underline{88.45$\pm${\scriptsize 0.21}} & 87.32$\pm${\scriptsize 0.14} & 87.08$\pm${\scriptsize 0.13} & 88.32$\pm${\scriptsize 0.19} & 88.06$\pm${\scriptsize 0.16} & 88.43$\pm${\scriptsize 0.19} & \textbf{89.27$\pm${\scriptsize 0.12}} & +2.04\\
      & $\text{CVaR}_{0.5}$ & 89.56$\pm${\scriptsize 0.19} & 89.96$\pm${\scriptsize 0.22} & 90.06$\pm${\scriptsize 0.18} & 90.01$\pm${\scriptsize 0.22} & \underline{91.16$\pm${\scriptsize 0.22}} & 90.65$\pm${\scriptsize 0.20} & 90.18$\pm${\scriptsize 0.19} & \textbf{91.52$\pm${\scriptsize 0.19}} & +1.96 \\
      & Avg & 93.52$\pm${\scriptsize 0.13} & \underline{94.28$\pm${\scriptsize 0.10}} & 94.13$\pm${\scriptsize 0.11} & 94.19$\pm${\scriptsize 0.15} & 94.15$\pm${\scriptsize 0.13} & 94.18$\pm${\scriptsize 0.14} & 94.11$\pm${\scriptsize 0.14} & \textbf{94.84$\pm${\scriptsize 0.08}} & +1.32\\
      \midrule
      ImageNet-R \citep{hendrycks2021many} 
      & $\text{CVaR}_{0.9}$ & 88.28$\pm${\scriptsize 0.13} & 88.89$\pm${\scriptsize 0.15} & 87.33$\pm${\scriptsize 0.27} & 88.73$\pm${\scriptsize 0.16} & \underline{89.20$\pm${\scriptsize 0.21}} & 88.52$\pm${\scriptsize 0.20} & 88.75$\pm${\scriptsize 0.15} & \textbf{89.78$\pm${\scriptsize 0.16}} & +1.50 \\
      & $\text{CVaR}_{0.7}$ & 91.47$\pm${\scriptsize 0.13} & \underline{91.95$\pm${\scriptsize 0.16}} & 91.11$\pm${\scriptsize 0.16} & 91.41$\pm${\scriptsize 0.19} & \underline{91.95$\pm${\scriptsize 0.23}} & 90.89$\pm${\scriptsize 0.20} & 91.32$\pm${\scriptsize 0.18} & \textbf{93.27$\pm${\scriptsize 0.19}} & +1.80\\
      & $\text{CVaR}_{0.5}$ & 93.42$\pm${\scriptsize 0.12} & \underline{93.87$\pm${\scriptsize 0.17}} & 93.20$\pm${\scriptsize 0.20} & 92.75$\pm${\scriptsize 0.15} & 93.27$\pm${\scriptsize 0.21} & 92.48$\pm${\scriptsize 0.18} & 93.20$\pm${\scriptsize 0.10} & \textbf{94.69$\pm${\scriptsize 0.18}} & +1.27 \\
      & Avg & 96.02$\pm${\scriptsize 0.14} & \underline{96.39$\pm${\scriptsize 0.16}} & 95.90$\pm${\scriptsize 0.18} & 95.32$\pm${\scriptsize 0.17} & 95.85$\pm${\scriptsize 0.16} & 95.64$\pm${\scriptsize 0.17} & 96.06$\pm${\scriptsize 0.13} & \textbf{96.91$\pm${\scriptsize 0.12}} & +0.89 \\
      \bottomrule
      Efficiency  &Runtime   & 1.00 & 1.33 & 1.00 & 1.09 & 1.68 & 1.68 & 1.70 & 1.01 &   \\
       &Memory  & 1.00 & 1.61 & 1.00 & 2.54 & 1.99 & 1.99 & 1.91 & 1.01 &   \\
      \bottomrule %
    \end{tabular}
    }
  \end{center}
\end{table}    
\end{small}
\endgroup

\begin{figure}[!htbp]
\begin{center}
\centerline{\includegraphics[height=0.80\textheight]{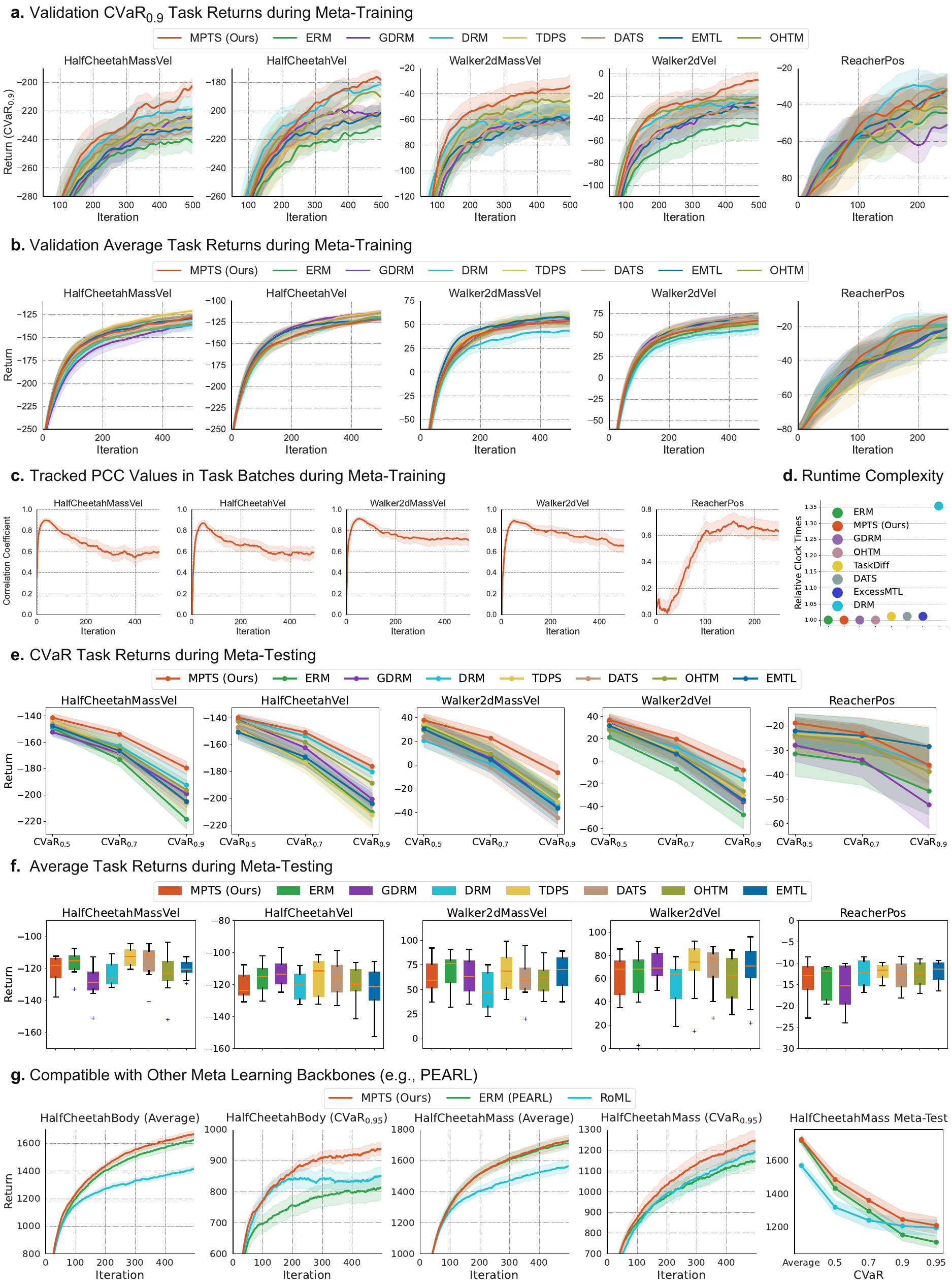}}
\caption{\textbf{Meta-RL Results on Mujoco Environments (10 Runs).}
\textbf{a.} The cumulative returns with SEM belonging to $\text{CVaR}_{0.9}$ validation MDPs are displayed during meta-training.
\textbf{b.} We compute the average cumulative returns with SEMs on validation MDPs during meta-training.
\textbf{c.} Tracked are the RPM's PCC values with SEMs over training iterations.
\textbf{d.} The relative clock time quantifies the computational complexity for all methods on Walker2dVel, where ERM's runtime works as the anchor.
\textbf{e.} We report $\text{CVaR}_{\alpha}$ returns with SEM of meta-testing MDPs.
\textbf{f.} The box-plot reports results averaged over meta-testing MDPs.
\textbf{g.} With PEARL \citep{rakelly2019efficient} as the Meta-RL backbone, we illustrate the learning curves and meta-testing results with SEM on HalfCheetahBody and HalfCheetahMass from RoML \citep{greenberg2023train} baseline.
}
\label{fig_merged_metarl}
\end{center}
\end{figure}

In terms of Meta-RL, we first analyze meta-training results in Fig. \ref{fig_merged_metarl}a-b. 
MPTS achieves the highest $\text{CVaR}_{0.9}$ validation returns on most benchmarks. 
DRM sacrifices average returns on HalfCheetahMassVel, HalfCheetahVel, and Walker2dVel, whereas MPTS maintains average performance comparable to ERM on HalfCheetahMassVel and Walker2dVel. 
EMTL, GDRM, OHTM, DATS, and TDPS are entangled in learning curves and behave intermediate performance, while DRM balances average and $\text{CVaR}_{0.9}$ returns, excelling on ReacherPos.
Fig. \ref{fig_merged_metarl}c witnesses the RPM's strong task difficulty discrimination capability, measured by $\rho_{\bar{\ell},\ell}$.
In Fig. \ref{fig_merged_metarl}d, DRM consumes 1.35$\times$ runtime on Walker2dVel due to extra environment interactions, while MPTS avoids this inefficiency. 
Meta-testing results in Fig. \ref{fig_merged_metarl}e-f highlight MPTS's robustness, with more return gains at higher $\alpha$ values. 
With increase of $\alpha$ in $\text{CVaR}_{\alpha}$, performance difference across methods grows significant.
In extreme cases, i.e., $\text{CVaR}_{0.9}$, MPTS surpasses ERM by over 20\% on all benchmarks, and DRM mostly dominates other baselines.
Average performance varies: Walker2dMassVel and Walker2dVel show minor differences, while HalfCheetahMassVel favors MPTS with slightly higher variance. 
HalfCheetah marginally benefits GDRM and ERM, whereas ReacherPos favors EMTL in $\text{CVaR}_{0.9}$, followed by MPTS and DRM with reduced variance. 
Overall, methods except DRM are mostly entangled in average performance, while DRM tends to sacrifice computational efficiency for adaptation robustness increase.
The primary goal of robust optimization is to enhance performance in tail risk or OOD tasks. 
Hence, methods often trade off worst-case and average performance in Meta-RL, as implied in work \citep{greenberg2024train}.

Moreover, Fig. \ref{fig_merged_metarl}g reports the performance of MPTS with PEARL \citep{rakelly2019efficient} on environments and configurations adopted from RoML \citep{greenberg2024train}, a representative robust Meta-RL algorithm. 
The learning curves verify the trade-off between average and worst-case performance in ERM and RoML. 
In contrast, MPTS maintains average returns while substantially outperforming RoML in $\text{CVaR}_{0.95}$, a behavior distinct from the MAML backbone results. 
The meta-testing results on HalfCheetahMass corroborate this trend, consistent with the meta-training phase, where MPTS outperforms ERM by over 9\% in $\text{CVaR}_{0.95}$ and exceeds RoML in average return by approximately 10\%.

\subsection{MPTS retains multi-faced advantages in zero-shot continuous control}

\begin{figure}[!htbp]
\begin{center}
\centerline{\includegraphics[width=0.95\textwidth]{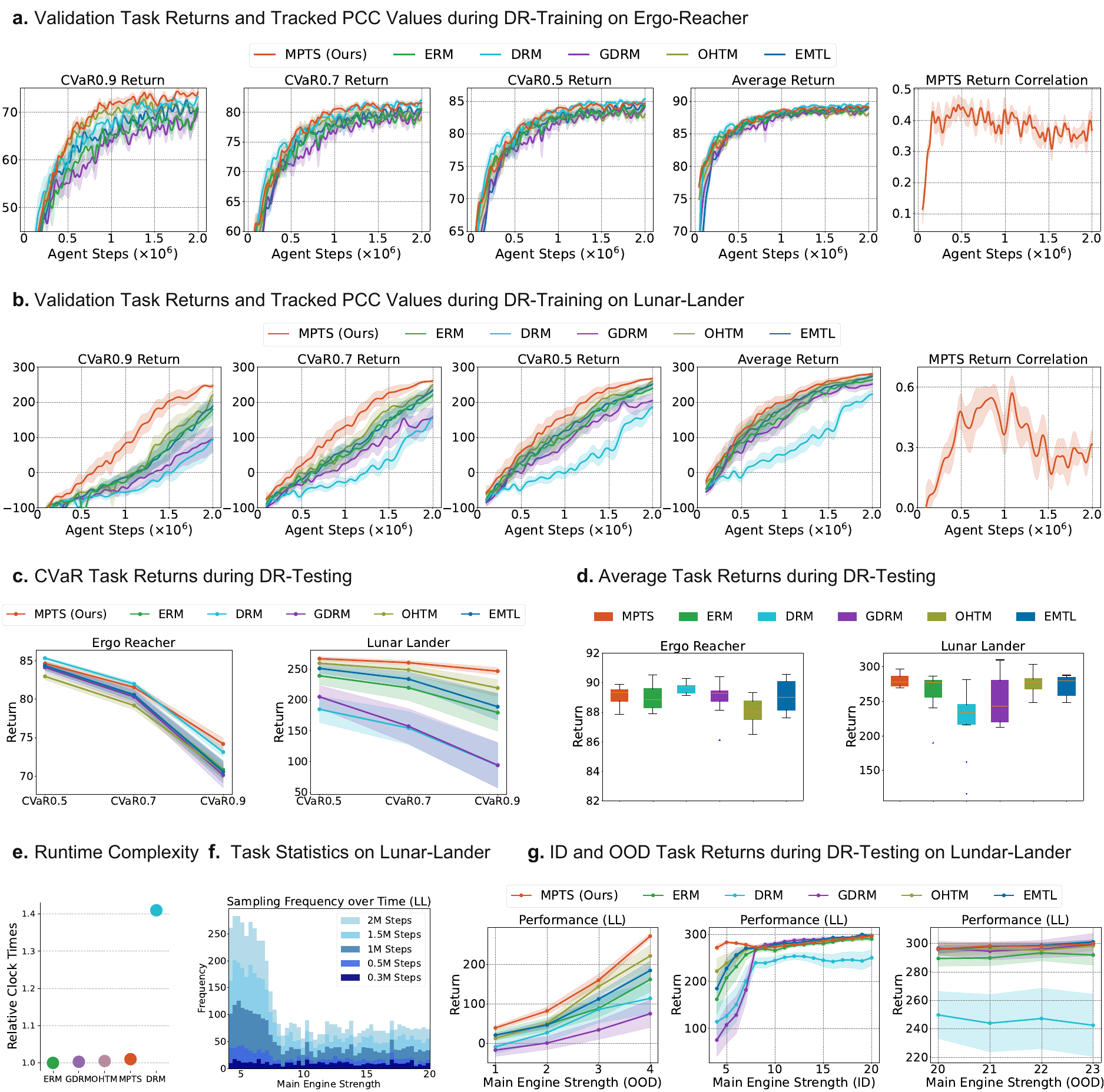}}
\caption{\textbf{DR Results on Ergo-Reacher and Lunar-Lander (10 Runs).}
\textbf{a.} In Ergo-Reacher, the $\text{CVaR}_{0.9}$, $\text{CVaR}_{0.7}$, $\text{CVaR}_{0.5}$, and average cumulative returns with SEM on validation MDPs are reported.
The RPM's PCC values with SEM are tracked during DR training.
\textbf{b.} In Lunar-Lander, the cumulative returns on validation MDPs with SEM are illustrated.
The RPM's PCC values with SEM are tracked during DR training.
\textbf{c.} We test the DR-trained policies on the fixed MDP set and report the $\text{CVaR}_{\alpha}$ cumulative returns with SEM.
\textbf{d.} The returns averaged over DR-testing MDPs are illustrated.
\textbf{e.} The required runtime is computed for all methods on Lunar-Lander.
\textbf{f.} In Lunar-Lander, shown are frequencies of sampled identifiers using MPTS during DR training.
\textbf{g.} In Lunar-Lander, we test the trained policies in both in-distribution (ID) domains and out-of-distribution (OOD) domains to report each task's average returns with SEM.
}
\label{fig_merged_dr}
\end{center}
\end{figure}

Overall, MPTS dominates the overall performance in DR training.
In Fig. \ref{fig_merged_dr}a Ergo-Reacher, distinguished from Meta-RL finding, MPTS and DRM improve both average and $\text{CVaR}_{\alpha}$ performance. 
This likely stems from MPTS’s broader task exploration via larger $\hat{\mathcal{B}}$. 
Meanwhile, $\rho_{\bar{\ell},\ell}$ fluctuates near 0.4 throughout training.
In Fig. \ref{fig_merged_dr}b Lunar-Lander, MPTS maintains the leading trend in average and $\text{CVaR}_{\alpha}$ returns, followed by EMTL and OHTM.
In contrast, DRM and GDRM not only underperform in average returns but also achieve the lowest $\text{CVaR}_{0.9}$ values, failing in robust optimization. 
Naively worst-case selection or reweighting tends to degrade performance when unsolvable tasks are frequently sampled. 
MPTS mitigates this by balancing worst-case and uncertainty-guided selection, preventing over-optimization on a finite set of difficult MDPs. 
Here, $\rho_{\bar{\ell},\ell}$ peaks above 0.6 before stabilizing near 0.3, consistent with prior findings that task selection converges, reducing task difficulty discrimination.
MPTS’s runtime in Lunar-Lander is comparable to ERM and GDRM in Fig. \ref{fig_merged_dr}e. 
In Lunar-Lander, the identifier $\bm\tau\in\mathbb{R}_{+}$ represents the main engine strength.
Fig. \ref{fig_merged_dr}f shows task sampling frequency, where MPTS favors lower-engine-strength MDPs while still exploring all engine-strengths early in training.

Also, MPTS facilitates policy adaptation in the worst-case and OOD MDPs.
For DR-trained policies, Fig. \ref{fig_merged_dr}c-d confirm MPTS and DRM's superior $\text{CVaR}_{\alpha}$ returns in Ergo-Reacher, while ERM exhibits a minor dip in average returns. 
In Lunar-Lander, MPTS attains the highest $\text{CVaR}_{\alpha}$ returns, remaining stable even as $\alpha$ increases—outperforming ERM by over 20\%. 
Additionally, MPTS, OHTM and ERM yield top average returns with minimal variance.
For OOD generalization, we shift $\bm\tau$'s range from training interval $\bm\tau\in[4.0, 20.0]$ to testing interval $\bm\tau\in[1.0, 4.0)\cup(20.0, 23.0]$. 
All methods struggle in hard OOD tasks (Fig. \ref{fig_merged_dr}g left), but MPTS dominates in difficult cases, demonstrating strong adaptation. 
DRM exhibits high variability and weak generalization, even for easier tasks (Fig. \ref{fig_merged_dr}g right).

\subsection{MPTS facilitates efficient robust SFT}

In SFT, each labeled example in the dataset can be viewed as a task. 
Following MaPLe, we execute prompt tuning on ImageNet \citep{russakovsky2015imagenet} and conduct standard evaluation. 
To assess post-SFT robustness, we test on four OOD datasets, i.e., ImageNet-A \citep{hendrycks2021natural}, ImageNet-S \citep{wang2019learning}, ImageNet-R \citep{hendrycks2021many}, and ImageNet-V \citep{recht2019imagenet} for capturing diverse domain shifts.

Table \ref{table_prompt_tune_test} shows MPTS consistently outperforms baselines in average and CVaR accuracies on ID and OOD datasets. 
MPTS achieves 0.82–3.03\% higher $\text{CVaR}_{0.9}$, $\text{CVaR}_{0.7}$ and $\text{CVaR}_{0.5}$ scores over ERM, with greater OOD advantages than on ImageNet. 
On 4/5 datasets, DRM ranks second to MPTS in $\text{CVaR}_{\alpha}$ but matches ERM in average accuracy.
GDRM’s performance varies with $\alpha$, showing only marginal gains over ERM.
OHTM slightly boosts ERM’s $\text{CVaR}_{\alpha}$ and mean accuracy in most settings, while consuming more memory during tuning processes.
Still, DRM sacrifices both memory and computational efficiency for robustness in the bottom of Table \ref{table_prompt_tune_test}. 
While MPTS shares DRM’s optimization goal, its risk predictive module and larger-batch simulation enable better task exploration at minimal computational cost, yielding a more robust machine learner.

\begingroup
\setlength{\tabcolsep}{8pt}
\begin{small}
\begin{table}[!htbp]
   \begin{center}
    \caption{\textbf{Testing Classification Results after Prompt-Tuning on ImageNet and Efficiency Comparison (10 Runs).}
    We report testing $\text{CVaR}_{0.9}$, $\text{CVaR}_{0.7}$, $\text{CVaR}_{0.5}$ and average accuracies together with corresponding SEMs evaluated by the prompt-tuned machine learner on ID and OOD datasets. 
    Evaluation on OOD datasets corresponds to the domain generalization setting.
    The \textbf{best results} are in bold with the \underline{runner-up} underlined, and MPTS's performance gains over ERM $\Delta\uparrow$ are marked in blue.
    During prompt-tuning ImageNet, we report the memory cost and runtime relative to ERM for all methods.
    }
    \label{table_prompt_tune_test}
    \resizebox{0.83\columnwidth}{!}{
    \begin{tabular}{|c|c|c|c|c|c|cc|}
      \toprule %
      Dataset & Metrics & ERM & DRM & GDRM & OHTM & MPTS (Ours) & $\Delta\uparrow$ \\
      \toprule %
      ImageNet \citep{russakovsky2015imagenet} & $\text{CVaR}_{0.9}$ & 31.70$\pm${\scriptsize 0.12} & \underline{32.23$\pm${\scriptsize 0.15}} & 31.37$\pm${\scriptsize 0.18} & 31.53$\pm${\scriptsize 0.21} & \textbf{32.52$\pm${\scriptsize 0.16}} & +0.82\\
       (ID) & $\text{CVaR}_{0.7}$ & 42.87$\pm${\scriptsize 0.14} & \underline{44.10$\pm${\scriptsize 0.14}} & 42.97$\pm${\scriptsize 0.18} & 43.28$\pm${\scriptsize 0.18} & \textbf{44.28$\pm${\scriptsize 0.14}} & +1.41\\
      & $\text{CVaR}_{0.5}$ & 51.43$\pm${\scriptsize 0.10} & \underline{52.61$\pm${\scriptsize 0.11}} & 51.75$\pm${\scriptsize 0.19} & 51.96$\pm${\scriptsize 0.16} & \textbf{52.72$\pm${\scriptsize 0.15}} & +1.29\\
      & Avg & 70.80$\pm${\scriptsize 0.08} & 70.90$\pm${\scriptsize 0.10} & \underline{71.00$\pm${\scriptsize 0.08}} & 70.80$\pm${\scriptsize 0.10} & \textbf{71.20$\pm${\scriptsize 0.09}} & +0.40\\
      \midrule
      ImageNet-A \citep{hendrycks2021natural} & $\text{CVaR}_{0.9}$ & 15.35$\pm${\scriptsize 0.21} & 15.50$\pm${\scriptsize 0.25} & \underline{15.62$\pm${\scriptsize 0.22}} & 15.44$\pm${\scriptsize 0.25} & \textbf{18.38$\pm${\scriptsize 0.22}} & +3.03\\
       (OOD) & $\text{CVaR}_{0.7}$ & 22.82$\pm${\scriptsize 0.23} & 23.05$\pm${\scriptsize 0.21} & \underline{23.13$\pm${\scriptsize 0.19}} & 23.00$\pm${\scriptsize 0.23} & \textbf{24.05$\pm${\scriptsize 0.20}} & +1.23\\
      & $\text{CVaR}_{0.5}$ & 30.05$\pm${\scriptsize 0.21} & 29.53$\pm${\scriptsize 0.18} & \underline{30.25$\pm${\scriptsize 0.20}} & 29.96$\pm${\scriptsize 0.23} & \textbf{31.26$\pm${\scriptsize 0.20}} & +1.21\\
      & Avg & \underline{49.88$\pm${\scriptsize 0.18}} & 48.42$\pm${\scriptsize 0.25} & 49.53$\pm${\scriptsize 0.14} & 49.70$\pm${\scriptsize 0.25} & \textbf{51.10$\pm${\scriptsize 0.18}} & +1.22\\
      \midrule
      ImageNet-R \citep{hendrycks2021many} & $\text{CVaR}_{0.9}$ & 26.20$\pm${\scriptsize 0.11} & \underline{28.01$\pm${\scriptsize 0.13}} & 26.01$\pm${\scriptsize 0.10} & 27.87$\pm${\scriptsize 0.14} & \textbf{28.21$\pm${\scriptsize 0.12}} & +2.01\\
       (OOD) & $\text{CVaR}_{0.7}$ & 43.55$\pm${\scriptsize 0.12} & \underline{45.25$\pm${\scriptsize 0.11}} & 43.95$\pm${\scriptsize 0.11} & 44.89$\pm${\scriptsize 0.15} & \textbf{45.50$\pm${\scriptsize 0.10}} & +1.95\\
      &$\text{CVaR}_{0.5}$ & 56.70$\pm${\scriptsize 0.11} & \underline{58.25$\pm${\scriptsize 0.12}} & 57.36$\pm${\scriptsize 0.11} & 58.03$\pm${\scriptsize 0.12} & \textbf{58.77$\pm${\scriptsize 0.12}} & +2.07\\
      & Avg & 76.91$\pm${\scriptsize 0.08} & \underline{77.40$\pm${\scriptsize 0.10}} & 77.30$\pm${\scriptsize 0.10} & 77.12$\pm${\scriptsize 0.11} & \textbf{77.60$\pm${\scriptsize 0.09}} & +0.69\\
      \midrule
      ImageNet-S \citep{wang2019learning} & $\text{CVaR}_{0.9}$ & 12.24$\pm${\scriptsize 0.20} & \underline{13.18$\pm${\scriptsize 0.20}} & 12.20$\pm${\scriptsize 0.19} & 12.73$\pm${\scriptsize 0.19} & \textbf{13.61$\pm${\scriptsize 0.17}} & +1.37\\
       (OOD) & $\text{CVaR}_{0.7}$ & 20.01$\pm${\scriptsize 0.18} & \underline{21.10$\pm${\scriptsize 0.22}} & 20.44$\pm${\scriptsize 0.20} & 20.65$\pm${\scriptsize 0.25} & \textbf{21.46$\pm${\scriptsize 0.21}} & +1.45\\
      & $\text{CVaR}_{0.5}$ & 26.70$\pm${\scriptsize 0.20} & \underline{27.44$\pm${\scriptsize 0.20}} & 27.42$\pm${\scriptsize 0.25} & 27.40$\pm${\scriptsize 0.26} & \textbf{27.98$\pm${\scriptsize 0.20}} & +1.28\\
      & Avg & 48.85$\pm${\scriptsize 0.23} & 48.89$\pm${\scriptsize 0.21} & 48.91$\pm${\scriptsize 0.23} & \underline{49.12$\pm${\scriptsize 0.20}} & \textbf{49.62$\pm${\scriptsize 0.22}} & +0.77\\
      \midrule
      ImageNet-V \citep{recht2019imagenet} & $\text{CVaR}_{0.9}$ & 24.65$\pm${\scriptsize 0.27} & \underline{25.52$\pm${\scriptsize 0.20}} & 24.17$\pm${\scriptsize 0.18} & 24.82$\pm${\scriptsize 0.17} & \textbf{25.95$\pm${\scriptsize 0.23}} & +1.30\\
       (OOD) & $\text{CVaR}_{0.7}$ & 34.72$\pm${\scriptsize 0.20} & \underline{35.70$\pm${\scriptsize 0.17}} & 35.00$\pm${\scriptsize 0.23} & 35.32$\pm${\scriptsize 0.25} & \textbf{35.86$\pm${\scriptsize 0.16}} & +1.14\\
      & $\text{CVaR}_{0.5}$ & 43.49$\pm${\scriptsize 0.24} & \underline{44.20$\pm${\scriptsize 0.19}} & 43.60$\pm${\scriptsize 0.22} & 43.95$\pm${\scriptsize 0.18} & \textbf{44.60$\pm${\scriptsize 0.20}} & +1.11\\
      & Avg & 64.02$\pm${\scriptsize 0.22} & 63.87$\pm${\scriptsize 0.19} & \underline{64.10$\pm${\scriptsize 0.20}} & \underline{64.10$\pm${\scriptsize 0.17}} & \textbf{64.55$\pm${\scriptsize 0.16}} & +0.53\\
      \bottomrule
      Efficiency  &Runtime   & 1.00 & 1.10 & 1.00  & 1.09  &  1.03 &  \\
       &Memory  & 1.00 & 1.50 & 1.00 & 2.08 & 1.01 &    \\
      \bottomrule
    \end{tabular}
    }
  \end{center}
\end{table}    
\end{small}
\endgroup

\FloatBarrier
\section{Discussion}

Rapid adaptation to novel scenarios is a cornerstone of artificial general intelligence.
However, challenges such as safety, limited annotations, and computational constraints necessitate robust and efficient adaptation mechanisms. 
This study explores learn-to-adapt optimization via generative modeling and introduces MPTS, a versatile framework for robust active task sampling.

Experiments demonstrate the feasibility of predicting optimization outcomes for active task selection.
Meanwhile, MPTS enhances adaptation robustness across diverse scenarios in an efficient manner.
These results highlight MPTS’s potential to scale $\text{CVaR}_{\alpha}$ principles for foundation model development and large-scale decision-making, without additional learning resources.

\section{Methods}\label{method_sec}

In alignment with the realistic necessities, this work focuses on robust adaptation while securing learning efficiency, such as circumventing partial expensive evaluation.
Such a purpose facilitates the birth of MPTS.
The framework is also agnostic to adaptation learning methods; hence, we leave out zero-shot learning, few-shot learning, and SFT details.

In Fig. \ref{fig_overall_mpts}a, several roles are involved in the optimization: (1) the adaptive machine learner, e.g., foundation models or generalist policies, learns to adapt given some optimizers; (2) the risk predictive model as a critic evaluates and forecasts the task-specific adaptation risk; (3) the task sampler as an actor works for screening the task subset for next iteration.
These components participate in episodic learning until convergence.

Technically, this work recasts task episodic learning to sequence generation and presents MPTS as the task sampling strategy to balance exploration and exploitation.
At first, we introduce the foundation of RPMs for ranking task difficulty.
To reconcile theory and practice, we introduce a tractable optimization approach to enable functional posterior inference towards adaptation risk.
Then, we devise the acquisition function informed by the captured risk landscapes. 
Finally, an understanding concerning the optimization pipeline is attached to conclude the this part.

\subsection{Theoretical Feasibility of Constructing RPMs}

\paragraph{Predictive Foundation.}
MPTS in Definition \ref{definition_mpts} leverages the risk history and generalization results under $\bm\theta_t$ to approximately score the difficulty of task samples for the $(t+1)$-th iteration. 
An unbiased Monte Carlo estimate of $\text{CVaR}_{1-\hat{\mathcal{B}}/\mathcal{B}}$ would require fully evaluation of the $(t+1)$-th adaptation risks across $\hat{\mathcal{B}}$ tasks in order to identify the Top-$\mathcal{B}$ subset, which is computationally prohibitive. 
Instead, MPTS circumvents this costly evaluation by utilizing the risk batch $\{\ell_{t,i}\}_{i=1}^{\mathcal{B}}$ from iteration $t$, together with historical information $H_{1:t-1}$, to train a generative model to rank difficulty of $\hat{\mathcal{B}}$ tasks for the $(t+1)$-th subset selection. 
This design inevitably introduces the on-step lagging bias (see Theorem \ref{theorem_prob_score} and Lemma \ref{lemma_mis_rank_quant}-\ref{lemma_rank_preserve_bound}) while avoiding direct access to the expensive evaluations of $\{\ell_{t+1,i}\}_{i=1}^{\hat{\mathcal{B}}}$.

We begin by introducing Assumptions \ref{assum_lip}/\ref{assum_bound}/\ref{assum_subg}/\ref{assum_mac}, which characterize the smoothness, boundedness and margin conditions essential to the optimization framework. 
Note that Assumption \ref{assum_mac} is not to impose unrealistic smoothness, but to formalize a ranking robustness condition, i.e., small changes in model parameters should not induce arbitrary flips in task difficulty ordering. 
This is natural in episodic learning settings where task losses vary smoothly with the learner parameters, and it is empirically supported by the consistently high rank correlations observed in the Fig. \ref{fig_meta_sin}g of the Results Section.
Specifically, under a fixed machine learner $\bm\theta$, it is reasonable to expect that similar tasks, represented by $\bm\tau$, will exhibit sufficiently close adaptation risk values.

\begin{assumption}[Lipschitz Continuity]\label{assum_lip}
    We assume the adaptation risk function $\ell(\cdot;\bm\theta)$ reserves the Lipschitz continuity w.r.t. $\bm\theta$ and $\bm\tau$, i.e.,
    \begin{equation}\label{eq_lipschitz_l}
        \begin{split}
            |\ell(\mathcal{D}_{\tau}^{Q},\mathcal{D}_{\tau}^{S};\bm\theta)-\ell(\mathcal{D}_{\tau}^{Q},\mathcal{D}_{\tau}^{S};\bm\theta^{\prime})|\leq\beta_{1}||\bm\theta-\bm\theta^{\prime}||\quad\text{and}\quad
            |\ell(\mathcal{D}_{\tau}^{Q},\mathcal{D}_{\tau}^{S};\bm\theta)-\ell(\mathcal{D}_{\tau^{\prime}}^{Q},\mathcal{D}_{\tau^{\prime}}^{S};\bm\theta)|\leq\beta_{2}||\bm\tau-\bm\tau^{\prime}||,
        \end{split}
    \end{equation}
    where $\forall\{\bm\theta,\bm\theta^{\prime}\}\in\bm\Theta$ and $\forall\{\bm\tau,\bm\tau^{\prime}\}\in\mathcal{T}$ with Lipschitz constants $\beta_1$ and $\beta_2$.
\end{assumption}

\begin{assumption}[Bounded Sample Gradient]\label{assum_bound}
    We assume the norm of the adaptation risk function's gradient $\nabla\ell(\cdot;\bm\theta_{t})$ is bounded:
    \begin{equation}
        \sup_{\tau\in\mathcal{T}}||\nabla_{\bm\theta}\ell(\mathcal{D}_{\tau}^{Q},\mathcal{D}_{\tau}^{S};\bm\theta_{t})||_{2}<G_{t}\
        \text{and}\
        \sup_{\tau\in\mathcal{T},t\in\mathbb{N}_{+}}||\nabla_{\bm\theta}\ell(\mathcal{D}_{\tau}^{Q},\mathcal{D}_{\tau}^{S};\bm\theta_{t})||_{2}<G,
    \end{equation}
    where $G_t$ is a positive constant and $G$ is a overall bound.
\end{assumption}

\begin{assumption}[Sub-Gaussian Stochastic Gradient]\label{assum_subg}
    The stochastic gradient $\Tilde{\bm g}:=\bm g+\bm\epsilon$ for the machine learner's adaptation at $t$-th iteration is $\sigma$-sub-Gaussian, which means:
    \begin{equation}
        \begin{split}
            \mathbb{E}\left[\exp{(\eta\bm v^{T}\bm\epsilon)}\right]\leq\exp{\left(\frac{\eta^{2}\sigma^{2}||\bm v||_{2}^{2}}{2}\right)}
            \quad
            \forall
            \eta\in\mathbb{R}\
            \text{and}\
            \bm v\in\mathbb{R}^{d},
        \end{split}
    \end{equation}
    where $\mathbb{E}[\Tilde{\bm g}]=\bm g$, $\mathbb{E}[||\Tilde{\bm g}-\bm g||_{2}^{2}]\leq\sigma^2$ and $\sigma\in\mathbb{R}_{+}$.
\end{assumption}

\begin{assumption}[Margin Anti-Concentration]\label{assum_mac}
    We assume $\Delta_{ij}(\bm\theta_{t})=\ell(\mathcal{D}_{\tau_i}^{Q},\mathcal{D}_{\tau_i}^{S};\bm\theta_{t})-\ell(\mathcal{D}_{\tau_j}^{Q},\mathcal{D}_{\tau_j}^{S};\bm\theta_{t})$ has a density uniformly bounded by $\rho$ in a neighborhood of zero.
    That is, $\forall \epsilon>0$, the probability inequality holds:
    \begin{equation}
        \mathbb{P}(|\Delta_{ij}(\bm\theta_{t})|\leq\epsilon)\leq\rho\epsilon.
    \end{equation}
\end{assumption}

Under the aforementioned assumptions, we derive Theorem \ref{theorem_prob_score}. 
Specifically, we define a random variable $\Delta_{ij}(\bm\theta_{t})$ as the sign of the adaptation risk difference and analyze its evolution following gradient updates across a population. 
Our theoretical analysis demonstrates that, under a sufficiently small learning rate for the machine learner update, a significant proportion of these sign variables remain largely unchanged in a probabilistic sense. 
This result establishes a rigorous foundation for evaluating relative task difficulty on $\bm\theta_{t+1}$ based on posterior inference outcomes derived from $\bm\theta_{t}$ and further guides amortizing the sample average Monte Carlo of $\text{CVaR}_{\alpha}$ optimization objective (see Fig. \ref{fig_overall_mpts}a-b).

\begin{theorem}[Provably Approximately Invariant Task Difficulties]\label{theorem_prob_score}
    Given arbitrary $K$ data points $\{(\bm\tau_i,\ell(\mathcal{D}_{\tau_i}^{Q},\mathcal{D}_{\tau_i}^{S};\bm\theta_t)\}_{i=1}^{K}$, the adaptation gradient $\nabla_{\bm\theta}\mathcal{\bm L}(\bm\theta_t)$ as a $\sigma$-sub-Gaussian random variable and $\bm\theta_{t+1}=\bm\theta_{t}-\eta_{t}\nabla_{\bm\theta}\mathcal{\bm L}(\bm\theta_t)$, we denote the relative difficulty via the difference $\Delta_{ij}(\bm\theta_{t+1})=\ell(\mathcal{D}_{\tau_i}^{Q},\mathcal{D}_{\tau_i}^{S};\bm\theta_{t+1})-\ell(\mathcal{D}_{\tau_j}^{Q},\mathcal{D}_{\tau_j}^{S};\bm\theta_{t+1})$ and $\Delta_{ij}(\bm\theta_{t})=\ell(\mathcal{D}_{\tau_i}^{Q},\mathcal{D}_{\tau_i}^{S};\bm\theta_{t})-\ell(\mathcal{D}_{\tau_j}^{Q},\mathcal{D}_{\tau_j}^{S};\bm\theta_{t})$ between $t$-th and $(t+1)$-th iterations, and the gradient difference as $\bm v_{ij}:=\nabla_{\bm\theta}\ell(\mathcal{D}_{\tau_i}^{Q},\mathcal{D}_{\tau_i}^{S};\bm\theta_{t})-\nabla_{\bm\theta}\ell(\mathcal{D}_{\tau_j}^{Q},\mathcal{D}_{\tau_j}^{S};\bm\theta_{t})$.

Under Assumption \ref{assum_lip}/\ref{assum_bound}/\ref{assum_subg}, the set of rank-preserving variable $E_{ij}:=\mathds{1}\left[\text{sign}(\Delta_{ij}(\bm\theta_{t+1}))=\text{sign}(\Delta_{ij}(\bm\theta_{t}))\right]$ satisfies the probability inequality:
$$\mathbb{P}(\bigcap_{i<j}E_{ij})\geq 1-\xi,$$
when $\eta_{t}\leq\frac{\delta_t}{2G_{t}M_{t}+\sqrt{8\sigma^{2}G_{t}^{2}\ln\left(\frac{K(K-1)}{2\xi}\right)}}$ with $G_t$ in Assumption \ref{assum_bound}, $\delta_{t}:=\min_{i\neq j}|\ell(\mathcal{D}_{\tau_i}^{Q},\mathcal{D}_{\tau_i}^{S};\bm\theta_t)-\ell(\mathcal{D}_{\tau_j}^{Q},\mathcal{D}_{\tau_j}^{S};\bm\theta_t)|\in\mathbb{R}_{+}$, the stochastic gradient norm $M_{t}:=||\nabla_{\bm\theta}\mathcal{\bm L}(\bm\theta_t)||_{2}$.
\end{theorem}

\begin{lemma}[Misranked Subset Quantity]\label{lemma_mis_rank_quant}
    In the presence of adaptation risk value, let $N_{f}$ be the number of order flipped cross-pairs between the ground-truth Top-$\mathcal{B}$ subset $\bar{\mathcal{T}}_{t}^{\mathcal{B}}$ and the remainder of the candidate tasks $\mathcal{T}_{t}^{C}:=\mathcal{T}^{\hat{\mathcal{B}}}_{t+1}\setminus\bar{\mathcal{T}}_{t}^{\mathcal{B}}$ when the machine learner's parameter changes from $\bm\theta_t$ to $\bm\theta_{t+1}$.
    We denote the number of tasks that change Top-$\mathcal{B}$ membership by $m_{t+1}=|\bar{\mathcal{T}}_{t}^{\mathcal{B}}\bigtriangleup\mathcal{T}_{t+1}^{\mathcal{B}}|$, where $\mathcal{T}_{t+1}^{\mathcal{B}}$ is the ground-truth Top-$\mathcal{B}$ task subset under $\bm\theta_{t+1}$.
    Then the inequality holds: $m_{t+1}\leq 2N_{f}$.
\end{lemma}

\begin{lemma}[Rank-Preserving Bound in Expectation]\label{lemma_rank_preserve_bound}
    With the risk difference notation $\Delta_{ij}(\bm\theta_{t})=\ell(\mathcal{D}_{\tau_i}^{Q},\mathcal{D}_{\tau_i}^{S};\bm\theta_{t})-\ell(\mathcal{D}_{\tau_j}^{Q},\mathcal{D}_{\tau_j}^{S};\bm\theta_{t})$ and $\Delta_{ij}(\bm\theta_{t+1})=\ell(\mathcal{D}_{\tau_i}^{Q},\mathcal{D}_{\tau_i}^{S};\bm\theta_{t+1})-\ell(\mathcal{D}_{\tau_j}^{Q},\mathcal{D}_{\tau_j}^{S};\bm\theta_{t+1})$, when the $(i,j)$ cross-pair flips its order from $\bm\theta_t$ to $\bm\theta_{t+1}$, we conclude that $|\Delta_{ij}(\bm\theta_t)|\leq|\Delta_{ij}(\bm\theta_{t+1})-\Delta_{ij}(\bm\theta_{t})|$.
\end{lemma}

Note that the rank of adaptation difficulty in a fixed task set might flip due to the model update.
Lemma \ref{lemma_mis_rank_quant} and \ref{lemma_rank_preserve_bound} provides a bound for the theoretically misranked tasks in the next-iteration selected subset and the adaptation risk difference over iteration.
Starting from the updated model parameter $\bm\theta_{t+1}$, the subsequent one-step gradient update can be written as
\begin{subequations}\label{eq_main_grad_decomp}
    \begin{align}
        \bm\theta_{t+2}=\bm\theta_{t+1}-\eta_{t+1}\nabla_{\bm\theta}\mathcal{L}(\bm\theta_{t+1})\\
        \nabla_{\bm\theta}\mathcal{L}(\bm\theta_{t+1})=\bar{\bm g}_{t+1}+\Delta\bm g_{t+1},
    \end{align}
\end{subequations}
where the perturbed gradient $\nabla_{\bm\theta}\mathcal{L}(\bm\theta_{t+1})$ reflects the average task gradient after rank-flipping.
We estimate the unbiased average gradient $\bar{\bm g}_{t+1}$ from the ground-truth Top-$\mathcal{B}$ subset under $\bm\theta_{t+1}$, i.e., $\mathcal{T}_{t+1}^{\mathcal{B}}$.
The result of rank-flipping tasks after model update contributes to the gradient difference as 
\[\Delta\bm g_{t+1}=\frac{2}{m_{t+1}}\left(\sum_{\tau\in\bar{\mathcal{T}}_{t}^{\mathcal{B}}\setminus\mathcal{T}_{t+1}^{\mathcal{B}}}\ell(\mathcal{D}_{\tau}^{Q},\mathcal{D}_{\tau}^{S};\bm\theta_{t+1})-\sum_{\tau\in\mathcal{T}_{t+1}^{\mathcal{B}}\setminus\bar{\mathcal{T}}_{t}^{\mathcal{B}}}\ell(\mathcal{D}_{\tau}^{Q},\mathcal{D}_{\tau}^{S};\bm\theta_{t+1})\right)\]
.
These terms serve the convergence analysis $\forall t\in\mathbb{N}_{+}$ in the following contents.

\begin{lemma}[Misranking Acute-Angle]\label{lemma_misrank_acute_angle}
    Let $\bar{\bm g}_{t}$ and $\Delta\bm g_{t}$ respectively denote the unbiased Monte Carlo estimate of CVaR and the gradient difference between it and the biased gradient $\nabla_{\bm\theta}\mathcal{L}(\bm\theta_t)$.
    For a given $c\in(0,1)$ and arbitrary small $\epsilon_*>0$, there exists $T$ such that $\forall t\geq T$, the following dichotomy holds: either (i) $||\bar{\bm g}_{t}||_{2}\leq\epsilon_*$, or (ii) $||\Delta\bm g_{t}||_{2}\leq c||\bar{\bm g}_{t}||_{2}$.
\end{lemma}

\begin{theorem}[Convergence with Diminishing Rank Flipping]\label{theorem_converge}
    Suppose there exists $c\in(0,1)$ and $T\geq 0$ such that $\forall t\geq T$, the inequality holds: $||\Delta\bm g_{t}||_{2}\leq c||\bar{\bm g}_{t}||_{2}$.
    Given the Assumptions \ref{assum_lip}/\ref{assum_bound}/\ref{assum_subg}/\ref{assum_mac}, and the appropriate construction of the step sizes $\{\eta_t\}$ from Theorem \ref{theorem_prob_score}, we conclude that:
    $\lim_{t\to\infty}\mathbb{E}\left[||\nabla_{\bm\theta}\mathcal{L}(\bm\theta_t)||_2\right]=0$.
    Hence, the iteration converges to first-order stationary points in expectation.
\end{theorem}

The key challenge addressed by Theorem \ref{theorem_converge} is that MPTS fundamentally alters the optimization trajectory by replacing exact task evaluation, as in standard CVaR-style optimization, with the predictive amortized task selection. 
Specifically, the amortized evaluation strategy injects systematic bias and additional stochasticity into the gradient through (i) imperfect RPMs and (ii) subset selection based on prediction rather than exact losses. 
In this setting, convergence is no longer guaranteed a priori, even if the underlying learner uses gradient descent, due to oscillation or divergence caused by persistent task misranking.
However, with {Lemmas} \ref{lemma_mis_rank_quant}/\ref{lemma_rank_preserve_bound}/\ref{lemma_misrank_acute_angle} and Assumptions \ref{assum_lip}/\ref{assum_bound}/\ref{assum_subg}/\ref{assum_mac}, we rigorously demonstrate that a learning rate schedule in the way of Theorem \ref{theorem_prob_score} can secure the convergence of MPTS.

In other words, Theorem \ref{theorem_converge} provides a convergence guarantee of using noisy Top-$\mathcal{B}$ task subset for robust optimization.
It is worth noting that, provided the RPM possesses sufficient expressiveness, the accumulated optimization history can theoretically drive down the generalization error arising from function approximation, as supported by statistical learning theory.
We leave more detailed discussions and MPTS relevant proof in Appendices~\ref{sec_supp_formulation} and~\ref{sec_supp_convergence}.

\paragraph{Diminishing Temporal Bias in the RPM.}
{Theorems} \ref{theorem_prob_score}/\ref{theorem_converge} have clarified the predictive foundation of MPTS and illustrated the theoretical convergence in stochastic optimization.
Intuitively, the rank flipping can be viewed as the noise injected into the gradient of the Monte Carlo estimate of $\text{CVaR}_{\alpha}$.
In implementation, MPTS uses the function approximation capability of neural networks and introduces the RPM to utilize the optimization history $H_{1:t}$, generalizing from previously sampled and evaluated tasks to a new candidate pool $\mathcal{T}_{t+1}^{\hat{\mathcal{B}}}$.
That is, the rank flipping can also be attributed to the error of the function approximation in the RPM.
However, this can be well alleviated if seeking an optimal RPM.
Here, we can analyze the temporal bias on an arbitrary task $\tau$ through the adaptation risk approximation lens:
\begin{subequations}
    \begin{align}
    \text{Temporal Bias:}\quad
    \left|\mathbb{E}_{p(\ell\vert\bm\tau, H_{1:t})}[\ell] - \ell(\mathcal{D}_{\tau}^{Q},\mathcal{D}_{\tau}^{S}; \bm\theta_{t+1})\right|\\
    \leq \left|\mathbb{E}_{p(\ell\vert\bm\tau, H_{1:t})}[\ell] - \ell(\mathcal{D}_{\tau}^{Q},\mathcal{D}_{\tau}^{S}; \bm\theta_t)\right| + \left|\ell(\mathcal{D}_{\tau}^{Q},\mathcal{D}_{\tau}^{S}; \bm\theta_t) - \ell(\mathcal{D}_{\tau}^{Q},\mathcal{D}_{\tau}^{S}; \bm\theta_{t+1})\right| \\
    \leq \epsilon_t + \beta_1 \|\bm\theta_{t+1} - \bm\theta_t\|_{2} 
    = \epsilon_t^{\text{approx error}} + \beta_1 \eta_t \|\nabla_{\bm\theta}\mathcal{L}(\bm\theta_t)\|_{2}.
    \end{align}
\end{subequations}
Here $\epsilon_t^{\text{approx error}}$ denotes the RPM's approximation error due to the function approximation of the RPM, while $\beta_1$ is the Lipschitz constant from the Assumption \ref{assum_lip}.
Note that if the RPM is expressive enough, the generalization error from function approximation can be theoretically well decreased with the accumulation of the optimization history based on the statistical learning theory.
Hence, the limit of the right-hand side term shrinks to tiny values over iterations.

The specific analysis of multiple approximation biases depends on the design of the RPM.
This work provides the VAE-like model in implementation, and we believe there exist more optimal candidates for the choice of RPMs.
And this can be a promising research direction in the future.
Compared with subset selection based robust optimization method like $\text{CVaR}_{\alpha}$, MPTS amortizes part of computation or agent-environment interaction intensive task evaluation; however, it also introduces approximation bias from the RPM and irreducible bias due to rank flipping when the model parameters change from $\bm\theta_t$ to $\bm\theta_{t+1}$. 

One notable contribution in MPTS is introducing the RPM as a predictor to surrogate the task evaluation in the next iteration.
This means RPMs can also be combined with other acquisition criteria, such as the medium-difficulty task prioritized sampling.
In the context of RL finetuning for large reasoning models, a variant of MPTS called MoPPS \citep{qu2025can} demonstrates that curriculum task selection with a lightweight predictive model is more efficient and effective than subset selection methods of exact evaluation.

\subsection{Generative Modeling Risk Functions and Posterior Inference}\label{subsec:risk_post}

Here, we design the sampling strategy through the lens of risk landscapes and pay more attention to datasets of learning optimization outcome $\{H_{t}\}_{t=1}^{T}$.
To characterize the adaptation risk during batch optimization, we introduce the latent variable $\bm z_t$ to summarize episodic information and present a versatile deep generative model as:
\begin{equation}\label{eq_risk_generation}
    \begin{split}
        p(H_{0:T},\bm z_{0:T}\vert\bm\theta_{0:T})
        =p(\bm z_{0})\prod_{t=0}^{T}p_{\bm\psi}(H_{t}\vert\bm z_{t};\bm\theta_{t})\prod_{t=0}^{T-1}p(\bm z_{t+1}\vert\bm z_{t}).
    \end{split}
\end{equation}

Within a Bayesian framework, we approximate the underlying function distribution with the latent variable, and the posterior $p(\bm z_{t}\vert H_{t})$ summarizes the historical risk information and accounts for uncertainty in distributions.
The following writes the form of $p(\bm z_{t}\vert H_{t})$ according to the Bayes rule \citep{stigler1982thomas}:
\begin{equation}
    p(\bm z_{t}\vert H_{t})=\frac{p(H_{t}\vert\bm z_{t})p(\bm z_{t}\vert H_{1:t-1})}{\int p(H_{t}\vert\bm z_{t})p(\bm z_{t}\vert H_{1:t-1})d\bm z_{t}},
\end{equation}
where $p(\bm z_{t}\vert H_{1:t-1})$ encodes the past evaluation results as the conditional prior.
Moreover, $p(H_{t}\vert\bm z_{t})$ conveys the likelihood of producing observations of the task batch risk values in the $t$-th iteration.
Notably, the exact computation \textit{w.r.t.} the posterior is intractable due to the complicated integral in the denominator.

\paragraph{Generative Process.}
As illustrated in Fig. \ref{fig_basic_setup}c, risk values of the task batch $\ell$ are correlated with the machine learner's parameters $\bm\theta$.
In specific, the factorization of the sequential optimization relevant variables arrives at:
\begin{equation}\label{cond_factorization}
    \begin{split}
        p_{\bm\psi}(H_{t}\vert H_{1:t-1})=\int p_{\bm\psi}(H_{t}\vert \bm z_{t})p(\bm z_{t}\vert H_{1:t-1})d\bm z_{t}
        =\int\Big[\prod_{i=1}^{\mathcal{B}}p_{\bm\psi}(\ell_{t,i}\vert\bm\tau_{t,i},\bm z_{t};\bm\theta_t)\Big]p(\bm z_{t}\vert H_{1:t-1})d\bm z_{t},
    \end{split}
\end{equation}
where $\bm z_{t}$ in the probabilistic graphical model constitutes the distribution over risk functions (For the sake of simplicity, we skip over other variables less relevant to our learning purposes).
Here, we assume the conditional independence between task-specific risk values given $\bm z$ and the machine learner's parameter $\bm\theta$ in Eq. (\ref{cond_factorization}).
And the primary optimization objective is to $\max_{\bm\psi\in\bm\Psi}\ln p_{\bm\psi}(H_{t}\vert H_{1:t-1})$ for the optimization outcome prediction.

\paragraph{Inference Process.}
The manner of episodic training, where the task batch and its evaluation arrive sequentially, inspires us to predict adaptation risk values online to actively sample tasks in a batch.
However, the exact inference \textit{w.r.t.} $p(\bm z_{t}\vert H_{t})$ is infeasible as there is no structural information regarding posteriors.
In each iteration, the risk function distribution relies on the updated machine learner $\bm\theta$; hence, such non-stationarity in the risk function distributions prompts us to involve the streaming VI \citep{broderick2013streaming,nguyen2017variational} to derive the approximate posterior.

To do so, we handle the streaming task batches and update the posterior in a recursive way:
\begin{equation}
    \begin{split}
        p(\bm z_{t}\vert H_{t})\propto p(H_{t}\vert\bm z_{t})p(\bm z_{t}\vert H_{1:t-1})
    \end{split}
\end{equation}
where $p(\bm z_{t}\vert H_{1:t-1})$ represents the conditional prior using the last time updated posterior as the proxy.
The role of the estimated functional posterior $p(\bm z_{t}\vert H_{t})$ is to provide uncertainty-aware prediction and serves the task sampling strategy design, which will be detailed in Section \ref{subsec:sampling strategy design}.

As a result, we can formulate the evidence lower bound (ELBO) as a tractable optimization objective in Eq. (\ref{stream_vi_obj}) from approximate inference.
\begin{equation}\label{stream_vi_obj}
    \begin{split}
\begin{small}
        \max_{\bm\psi\in\bm\Psi,\bm\phi\in\bm\Phi}\hat{\mathcal{G}}_{\text{ELBO}}(\bm\psi,\bm\phi):=\mathbb{E}_{q_{\bm\phi}(\bm z_{t}\vert H_{t})}\left[\sum_{i=1}^{\mathcal{B}}\ln p_{\bm\psi}(\ell_{t,i}\vert\bm\tau_{t,i},\bm z_{t})\right]-D_{KL}\Big[q_{\bm\phi}(\bm z_{t}\vert H_{t})\parallel p(\bm z_{t}\vert H_{1:t-1})\Big]
\end{small}
    \end{split}    
\end{equation}
For implementation convenience, we adopt the parameterized Gaussian distribution with diagonal covariance matrices as variational distributions similar to vanilla VAEs \citep{kingma2013auto,rezende2014stochastic} and neural processes (NPs) \citep{garnelo2018neural}.
In other words, these distribution parameters are approximated with neural networks, e.g., $q_{\bm\phi}(\bm z_{t}\vert H_{t})=\mathcal{N}(\bm z_{t};\mu_{\bm\phi}(H_{t}),\Sigma_{\bm\phi}(H_{t}))$, and the reparameterization trick \citep{kingma2013auto} is used for stochastic gradient estimate.

\begin{definition}[Permutation Invariant Function]\label{def_pif}
    With an $n$-element permutation group $\mathcal{S}_{n}$, the operator $g\in\mathcal{S}_{n}$ maps the order set to itself:
    \begin{equation}
        g:[1,2,\dots,n]\mapsto[g_1,g_2,\dots,g_n].
    \end{equation}
    Then the function $\Phi$ is called permutation invariant if for any set of data points $\bm x_{1},\dots,\bm x_{n}$, the following condition holds:
    \begin{equation}
        \Phi(g\circ[\bm x_{1},\dots,\bm x_{n}])=\Phi([\bm x_{g_1},\dots,\bm x_{g_n}])=\Phi([\bm x_{1},\dots,\bm x_{n}])\quad\forall g\in\mathcal{S}_{n}.
    \end{equation}
\end{definition}

As for the neural architecture, we employ the DeepSet encoding module \citep{zaheer2017deep} to process the set dataset $H_{t}$, which corresponds to the permutation invariant function family in Definition \ref{def_pif}.
Also, in the context of streaming VI, $q_{\bm\phi}(\bm z_{t}\vert H_{t-1})$ mostly works as the proxy for the conditional prior as default.
Consequently, we can modify the exact ELBO in Eq. (\ref{stream_vi_obj}) and further translate the practical optimization process with the Lagrange multiplier $\beta$ into:
\begin{subequations}\label{eq_approx_elbo}
    \begin{align}
        \max_{\bm\psi\in\bm\Psi,\bm\phi\in\bm\Phi}\mathbb{E}_{q_{\bm\phi}(\bm z_{t}\vert H_{t})}\left[\sum_{i=1}^{\mathcal{B}}\ln p_{\bm\psi}(\ell_{t,i}\vert\bm\tau_{t,i},\bm z_{t})\right]\quad\text{s.t.}\quad
        D_{KL}\Big[q_{\bm\phi}(\bm z_{t}\vert H_{t})\parallel q_{\bar{\bm\phi}}(\bm z_{t}\vert H_{t-1})\Big]\leq\epsilon\Leftrightarrow\\
        \max_{\bm\psi\in\bm\Psi,\bm\phi\in\bm\Phi}\mathcal{G}_{\text{ELBO}}(\bm\psi,\bm\phi):=\mathbb{E}_{q_{\bm\phi}(\bm z_{t}\vert H_{t})}\left[\sum_{i=1}^{\mathcal{B}}\ln p_{\bm\psi}(\ell_{t,i}\vert\bm\tau_{t,i},\bm z_{t})\right]-\beta D_{KL}\Big[q_{\bm\phi}(\bm z_{t}\vert H_{t})\parallel q_{\bar{\bm\phi}}(\bm z_{t}\vert H_{t-1})\Big],
    \end{align}
\end{subequations}
where $\bar{\bm\phi}$ indicates no gradients computed through $\bm\phi$ in the term, and $\{\beta\in\mathbb{R}_{+},\epsilon\in\mathbb{R}_{+}\}$ constrains the machine learner's parameter search in next iteration.

\subsection{Task Sampling Strategy Design}\label{subsec:sampling strategy design}

In robust active task sampling, existing strategies evaluate task batches to rank their difficulties in adaptation and then prioritize challenging subsets for optimization \citep{rajeswaran2022epopt,evans2023bad,greenberg2024train,wang2024simple,lv2024theoretical}.
Besides the expensive evaluation cost, these strategies are weak in the efficient exploration of the task space.

As Theorem \ref{theorem_prob_score} has established the theoretical foundation of approximately rank task difficulty, this necessitates the development of the RPM from cumulated risk histories.
With the model predictive results as amortized evaluation, specific rules can be incorporated into the acquisition function for active sampling.
Meanwhile, it is fascinating for the RPM to evaluate the machine learner's adaptation to arbitrarily many tasks with minimal computational cost.
Hence, we can easily enlarge the pseudo batch size $\hat{\mathcal{B}}$ for more selection candidates and exploit the epistemic uncertainty from the RPM, encouraging more exploration in the task space.

\paragraph{Evaluating Adaptation Performance through Stochastic Forward Passes.}
The RPM and estimated functional posteriors in Eq. (\ref{cond_factorization})/(\ref{stream_vi_obj}) work as tools for the active selection of the task batch.
Specifically, the predictive distribution can be depicted as:
\begin{equation}
    \begin{split}
        p_{\bm\psi}(\ell\vert\bm\tau,H_{1:t})=\int p_{\bm\psi}(\ell\vert\bm\tau,\bm z_{t})p(\bm z_{t}\vert H_{1:t})d\bm z_{t}
        \overset{\triangle}{=}
         \int p_{\bm\psi}(\ell\vert\bm\tau,\bm z_{t})q_{\bm\phi}(\bm z_{t}\vert H_{t})d\bm z_{t}
         \\
         \approx
         \frac{1}{K}\sum_{k=1}^{K}p_{\bm\psi}(\ell\vert\bm\tau,\bm z_{t}^{(k)}),
         \
         \text{with}
         \
         \bm z_{t}^{(k)}\sim q_{\bm\phi}(\bm z_{t}\vert H_{t})
         \
         \forall\bm\tau\sim p(\tau).
    \end{split}
\end{equation}
The above predictive distribution $p_{\bm\psi}(\ell\vert\bm\tau,H_{1:t})$ benefits from the Bayesian modeling and provides a tractable way to roughly assess difficulties of tasks throughout the whole task space.

\paragraph{Rank-Flitering the Next Task Batch to Episodically Train.}
After obtaining $p_{\bm\psi}(\ell\vert\bm\tau,H_{1:t})$, we draw up a batch sampling strategy on the basis of its quantified statistics.
The criteria resembles the acquisition function in classical Bayesian optimization (BO), which includes a collection of available evaluation principles, such as expected improvement \citep{mockus1978application}, output information theoretical index \citep{ru2018fast} or UCB \citep{auer2002finite}.

However, it is also necessary to clarify that the search space is on the sequential task batch instead of machine learners' parameters, which differs from the ultimate purpose in BO.
Central to our approach is the principle of optimism in the face of uncertainty \citep{auer2002using}. 
We consider the difficult task's prioritization for robustness and the epistemic uncertainty as pivotal elements in developing acquisition functions. 
The grounds behind this idea are that (i) the subset with the worst performance deserves extra attention in optimization for adaptation robustness, and (ii) task regions with high predictive uncertainty tend to be underexplored in the last few iterations. 

As a result, we present the acquisition function built on the UCB principle \citep{auer2002finite}:
\begin{equation}\label{eq_acq}
    \begin{split}
        \mathcal{A}(\mathcal{T}^{\mathcal{B}};\bm\phi,\bm\psi)=\sum_{i=1}^{\mathcal{B}}a(\bm\tau_i)
        =\sum_{i=1}^{\mathcal{B}}\gamma_0 m(\ell_i)+\gamma_1\sigma(\ell_i),
        \
        \text{where}
        \
        \bm\tau_i\sim p(\tau)
        \\
        \text{with}
        \
        m(\ell_i)=\mathbb{E}_{q_{\bm\phi}(\bm z_{t}\vert H_{t})}\Big[p_{\bm\psi}(\ell\vert\bm\tau_i,\bm z_{t})\Big]
        \
        \text{and}
        \
        \sigma(\ell_i)=\mathbb{V}_{q_{\bm\phi}(\bm z_{t}\vert H_{t})}^{\frac{1}{2}}\Big[p_{\bm\psi}(\ell\vert\bm\tau_i,\bm z_{t})\Big],
    \end{split}
\end{equation}
where $m(\ell_i)$ and $\sigma(\ell_i)$ are, respectively, the adaptation risk mean and standard deviations, which can be estimated from multiple stochastic forward passes $\bm z_t\sim p(\bm z_t\vert H_{1:t})$ and $\ell\sim p_{\bm\psi}(\ell\vert\bm\tau_{i},\bm z_t)$ using the risk generative model.
And $\{\gamma_0,\gamma_1\}$ are hyperparameters to balance considerations.

Then, the Simulate-Rank-Filter operation in Eq. (\ref{eq_mpts_workflows})c arrives at the task batch for $(t+1)$-th iteration, i.e.,
$\mathcal{T}_{t+1}^{\mathcal{B}}=\arg\max_{\mathcal{T}^{\mathcal{B}}\subseteq\mathcal{T}^{\hat{\mathcal{B}}}_{t+1}:|\mathcal{T}^{\mathcal{B}}|=\mathcal{B}}\mathcal{A}(\mathcal{T}^{\mathcal{B}};\bm\phi,\bm\psi)$.
This characterizes the step of the active subset selection from $\mathcal{T}^{\hat{\mathcal{B}}}_{t+1}$, the randomly sampled identifier candidate set with $|\mathcal{T}^{\hat{\mathcal{B}}}_{t+1}|=\hat{\mathcal{B}}$.
In an implementation, we still perform random sampling from $p(\tau)$ and forecast the task-wise acquisition score $a(\cdot)$ from the RPM. 
Candidates in Top-$\mathcal{B}$ acquisition scores are screened to formulate the task batch $\mathcal{T}_{t+1}^{\mathcal{B}}$ for episodic optimization, as illustrated in Fig. \ref{fig_overall_mpts}d.
These steps approximately solve Eq. (\ref{eq_mpts_workflows})c and obtain $\mathcal{T}_{t+1}^{\mathcal{B}}$ in a heuristic way.

\subsection{Sequentially Optimize the Adaptive Machine Learner}

Given the screened $\mathcal{T}_{t+1}$, we execute optimization to update the machine learner's parameters.
The task-specific adaptation risk in $(t+1)$-th iteration is written as $\ell_{t+1,i}(\bm\theta)$ for the selected task $\tau_{i}$.
The developed MPTS is agnostic to any-shot learning methods, and the following includes the standard update rule for zero-shot, few-shot, and SFT scenarios.

\paragraph{Machine Learner Updates in Zero-Shot Adaptation:}
The zero-shot setup does not require the support dataset to identify the task.
Hence, taking the vanilla DR \citep{tobin2017domain} as an instantiation, we can obtain the update rule as:
\begin{equation}
    \begin{split}
        \bm\theta_{t+1}=\bm\theta_{t}-\frac{\lambda}{\mathcal{B}}\sum_{i=1}^{\mathcal{B}}\nabla_{\bm\theta}\ell(\mathcal{D}_{\tau_{t+1,i}}^{Q};\bm\theta_{t}),
    \end{split}
\end{equation}
where $\bm\theta$ denotes the zero-shot learning model parameter with $\lambda$ the learning rate.

\paragraph{Machine Learner Updates in Few-Shot Adaptation:}
Still, we take the typical optimization-based method MAML \citep{finn2017model} as an instantiation and provide the update rule as follows:
\begin{subequations}
    \begin{align}
        \ell_{t+1,i}(\bm\theta)
        =\ell(\mathcal{D}_{\tau_{t+1,i}}^{Q};\bm\theta^{\text{meta}}_{t}-\lambda_{1,1}\nabla_{\bm\theta}\ell(\mathcal{D}_{\tau_{t+1,i}}^{S}))
        \\
        \bm\theta^{\text{meta}}_{t+1}=
        \bm\theta^{\text{meta}}_{t}-\frac{\lambda_{1,2}}{\mathcal{B}}\sum_{i=1}^{\mathcal{B}}\nabla_{\bm\theta}\ell_{t+1,i}(\bm\theta),\quad\forall i\in\{1,\dots,\mathcal{B}\}
    \end{align}
\end{subequations}
where $\bm\theta^{\text{meta}}$ denotes the meta initialization, and $\lambda_{1,1}$ and $\lambda_{1,2}$ are, respectively, learning rates in the inner and outer loops.

\paragraph{Machine Learner Updates in SFT:}
Here, we take finetuning pretrained models to downstream tasks \citep{ding2023parameter} as an instantiation.
In this case, each data point $[\bm x,\bm y]$ can be viewed as a task with either its embedding $\bm\tau$ or $\bm x$ as the task identifier.
Then the model update rule can be:

\begin{equation}
    \begin{split}
        \bm\theta_{t+1}=\bm\theta_{t}-\frac{\lambda}{\mathcal{B}}\sum_{i=1}^{\mathcal{B}}\nabla_{\bm\theta}\ell([\bm x_{t+1,i},\bm y_{t+1,i}];\bm\theta_{t}),
    \end{split}
\end{equation}
where $\{[\bm x_{t+1,i},\bm y_{t+1,i}]\}_{i=1}^{\mathcal{B}}$ denote the sampled task batch for the $(t+1)$-th iteration.

\subsection{Overall Algorithm and Interpretation}

\paragraph{Implementation Pipelines.}
Here, we write the general form of MPTS in Algorithm \ref{alg_mpts}, where the past risk episodes are reused to train the RPM and serve the active subset selection.
We also provide some implementation examples by putting all the ingredients and optimization recipes together in the zero-shot, few-shot, and SFT scenarios.
See Algorithms~\ref{dr_pseudo}--\ref{ft_cts_pseudo} in Appendix~\ref{sec_supp_guideline} for details.
Since the first iteration in Algorithm 2/4/6 does not involve active sampling, due to no latest history, and the task batch follows the standard random sampling setup.

\paragraph{Connection with Sequential Decision-making and Control.}
Intuitively, MPTS resembles model predictive control (MPC) \citep{morari1999model} when treating task sampling under some criteria as an optimal planning problem.
In this case, the episodic learning process specifies an underlying dynamical system for MPTS to predict with only one future time step in the simulation to assess the influence of selecting the task batch, and the feedback as exact adaptation risk information further helps improve the episodic risk prediction system.
In addition, through the lens of sequential decision-making, we can interpret the optimization pipeline of MPTS from the actor-critic framework in RL \citep{konda1999actor}.
In detail, the RPM works as the critic that predicts adaptation performance in the task $\tau$ given a fixed machine learner.
Accordingly, the actor plays the role of selecting the task batch from the acquisition function and then executing the machine learner's optimization.
These two roles are entangled in the MPTS pipeline to achieve robust yet efficient adaptation.

\paragraph{Sample and Compute Complexity Analysis in Amortized Task Evaluation.} 
Within the framework of subset selection methods like $\text{CVaR}_{\alpha}$, while the efficiency gain of MPTS represents a constant factor, it constitutes the definitive boundary between computational feasibility and practical proscription in regimes where $\hat{\mathcal{B}} \gg \mathcal{B}$. 
For example, optimizing $\text{CVaR}_{0.9}$ requires a candidate pool ten times the size of the update batch ($\hat{\mathcal{B}} = 10\mathcal{B}$); standard methods would thus incur the redundant cost of evaluating $9\mathcal{B}$ tasks that are ultimately discarded in each iteration. 
By substituting these exact evaluations with predictive signals, MPTS renders robust adaptation viable for foundation models and RL scenarios. 
Essentially, as the risk-sensitivity parameter $\alpha$ increases, the amortized inference of MPTS becomes more significant for mitigating the prohibitive costs of exact task evaluation.

\paragraph{MPTS versus $\text{CVaR}_{\alpha}$ in Convergence Behaviors.}
Under worst subset selection principles, MPTS and vanilla $\text{CVaR}_{\alpha}$ target different convergence points, as identical solutions would require unrealistic perfect prediction assumptions. 
Instead, we formalize MPTS as a biased surrogate: Theorem \ref{theorem_prob_score} guarantees that task difficulty rankings are roughly preserved, while Theorem \ref{theorem_converge} ensures stable convergence of the resulting biased surrogate objective under this controlled approximation.
The vanilla $\text{CVaR}_{\alpha}$ relies on purely greedy selection and requires prohibitively large batches for accurate quantile estimation, i.e., suffering from $\mathcal{O}(1/\sqrt{\mathcal{B}})$ error.
In contrast, MPTS incorporates an exploration bonus in Eq. (\ref{eq_acq}) besides the robust optimization.
And this allows the model to balance worst-case risk with task-space uncertainty, avoiding the significant drop in average performance often seen in standard $\text{CVaR}_{\alpha}$. 
We also emphasize that MPTS intentionally provides weaker guarantees than exact $\text{CVaR}_\alpha$ optimization. 
while irreducible gradient bias and exploration terms lead to different stationary points, MPTS fulfills the same robust intent with provable stability.
Ultimately, MPTS depends on the empirical quality of task ranking predictions, and improving RPM's expressiveness to maintain ranking accuracy remains a promising direction for future work.

\paragraph{Comparison between Two Robust Optimization Paradigms.}
Robust optimization methods generally follow two paradigms.
(1) Subset selection methods, e.g., $\text{CVaR}_\alpha$, which train on worst-case task subsets to reduce per-iteration cost. 
However, exact subset construction in $\text{CVaR}_\alpha$ still requires evaluating all $\hat{\mathcal{B}}$ candidate tasks.
(2) Task balancing methods, e.g., GroupDRO \citep{sagawa2019distributionally}, ExcessMTL \citep{he2024robust}, to optimize all tasks jointly with adaptive weights, which utilize the reweighting mechanism yet is limited by the task batch size due to computational and memory costs.
MPTS bridges these paradigms by combining efficient subset selection with lightweight predictive inference, replacing exhaustive evaluation with RPM-based approximation to enable scalable worst-case optimization in large-scale settings.

\section*{Data Availability}
Here, we declare that the source of data used in experiments is from open-source dataset repositories or widely adopted simulators with free licenses.
The sinusoid simulator at \url{https://github.com/cbfinn/maml} is from open-sourced MAML \citep{finn2017model}.
The six few-shot image classification datasets are from ImageNet-CG \citep{hendrycks2019benchmarking}, ImageNet-CI \citep{hendrycks2019benchmarking}, ImageNet-CS \citep{hendrycks2019benchmarking}, ImageNet-A \citep{hendrycks2021natural}, ImageNet-S \citep{wang2019learning} and ImageNet-R \citep{hendrycks2021many}.
The image dataset for SFT is from ImageNet \citep{russakovsky2015imagenet} and four OOD datasets-ImageNet-A \citep{hendrycks2021natural}, ImageNet-S \citep{wang2019learning}, ImageNet-R \citep{hendrycks2021many}, and ImageNet-V \citep{recht2019imagenet}. 
Note that the ImageNet data is available and free to researchers for non-commercial use
The Meta-RL simulators are based on the Mujoco robotic software with configurations specified in the manuscript.
The DR environments Ergo-Reacher and Lunar-Lander at \url{https://github.com/montrealrobotics/active-domainrand/tree/master/common/envs} are from open-sourced work \citep{mehta2020active}.
As for the data of experimental results, it is reflected in the Tables and Figures, which can be found in the attached manuscript file and the source data file.

\section*{Code Availability}
The code of Model Predictive Task Sampling and demonstration is accessible at \url{https://github.com/thu-rllab/MPTS}.
Meanwhile, the specified Github code with DOI is made at \url{https://doi.org/10.5281/zenodo.20048588}. 

\section*{Acknowledgments}

We thank Yuhang Jiang, Heming Zou, Lizhou Cai, and Yi Zhang for helpful feedback and discussions on this work.

\section*{Funding Statement}

This work was supported by the Fundamental and Interdisciplinary Disciplines Breakthrough Plan of the Ministry of Education of China (JYB2025XDXM503).

\section*{Author Contributions Statement}

The authors of this work are Qi Cheems Wang (Q.C.W.), Zehao Xiao (Z.X.), Yixiu Mao (Y.M.), Yun Qu (Y.Q.), Jiayi Shen (J.S.), Yiqin Lv (Y.L.), and Xiangyang Ji (X.J.). X.J. initiated and sponsored this research; Z.X. and J.S., affiliated with the University of Amsterdam, contributed during an internship at Tsinghua University. 
Under X.J.'s supervision, Q.C.W. conceptualized the MPTS framework, developed the mathematical formulations, and wrote the manuscript. 
Y.Q., Y.M., and Y.L. implemented MPTS in sinusoid regression, Meta-RL, and robotic domain randomization, while Z.X. and J.S. implemented MPTS in prompt-based few-shot image classification, and SFT image classification.
Y.M., Y.Q., and Z.X. collected and visualized experimental results and contributed implementation details to the Supplementary Information (Appendix). 
X.J. supervised the research progress, organized technical discussions, and reviewed and revised the manuscript. 
All authors have read and approved the final manuscript for publication.

\section*{Competing Interests Statement}

The authors declare no competing interests.

\bibliography{refs}

@article{brown2020language,
  title={Language models are few-shot learners},
  author={Brown, Tom and Mann, Benjamin and Ryder, Nick and Subbiah, Melanie and Kaplan, Jared D and Dhariwal, Prafulla and Neelakantan, Arvind and Shyam, Pranav and Sastry, Girish and Askell, Amanda and others},
  journal={Advances in neural information processing systems},
  volume={33},
  pages={1877--1901},
  year={2020}
}

@inproceedings{he2024robust,
  title={Robust Multi-Task Learning with Excess Risks},
  author={He, Yifei and Zhou, Shiji and Zhang, Guojun and Yun, Hyokun and Xu, Yi and Zeng, Belinda and Chilimbi, Trishul and Zhao, Han},
  booktitle={International Conference on Machine Learning},
  pages={18094--18114},
  year={2024},
  organization={PMLR}
}

@article{qu2025can,
  title={Can prompt difficulty be online predicted for accelerating rl finetuning of reasoning models?},
  author={Qu, Yun and Wang, Qi and Mao, Yixiu and Hu, Vincent Tao and Ommer, Bj{\"o}rn and Ji, Xiangyang},
  journal={arXiv preprint arXiv:2507.04632},
  year={2025}
}

@article{wang2024towards,
  title={Towards task sampler learning for meta-learning},
  author={Wang, Jingyao and Qiang, Wenwen and Su, Xingzhe and Zheng, Changwen and Sun, Fuchun and Xiong, Hui},
  journal={International Journal of Computer Vision},
  volume={132},
  number={12},
  pages={5534--5564},
  year={2024},
  publisher={Springer}
}

@inproceedings{kumar2023effect,
  title={The effect of diversity in meta-learning},
  author={Kumar, Ramnath and Deleu, Tristan and Bengio, Yoshua},
  booktitle={Proceedings of the AAAI Conference on Artificial Intelligence},
  volume={37},
  number={7},
  pages={8396--8404},
  year={2023}
}

@inproceedings{toloubidokhti2023dats,
  title={Dats: Difficulty-aware task sampler for meta-learning physics-informed neural networks},
  author={Toloubidokhti, Maryam and Ye, Yubo and Missel, Ryan and Jiang, Xiajun and Kumar, Nilesh and Shrestha, Ruby and Wang, Linwei},
  booktitle={The Twelfth International Conference on Learning Representations},
  year={2023}
}

@article{stephan2017stochastic,
  title={Stochastic gradient descent as approximate bayesian inference},
  author={Stephan, Mandt and Hoffman, Matthew D and Blei, David M and others},
  journal={Journal of Machine Learning Research},
  volume={18},
  number={134},
  pages={1--35},
  year={2017}
}

@article{akkaya2019solving,
  title={Solving rubik's cube with a robot hand},
  author={Akkaya, Ilge and Andrychowicz, Marcin and Chociej, Maciek and Litwin, Mateusz and McGrew, Bob and Petron, Arthur and Paino, Alex and Plappert, Matthias and Powell, Glenn and Ribas, Raphael and others},
  journal={arXiv preprint arXiv:1910.07113},
  year={2019}
}

@article{ajay2022distributionally,
  title={Distributionally adaptive meta reinforcement learning},
  author={Ajay, Anurag and Gupta, Abhishek and Ghosh, Dibya and Levine, Sergey and Agrawal, Pulkit},
  journal={Advances in Neural Information Processing Systems},
  volume={35},
  pages={25856--25869},
  year={2022}
}

@misc{auer2002finite,
  title={Finite-time Analysis of the Multiarmed Bandit Problem},
  author={Auer, P},
  year={2002},
  publisher={Kluwer Academic Publishers}
}

@book{tomczak2024deep,
  title={Deep Generative Modeling},
  author={Tomczak, Jakub M},
  publisher={Springer Cham},
  year={2024}
}

@inproceedings{rezende2014stochastic,
  title={Stochastic backpropagation and approximate inference in deep generative models},
  author={Rezende, Danilo Jimenez and Mohamed, Shakir and Wierstra, Daan},
  booktitle={International conference on machine learning},
  pages={1278--1286},
  year={2014},
  organization={PMLR}
}

@inproceedings{gondal2024domain,
  title={Domain aligned CLIP for few-shot classification},
  author={Gondal, Muhammad Waleed and Gast, Jochen and Ruiz, Inigo Alonso and Droste, Richard and Macri, Tommaso and Kumar, Suren and Staudigl, Luitpold},
  booktitle={Proceedings of the IEEE/CVF Winter Conference on Applications of Computer Vision},
  pages={5721--5730},
  year={2024}
}

@article{zheng2024rapid,
  title={Rapid context inference in a thalamocortical model using recurrent neural networks},
  author={Zheng, Wei-Long and Wu, Zhongxuan and Hummos, Ali and Yang, Guangyu Robert and Halassa, Michael M},
  journal={Nature Communications},
  volume={15},
  number={1},
  pages={8275},
  year={2024},
  publisher={Nature Publishing Group UK London}
}

@article{dubey2024llama,
  title={The llama 3 herd of models},
  author={Dubey, Abhimanyu and Jauhri, Abhinav and Pandey, Abhinav and Kadian, Abhishek and Al-Dahle, Ahmad and Letman, Aiesha and Mathur, Akhil and Schelten, Alan and Yang, Amy and Fan, Angela and others},
  journal={arXiv preprint arXiv:2407.21783},
  year={2024}
}

@article{rigollet2023high,
  title={High-dimensional statistics},
  author={Rigollet, Philippe and H{\"u}tter, Jan-Christian},
  journal={arXiv preprint arXiv:2310.19244},
  year={2023}
}

@inproceedings{hejnaremix,
  title={ReMix: Optimizing Data Mixtures for Large Scale Imitation Learning},
  author={Hejna, Joey and Bhateja, Chethan Anand and Jiang, Yichen and Pertsch, Karl and Sadigh, Dorsa},
  booktitle={8th Annual Conference on Robot Learning},
  year={2024}
}

@inproceedings{todorov2012mujoco,
  title={Mujoco: A physics engine for model-based control},
  author={Todorov, Emanuel and Erez, Tom and Tassa, Yuval},
  booktitle={2012 IEEE/RSJ international conference on intelligent robots and systems},
  pages={5026--5033},
  year={2012},
  organization={IEEE}
}

@article{xie2024doremi,
  title={Doremi: Optimizing data mixtures speeds up language model pretraining},
  author={Xie, Sang Michael and Pham, Hieu and Dong, Xuanyi and Du, Nan and Liu, Hanxiao and Lu, Yifeng and Liang, Percy S and Le, Quoc V and Ma, Tengyu and Yu, Adams Wei},
  journal={Advances in Neural Information Processing Systems},
  volume={36},
  year={2024}
}

@inproceedings{recht2019imagenet,
  title={Do imagenet classifiers generalize to imagenet?},
  author={Recht, Benjamin and Roelofs, Rebecca and Schmidt, Ludwig and Shankar, Vaishaal},
  booktitle={ICML},
  year={2019},
}

@article{russakovsky2015imagenet,
  title={Imagenet large scale visual recognition challenge},
  author={Russakovsky, Olga and Deng, Jia and Su, Hao and Krause, Jonathan and Satheesh, Sanjeev and Ma, Sean and Huang, Zhiheng and Karpathy, Andrej and Khosla, Aditya and Bernstein, Michael and others},
  journal={IJCV},
  year={2015},
}

@inproceedings{hendrycks2019benchmarking,
  title={Benchmarking Neural Network Robustness to Common Corruptions and Perturbations},
  author={Hendrycks, Dan and Dietterich, Thomas},
  booktitle={ICLR},
  year={2019}
}

@article{friston2016active,
  title={Active inference and learning},
  author={Friston, Karl and FitzGerald, Thomas and Rigoli, Francesco and Schwartenbeck, Philipp and Pezzulo, Giovanni and others},
  journal={Neuroscience \& Biobehavioral Reviews},
  volume={68},
  pages={862--879},
  year={2016},
  publisher={Elsevier}
}

@article{hendrycks2021many,
  title={The Many Faces of Robustness: A Critical Analysis of Out-of-Distribution Generalization},
  author={Dan Hendrycks and Steven Basart and Norman Mu and Saurav Kadavath and Frank Wang and Evan Dorundo and Rahul Desai and Tyler Zhu and Samyak Parajuli and Mike Guo and Dawn Song and Jacob Steinhardt and Justin Gilmer},
  journal={ICCV},
  year={2021}
}

@article{wang2019learning,
  title={Learning robust global representations by penalizing local predictive power},
  author={Wang, Haohan and Ge, Songwei and Lipton, Zachary and Xing, Eric P},
  journal={Advances in Neural Information Processing Systems},
  volume={32},
  year={2019}
}

@inproceedings{hendrycks2021natural,
  title={Natural adversarial examples},
  author={Hendrycks, Dan and Zhao, Kevin and Basart, Steven and Steinhardt, Jacob and Song, Dawn},
  booktitle={Proceedings of the IEEE/CVF conference on computer vision and pattern recognition},
  pages={15262--15271},
  year={2021}
}

@inproceedings{zhou2022conditional,
  title={Conditional prompt learning for vision-language models},
  author={Zhou, Kaiyang and Yang, Jingkang and Loy, Chen Change and Liu, Ziwei},
  booktitle={Proceedings of the IEEE/CVF conference on computer vision and pattern recognition},
  pages={16816--16825},
  year={2022}
}

@article{zhou2022learning,
  title={Learning to prompt for vision-language models},
  author={Zhou, Kaiyang and Yang, Jingkang and Loy, Chen Change and Liu, Ziwei},
  journal={International Journal of Computer Vision},
  volume={130},
  number={9},
  pages={2337--2348},
  year={2022},
  publisher={Springer}
}

@article{broderick2013streaming,
  title={Streaming variational bayes},
  author={Broderick, Tamara and Boyd, Nicholas and Wibisono, Andre and Wilson, Ashia C and Jordan, Michael I},
  journal={Advances in neural information processing systems},
  volume={26},
  year={2013}
}

@article{wang2023large,
  title={Large-scale generative simulation artificial intelligence: The next hotspot},
  author={Wang, Qi and Feng, Yanghe and Huang, Jincai and Lv, Yiqin and Xie, Zheng and Gao, Xiaoshan},
  journal={The Innovation},
  volume={4},
  number={6},
  year={2023},
  publisher={Elsevier}
}

@article{duan2016rl,
  title={Rl2: Fast reinforcement learning via slow reinforcement learning},
  author={Duan, Yan and Schulman, John and Chen, Xi and Bartlett, Peter L and Sutskever, Ilya and Abbeel, Pieter},
  journal={arXiv preprint arXiv:1611.02779},
  year={2016}
}

@inproceedings{allen2019infinite,
  title={Infinite mixture prototypes for few-shot learning},
  author={Allen, Kelsey and Shelhamer, Evan and Shin, Hanul and Tenenbaum, Joshua},
  booktitle={International conference on machine learning},
  pages={232--241},
  year={2019},
  organization={PMLR}
}

@inproceedings{sendera2023hypershot,
  title={Hypershot: Few-shot learning by kernel hypernetworks},
  author={Sendera, Marcin and Przewi{\k{e}}{\'z}likowski, Marcin and Karanowski, Konrad and Zi{\k{e}}ba, Maciej and Tabor, Jacek and Spurek, Przemys{\l}aw},
  booktitle={Proceedings of the IEEE/CVF winter conference on applications of computer vision},
  pages={2469--2478},
  year={2023}
}

@article{ha2016hypernetworks,
  title={Hypernetworks},
  author={Ha, David and Dai, Andrew and Le, Quoc V},
  journal={arXiv preprint arXiv:1609.09106},
  year={2016}
}

@article{snell2017prototypical,
  title={Prototypical networks for few-shot learning},
  author={Snell, Jake and Swersky, Kevin and Zemel, Richard},
  journal={Advances in neural information processing systems},
  volume={30},
  year={2017}
}

@article{rajeswaran2019meta,
  title={Meta-learning with implicit gradients},
  author={Rajeswaran, Aravind and Finn, Chelsea and Kakade, Sham M and Levine, Sergey},
  journal={Advances in neural information processing systems},
  volume={32},
  year={2019}
}

@inproceedings{abbas2022sharp,
  title={Sharp-maml: Sharpness-aware model-agnostic meta learning},
  author={Abbas, Momin and Xiao, Quan and Chen, Lisha and Chen, Pin-Yu and Chen, Tianyi},
  booktitle={International conference on machine learning},
  pages={10--32},
  year={2022},
  organization={PMLR}
}

@article{finn2018probabilistic,
  title={Probabilistic model-agnostic meta-learning},
  author={Finn, Chelsea and Xu, Kelvin and Levine, Sergey},
  journal={Advances in neural information processing systems},
  volume={31},
  year={2018}
}

@inproceedings{garnelo2018conditional,
  title={Conditional neural processes},
  author={Garnelo, Marta and Rosenbaum, Dan and Maddison, Christopher and Ramalho, Tiago and Saxton, David and Shanahan, Murray and Teh, Yee Whye and Rezende, Danilo and Eslami, SM Ali},
  booktitle={International conference on machine learning},
  pages={1704--1713},
  year={2018},
  organization={PMLR}
}

@inproceedings{finn2017model,
  title={Model-agnostic meta-learning for fast adaptation of deep networks},
  author={Finn, Chelsea and Abbeel, Pieter and Levine, Sergey},
  booktitle={International conference on machine learning},
  pages={1126--1135},
  year={2017},
  organization={PMLR}
}

@inproceedings{wang2022bridge,
  title={Bridge the inference gaps of neural processes via expectation maximization},
  author={Wang, Qi and Federici, Marco and van Hoof, Herke},
  booktitle={The Eleventh International Conference on Learning Representations},
  year={2022}
}

@inproceedings{gondal2021function,
  title={Function contrastive learning of transferable meta-representations},
  author={Gondal, Muhammad Waleed and Joshi, Shruti and Rahaman, Nasim and Bauer, Stefan and Wuthrich, Manuel and Sch{\"o}lkopf, Bernhard},
  booktitle={International Conference on Machine Learning},
  pages={3755--3765},
  year={2021},
  organization={PMLR}
}

@article{garnelo2018neural,
  title={Neural processes},
  author={Garnelo, Marta and Schwarz, Jonathan and Rosenbaum, Dan and Viola, Fabio and Rezende, Danilo J and Eslami, SM and Teh, Yee Whye},
  journal={arXiv preprint arXiv:1807.01622},
  year={2018}
}

@article{wang2020generalizing,
  title={Generalizing from a few examples: A survey on few-shot learning},
  author={Wang, Yaqing and Yao, Quanming and Kwok, James T and Ni, Lionel M},
  journal={ACM computing surveys (csur)},
  volume={53},
  number={3},
  pages={1--34},
  year={2020},
  publisher={ACM New York, NY, USA}
}

@article{hospedales2021meta,
  title={Meta-learning in neural networks: A survey},
  author={Hospedales, Timothy and Antoniou, Antreas and Micaelli, Paul and Storkey, Amos},
  journal={IEEE transactions on pattern analysis and machine intelligence},
  volume={44},
  number={9},
  pages={5149--5169},
  year={2021},
  publisher={IEEE}
}

@article{kaddour2020probabilistic,
  title={Probabilistic active meta-learning},
  author={Kaddour, Jean and S{\ae}mundsson, Steind{\'o}r and others},
  journal={Advances in Neural Information Processing Systems},
  volume={33},
  pages={20813--20822},
  year={2020}
}

@article{wang2024simple,
  title={A Simple Yet Effective Strategy to Robustify the Meta Learning Paradigm},
  author={Wang, Qi and Lv, Yiqin and Xie, Zheng and Huang, Jincai and others},
  journal={Advances in Neural Information Processing Systems},
  volume={36},
  year={2024}
}

@article{oren2019distributionally,
  title={Distributionally robust language modeling},
  author={Oren, Yonatan and Sagawa, Shiori and Hashimoto, Tatsunori B and Liang, Percy},
  journal={arXiv preprint arXiv:1909.02060},
  year={2019}
}

@inproceedings{
lv2024theoretical,
title={Theoretical Investigations and Practical Enhancements on Tail Task Risk Minimization in Meta Learning},
author={Yiqin Lv and Cheems Wang and Dong Liang and Zheng Xie},
booktitle={The Thirty-eighth Annual Conference on Neural Information Processing Systems},
year={2024},
url={https://openreview.net/forum?id=McrzOo0hwr}
}

@article{konda1999actor,
  title={Actor-critic algorithms},
  author={Konda, Vijay and Tsitsiklis, John},
  journal={Advances in neural information processing systems},
  volume={12},
  year={1999}
}

@article{morari1999model,
  title={Model predictive control: past, present and future},
  author={Morari, Manfred and Lee, Jay H},
  journal={Computers \& chemical engineering},
  volume={23},
  number={4-5},
  pages={667--682},
  year={1999},
  publisher={Elsevier}
}

@article{hessel2021clipscore,
  title={Clipscore: A reference-free evaluation metric for image captioning},
  author={Hessel, Jack and Holtzman, Ari and Forbes, Maxwell and Bras, Ronan Le and Choi, Yejin},
  journal={arXiv preprint arXiv:2104.08718},
  year={2021}
}

@inproceedings{koh2021wilds,
  title={Wilds: A benchmark of in-the-wild distribution shifts},
  author={Koh, Pang Wei and Sagawa, Shiori and Marklund, Henrik and Xie, Sang Michael and Zhang, Marvin and Balsubramani, Akshay and Hu, Weihua and Yasunaga, Michihiro and Phillips, Richard Lanas and Gao, Irena and others},
  booktitle={International conference on machine learning},
  pages={5637--5664},
  year={2021},
  organization={PMLR}
}

@article{evans2023bad,
  title={Bad students make great teachers: Active learning accelerates large-scale visual understanding},
  author={Evans, Talfan and Pathak, Shreya and Merzic, Hamza and Schwarz, Jonathan and Tanno, Ryutaro and Henaff, Olivier J},
  journal={arXiv preprint arXiv:2312.05328},
  year={2023}
}

@inproceedings{mehta2020active,
  title={Active domain randomization},
  author={Mehta, Bhairav and Diaz, Manfred and Golemo, Florian and Pal, Christopher J and Paull, Liam},
  booktitle={Conference on Robot Learning},
  pages={1162--1176},
  year={2020},
  organization={PMLR}
}

@inproceedings{mirzasoleiman2020coresets,
  title={Coresets for data-efficient training of machine learning models},
  author={Mirzasoleiman, Baharan and Bilmes, Jeff and Leskovec, Jure},
  booktitle={International Conference on Machine Learning},
  pages={6950--6960},
  year={2020},
  organization={PMLR}
}

@inproceedings{joshi2023data,
  title={Data-efficient contrastive self-supervised learning: Most beneficial examples for supervised learning contribute the least},
  author={Joshi, Siddharth and Mirzasoleiman, Baharan},
  booktitle={International conference on machine learning},
  pages={15356--15370},
  year={2023},
  organization={PMLR}
}

@inproceedings{yang2023towards,
  title={Towards sustainable learning: Coresets for data-efficient deep learning},
  author={Yang, Yu and Kang, Hao and Mirzasoleiman, Baharan},
  booktitle={International Conference on Machine Learning},
  pages={39314--39330},
  year={2023},
  organization={PMLR}
}

@article{paul2021deep,
  title={Deep learning on a data diet: Finding important examples early in training},
  author={Paul, Mansheej and Ganguli, Surya and Dziugaite, Gintare Karolina},
  journal={Advances in neural information processing systems},
  volume={34},
  pages={20596--20607},
  year={2021}
}

@article{sorscher2022beyond,
  title={Beyond neural scaling laws: beating power law scaling via data pruning},
  author={Sorscher, Ben and Geirhos, Robert and Shekhar, Shashank and Ganguli, Surya and Morcos, Ari},
  journal={Advances in Neural Information Processing Systems},
  volume={35},
  pages={19523--19536},
  year={2022}
}

@article{ouyang2022training,
  title={Training language models to follow instructions with human feedback},
  author={Ouyang, Long and Wu, Jeffrey and Jiang, Xu and Almeida, Diogo and Wainwright, Carroll and Mishkin, Pamela and Zhang, Chong and Agarwal, Sandhini and Slama, Katarina and Ray, Alex and others},
  journal={Advances in neural information processing systems},
  volume={35},
  pages={27730--27744},
  year={2022}
}

@inproceedings{rajeswaran2022epopt,
  title={EPOpt: Learning Robust Neural Network Policies Using Model Ensembles},
  author={Rajeswaran, Aravind and Ghotra, Sarvjeet and Ravindran, Balaraman and Levine, Sergey},
  booktitle={International Conference on Learning Representations},
  year={2022}
}

@inproceedings{khattak2023maple,
  title={Maple: Multi-modal prompt learning},
  author={Khattak, Muhammad Uzair and Rasheed, Hanoona and Maaz, Muhammad and Khan, Salman and Khan, Fahad Shahbaz},
  booktitle={Proceedings of the IEEE/CVF Conference on Computer Vision and Pattern Recognition},
  pages={19113--19122},
  year={2023}
}

@inproceedings{fujimoto2018addressing,
  title={Addressing function approximation error in actor-critic methods},
  author={Fujimoto, Scott and Hoof, Herke and Meger, David},
  booktitle={International conference on machine learning},
  pages={1587--1596},
  year={2018},
  organization={PMLR}
}

@article{stigler1982thomas,
  title={Thomas Bayes's bayesian inference},
  author={Stigler, Stephen M},
  journal={Journal of the Royal Statistical Society: Series A (General)},
  volume={145},
  number={2},
  pages={250--258},
  year={1982},
  publisher={Wiley Online Library}
}

@article{kingma2013auto,
  title={Auto-encoding variational bayes},
  author={Kingma, Diederik P and Welling, Max},
  journal={arXiv preprint arXiv:1312.6114},
  year={2013}
}

@article{wilson2018maximizing,
  title={Maximizing acquisition functions for Bayesian optimization},
  author={Wilson, James and Hutter, Frank and Deisenroth, Marc},
  journal={Advances in neural information processing systems},
  volume={31},
  year={2018}
}

@article{snoek2012practical,
  title={Practical bayesian optimization of machine learning algorithms},
  author={Snoek, Jasper and Larochelle, Hugo and Adams, Ryan P},
  journal={Advances in neural information processing systems},
  volume={25},
  year={2012}
}

@inproceedings{liu2020adaptive,
  title={Adaptive task sampling for meta-learning},
  author={Liu, Chenghao and Wang, Zhihao and Sahoo, Doyen and Fang, Yuan and Zhang, Kun and Hoi, Steven CH},
  booktitle={European Conference on Computer Vision},
  pages={752--769},
  year={2020},
  organization={Springer}
}

@article{collins2020task,
  title={Task-robust model-agnostic meta-learning},
  author={Collins, Liam and Mokhtari, Aryan and Shakkottai, Sanjay},
  journal={Advances in Neural Information Processing Systems},
  volume={33},
  pages={18860--18871},
  year={2020}
}

@article{yao2021meta,
  title={Meta-learning with an adaptive task scheduler},
  author={Yao, Huaxiu and Wang, Yu and Wei, Ying and Zhao, Peilin and Mahdavi, Mehrdad and Lian, Defu and Finn, Chelsea},
  journal={Advances in Neural Information Processing Systems},
  volume={34},
  pages={7497--7509},
  year={2021}
}

@article{nguyen2017variational,
  title={Variational continual learning},
  author={Nguyen, Cuong V and Li, Yingzhen and Bui, Thang D and Turner, Richard E},
  journal={arXiv preprint arXiv:1710.10628},
  year={2017}
}

@article{vapnik1998statistical,
  title={Statistical learning theory},
  author={Vapnik, Vladimir Naumovich and Vapnik, Vlamimir and others},
  year={1998},
  publisher={wiley New York}
}

@inproceedings{damianou2013deep,
  title={Deep gaussian processes},
  author={Damianou, Andreas and Lawrence, Neil D},
  booktitle={Artificial intelligence and statistics},
  pages={207--215},
  year={2013},
  organization={PMLR}
}

@article{greenberg2024train,
  title={Train hard, fight easy: Robust meta reinforcement learning},
  author={Greenberg, Ido and Mannor, Shie and Chechik, Gal and Meirom, Eli},
  journal={Advances in Neural Information Processing Systems},
  volume={36},
  year={2024}
}

@article{friedman2022role,
  title={The role of prefrontal cortex in cognitive control and executive function},
  author={Friedman, Naomi P and Robbins, Trevor W},
  journal={Neuropsychopharmacology},
  volume={47},
  number={1},
  pages={72--89},
  year={2022},
  publisher={Springer International Publishing Cham}
}

@article{wang2018prefrontal,
  title={Prefrontal cortex as a meta-reinforcement learning system},
  author={Wang, Jane X and Kurth-Nelson, Zeb and Kumaran, Dharshan and Tirumala, Dhruva and Soyer, Hubert and Leibo, Joel Z and Hassabis, Demis and Botvinick, Matthew},
  journal={Nature neuroscience},
  volume={21},
  number={6},
  pages={860--868},
  year={2018},
  publisher={Nature Publishing Group}
}

@article{auer2002using,
  title={Using confidence bounds for exploitation-exploration trade-offs},
  author={Auer, Peter},
  journal={Journal of Machine Learning Research},
  volume={3},
  number={Nov},
  pages={397--422},
  year={2002}
}

@inproceedings{sagawa2019distributionally,
  title={Distributionally Robust Neural Networks},
  author={Sagawa, Shiori and Koh, Pang Wei and Hashimoto, Tatsunori B and Liang, Percy},
  booktitle={International Conference on Learning Representations},
  year={2019}
}

@article{wang2022learning,
  title={Learning expressive meta-representations with mixture of expert neural processes},
  author={Wang, Qi and Van Hoof, Herke},
  journal={Advances in neural information processing systems},
  volume={35},
  pages={26242--26255},
  year={2022}
}

@article{kim2019attentive,
  title={Attentive neural processes},
  author={Kim, Hyunjik and Mnih, Andriy and Schwarz, Jonathan and Garnelo, Marta and Eslami, Ali and Rosenbaum, Dan and Vinyals, Oriol and Teh, Yee Whye},
  journal={arXiv preprint arXiv:1901.05761},
  year={2019}
}

@article{foong2020meta,
  title={Meta-learning stationary stochastic process prediction with convolutional neural processes},
  author={Foong, Andrew and Bruinsma, Wessel and Gordon, Jonathan and Dubois, Yann and Requeima, James and Turner, Richard},
  journal={Advances in Neural Information Processing Systems},
  volume={33},
  pages={8284--8295},
  year={2020}
}

@article{gordon2019convolutional,
  title={Convolutional conditional neural processes},
  author={Gordon, Jonathan and Bruinsma, Wessel P and Foong, Andrew YK and Requeima, James and Dubois, Yann and Turner, Richard E},
  journal={arXiv preprint arXiv:1910.13556},
  year={2019}
}

@article{zaheer2017deep,
  title={Deep sets},
  author={Zaheer, Manzil and Kottur, Satwik and Ravanbakhsh, Siamak and Poczos, Barnabas and Salakhutdinov, Russ R and Smola, Alexander J},
  journal={Advances in neural information processing systems},
  volume={30},
  year={2017}
}

@article{mockus1978application,
  title={The application of Bayesian methods for seeking the extremum, vol. 2},
  author={Mockus, J and Tiesis, V and Zilinskas, A},
  journal={L Dixon and G Szego. Toward Global Optimization},
  volume={2},
  year={1978}
}

@inproceedings{ru2018fast,
  title={Fast information-theoretic Bayesian optimisation},
  author={Ru, Binxin and Osborne, Michael A and McLeod, Mark and Granziol, Diego},
  booktitle={International Conference on Machine Learning},
  pages={4384--4392},
  year={2018},
  organization={PMLR}
}

@book{garnett2023bayesian,
  title={Bayesian optimization},
  author={Garnett, Roman},
  year={2023},
  publisher={Cambridge University Press}
}

@inproceedings{kirillov2023segment,
  title={Segment anything},
  author={Kirillov, Alexander and Mintun, Eric and Ravi, Nikhila and Mao, Hanzi and Rolland, Chloe and Gustafson, Laura and Xiao, Tete and Whitehead, Spencer and Berg, Alexander C and Lo, Wan-Yen and others},
  booktitle={Proceedings of the IEEE/CVF International Conference on Computer Vision},
  pages={4015--4026},
  year={2023}
}

@article{liu2018generalized,
  title={Generalized zero-shot learning with deep calibration network},
  author={Liu, Shichen and Long, Mingsheng and Wang, Jianmin and Jordan, Michael I},
  journal={Advances in neural information processing systems},
  volume={31},
  year={2018}
}

@inproceedings{radford2021learning,
  title={Learning transferable visual models from natural language supervision},
  author={Radford, Alec and Kim, Jong Wook and Hallacy, Chris and Ramesh, Aditya and Goh, Gabriel and Agarwal, Sandhini and Sastry, Girish and Askell, Amanda and Mishkin, Pamela and Clark, Jack and others},
  booktitle={International conference on machine learning},
  pages={8748--8763},
  year={2021},
  organization={PMLR}
}

@inproceedings{xian2018feature,
  title={Feature generating networks for zero-shot learning},
  author={Xian, Yongqin and Lorenz, Tobias and Schiele, Bernt and Akata, Zeynep},
  booktitle={Proceedings of the IEEE conference on computer vision and pattern recognition},
  pages={5542--5551},
  year={2018}
}

@inproceedings{keshari2020generalized,
  title={Generalized zero-shot learning via over-complete distribution},
  author={Keshari, Rohit and Singh, Richa and Vatsa, Mayank},
  booktitle={Proceedings of the IEEE/CVF conference on computer vision and pattern recognition},
  pages={13300--13308},
  year={2020}
}

@inproceedings{schonfeld2019generalized,
  title={Generalized zero-and few-shot learning via aligned variational autoencoders},
  author={Schonfeld, Edgar and Ebrahimi, Sayna and Sinha, Samarth and Darrell, Trevor and Akata, Zeynep},
  booktitle={Proceedings of the IEEE/CVF conference on computer vision and pattern recognition},
  pages={8247--8255},
  year={2019}
}

@article{xu2022attribute,
  title={Attribute prototype network for any-shot learning},
  author={Xu, Wenjia and Xian, Yongqin and Wang, Jiuniu and Schiele, Bernt and Akata, Zeynep},
  journal={International Journal of Computer Vision},
  volume={130},
  number={7},
  pages={1735--1753},
  year={2022},
  publisher={Springer}
}

@article{li2020transferrable,
  title={Transferrable feature and projection learning with class hierarchy for zero-shot learning},
  author={Li, Aoxue and Lu, Zhiwu and Guan, Jiechao and Xiang, Tao and Wang, Liwei and Wen, Ji-Rong},
  journal={International Journal of Computer Vision},
  volume={128},
  pages={2810--2827},
  year={2020},
  publisher={Springer}
}

@article{rockafellar2000optimization,
  title={Optimization of conditional value-at-risk},
  author={Rockafellar, R Tyrrell and Uryasev, Stanislav and others},
  journal={Journal of risk},
  volume={2},
  pages={21--42},
  year={2000}
}

@article{ding2023parameter,
  title={Parameter-efficient fine-tuning of large-scale pre-trained language models},
  author={Ding, Ning and Qin, Yujia and Yang, Guang and Wei, Fuchao and Yang, Zonghan and Su, Yusheng and Hu, Shengding and Chen, Yulin and Chan, Chi-Min and Chen, Weize and others},
  journal={Nature Machine Intelligence},
  volume={5},
  number={3},
  pages={220--235},
  year={2023},
  publisher={Nature Publishing Group UK London}
}

@inproceedings{tobin2017domain,
  title={Domain randomization for transferring deep neural networks from simulation to the real world},
  author={Tobin, Josh and Fong, Rachel and Ray, Alex and Schneider, Jonas and Zaremba, Wojciech and Abbeel, Pieter},
  booktitle={2017 IEEE/RSJ international conference on intelligent robots and systems (IROS)},
  pages={23--30},
  year={2017},
  organization={IEEE}
}

@article{evans2024data,
  title={Data curation via joint example selection further accelerates multimodal learning},
  author={Evans, Talfan and Parthasarathy, Nikhil and Merzic, Hamza and Henaff, Olivier J},
  journal={arXiv preprint arXiv:2406.17711},
  year={2024}
}

@article{achiam2023gpt,
  title={Gpt-4 technical report},
  author={Achiam, Josh and Adler, Steven and Agarwal, Sandhini and Ahmad, Lama and Akkaya, Ilge and Aleman, Florencia Leoni and Almeida, Diogo and Altenschmidt, Janko and Altman, Sam and Anadkat, Shyamal and others},
  journal={arXiv preprint arXiv:2303.08774},
  year={2023}
}

@article{zhou2024larger,
  title={Larger and more instructable language models become less reliable},
  author={Zhou, Lexin and Schellaert, Wout and Mart{\'\i}nez-Plumed, Fernando and Moros-Daval, Yael and Ferri, C{\`e}sar and Hern{\'a}ndez-Orallo, Jos{\'e}},
  journal={Nature},
  pages={1--8},
  year={2024},
  publisher={Nature Publishing Group UK London}
}

@inproceedings{sun2024evaluating,
  title={Evaluating the Zero-shot Robustness of Instruction-tuned Language Models},
  author={Sun, Jiuding and Shaib, Chantal and Wallace, Byron C},
  booktitle={International Conference on Learning Representations},
  year={2024},
  organization={ICLR}
}

@inproceedings{rempe2022generating,
  title={Generating useful accident-prone driving scenarios via a learned traffic prior},
  author={Rempe, Davis and Philion, Jonah and Guibas, Leonidas J and Fidler, Sanja and Litany, Or},
  booktitle={Proceedings of the IEEE/CVF Conference on Computer Vision and Pattern Recognition},
  pages={17305--17315},
  year={2022}
}

@article{alayrac2022flamingo,
  title={Flamingo: a visual language model for few-shot learning},
  author={Alayrac, Jean-Baptiste and Donahue, Jeff and Luc, Pauline and Miech, Antoine and Barr, Iain and Hasson, Yana and Lenc, Karel and Mensch, Arthur and Millican, Katherine and Reynolds, Malcolm and others},
  journal={Advances in neural information processing systems},
  volume={35},
  pages={23716--23736},
  year={2022}
}

@inproceedings{golemo2018sim,
  title={Sim-to-real transfer with neural-augmented robot simulation},
  author={Golemo, Florian and Taiga, Adrien Ali and Courville, Aaron and Oudeyer, Pierre-Yves},
  booktitle={Conference on Robot Learning},
  pages={817--828},
  year={2018},
  organization={PMLR}
}

@article{greenberg2023train,
  title={Train hard, fight easy: Robust meta reinforcement learning},
  author={Greenberg, Ido and Mannor, Shie and Chechik, Gal and Meirom, Eli},
  journal={Advances in Neural Information Processing Systems},
  volume={36},
  pages={68276--68299},
  year={2023}
}

@inproceedings{rakelly2019efficient,
  title={Efficient off-policy meta-reinforcement learning via probabilistic context variables},
  author={Rakelly, Kate and Zhou, Aurick and Finn, Chelsea and Levine, Sergey and Quillen, Deirdre},
  booktitle={International conference on machine learning},
  pages={5331--5340},
  year={2019},
  organization={PMLR}
}

@article{nichol2018first,
  title={On first-order meta-learning algorithms},
  author={Nichol, Alex and Achiam, Joshua and Schulman, John},
  journal={arXiv preprint arXiv:1803.02999},
  year={2018}
}

@incollection{coumans2015bullet,
  title={Bullet physics simulation},
  author={Coumans, Erwin},
  booktitle={ACM SIGGRAPH 2015 Courses},
  pages={1},
  year={2015}
}

@misc{box2d,
  author = {Erin Catto},
  title = {Box2D: A 2D Physics Engine for Games},
  year = {2007},
  url = {http://box2d.org}
}

\clearpage
\appendix
\etocdepthtag.toc{supppart}

\setcounter{figure}{0}\renewcommand{\thefigure}{S\arabic{figure}}
\setcounter{table}{0}\renewcommand{\thetable}{S\arabic{table}}
\setcounter{algocf}{0}\renewcommand{\thealgocf}{S\arabic{algocf}}
\providecommand{\theHfigure}{}\renewcommand{\theHfigure}{S\arabic{figure}}
\providecommand{\theHtable}{}\renewcommand{\theHtable}{S\arabic{table}}
\providecommand{\theHalgocf}{}\renewcommand{\theHalgocf}{S\arabic{algocf}}

\begin{tcolorbox}[enhanced, colback=thupanel, colframe=thupanel, boxrule=0pt,
    arc=7pt, outer arc=7pt, borderline west={2.6pt}{0pt}{thupurple},
    left=16pt, right=16pt, top=14pt, bottom=12pt]
  {\sffamily\bfseries\fontsize{16}{20}\selectfont\color{thupurpledark}
   Supplementary Information\par}
  \vspace{4pt}
  {\fontsize{9.3}{12.3}\selectfont\color{thugray}
   for \textit{Model Predictive Task Sampling for Efficient and Robust Adaptation}\par}
  \vspace{6pt}
  \etocsettocstyle{}{}
  \etocsettagdepth{mainpart}{none}
  \etocsettagdepth{supppart}{subsection}
  \tableofcontents
\end{tcolorbox}
\pdfbookmark[0]{Supplementary Information}{suppinfo}

\section{Quick Guideline to MPTS}\label{sec_supp_guideline}

Task episodic learning serves as a cornerstone in developing adaptive models by structuring diverse, context-rich learning experiences. 
One of the pivotal insights underpinning this process is the neural scaling law, which establishes a relationship between task volume, model complexity, and computational resources, offering a principled insight into training foundation models at a certain budget. 
Recent viewpoints have also shed light on the importance of task quality \citep{paul2021deep,sorscher2022beyond,hessel2021clipscore,evans2023bad,evans2024data,mirzasoleiman2020coresets,yang2023towards,joshi2023data}, prompting innovative data curation strategies to refine datasets for pretraining, meta-training, and post-training. 
Evidence suggests that carefully curated data can significantly reduce task sampling complexity, decrease computational demands, and enhance robustness against distributional shifts, sometimes achieving these goals simultaneously. 
Despite these advancements, a practical operation such as Evaluate-Rank-Filter still faces challenges associated with costly evaluations from intensive task queries, computational overhead, and massive annotations. 
Addressing these bottlenecks remains essential to fully realize the potential of task episodic learning in robust efficient foundation model training.

\begin{figure}[!htbp]
\noindent\begin{minipage}[t]{0.485\textwidth}
\begin{algorithm}[H]
\SetAlgoLined
\SetKwInOut{Input}{Input}
\SetKwInOut{Output}{Output}
\Input{Task distribution $p(\tau)$;
 Task batch size $\mathcal{B}$;
 Learning rate $\lambda_1$.}
\Output{Adapted machine learner $\bm\theta$.}
Set the initial iteration number $t=1$;

Randomly initialize machine learner $\bm\theta$;

Randomly initialize RPM $\{\bm\psi,\bm\phi\}$;

\While{not converged}{
Execute \textbf{Algorithm} \ref{dr_cts_pseudo} to access the batch $\{\bm\tau_{t,i}\}_{i=1}^{\mathcal{B}}$ and induced $\{\mathcal{D}_{\tau_{t,i}}^{Q}\}_{i=1}^{\mathcal{B}}$;

\tcp{\textcolor{blue}{\textbf{Eval Adaptation Performance}}}
Compute the task specific adaptation risk $\{\ell_{t,i}:=\ell(\mathcal{D}_{\tau_{t,i}}^{Q};\bm\theta_{t})\}_{i=1}^{\mathcal{B}}$;

Return $H_{t}=\{[\bm\tau_{t,i},\ell_{t,i}]\}_{i=1}^{\mathcal{B}}$ as the Input to \textbf{Algorithm} \ref{dr_cts_pseudo};

\tcp{\textcolor{blue}{\textbf{Update Machine Learner}}}
Perform batch gradient updates:

$\bm\theta_{t+1}\leftarrow\bm\theta_{t}-\frac{\lambda_1}{\mathcal{B}}\sum_{i=1}^{\mathcal{B}}\nabla_{\bm\theta}\ell_{t,i}$;

Update the iteration number: $t\leftarrow t+1$;

}
\caption{MPTS for DR (Zero-Shot Scenarios)}
\label{dr_pseudo}
\end{algorithm}
\end{minipage}\hfill
\begin{minipage}[t]{0.485\textwidth}
\begin{algorithm}[H]
\SetAlgoLined
\SetKwInOut{Input}{Input}
\SetKwInOut{Output}{Output}
\Input{Task distribution $p(\tau)$;
 Task batch size $\mathcal{B}$;
 Candidate batch size $\hat{\mathcal{B}}$;
 Latest updated $\{\bm\psi,\bm\phi\}$;
 Latest history $H_{t-1}$;
 Iteration number $K$;
 Learning rate $\lambda_{2}$.}
\Output{Task identifier batch $\{\bm\tau_{t,i}\}_{i=1}^{\mathcal{B}}$.}

\tcp{\textcolor{red}{\textbf{Functional Posterior Inference}}}
\For{$i=1$ \KwTo $K$}
{   
Perform gradient updates given $H_{t-1}$:

$\bm\phi\leftarrow\bm\phi+\lambda_{2}\nabla_{\bm\phi}\mathcal{G}_{\text{ELBO}}(\bm\psi,\bm\phi)$ in Eq. (17b);

$\bm\psi\leftarrow\bm\psi+\lambda_{2}\nabla_{\bm\psi}\mathcal{G}_{\text{ELBO}}(\bm\psi,\bm\phi)$ in Eq. (17b);

}

\tcp{\textcolor{red}{\textbf{Simulating Adaptation Results}}}
Randomly sample $\{\bm\hat{\bm\tau}_{t,i}\}_{i=1}^{\hat{\mathcal{B}}}$ from $p(\tau)$;

Run amortized evaluation on candidate tasks $\{\delta_{i}:=\gamma_{0}m(\ell_i)+\gamma_{1}\sigma(\ell_i)\}_{i=1}^{\hat{\mathcal{B}}}$ in Eq. (19);

\tcp{\textcolor{red}{\textbf{Active Subset Selection from Predicted Results}}}
Rank $\{\delta_{i}\}_{i=1}^{\hat{\mathcal{B}}}$ and screen Top-$\mathcal{B}$ values;

Return the screened task batch $\{\bm\tau_{t,i}\}_{i=1}^{\mathcal{B}}$.

\caption{Model Predictive Task Sampling}
\label{dr_cts_pseudo}
\end{algorithm}
\end{minipage}
\end{figure}

\begin{figure}[!htbp]
\noindent\begin{minipage}[t]{0.485\textwidth}
\begin{algorithm}[H]
\SetAlgoLined
\SetKwInOut{Input}{Input}
\SetKwInOut{Output}{Output}
\Input{Task distribution $p(\tau)$;
 Task batch size $\mathcal{B}$;
 Learning rates: $\{\lambda_{1,1},\lambda_{1,2}\}$.}
\Output{Meta-trained initialization $\bm\theta^{\text{meta}}$.}
Set the initial iteration number $t=1$;

Randomly initialize meta learner $\bm\theta^{\text{meta}}$;

Randomly initialize RPM $\{\bm\psi,\bm\phi\}$;

\While{not converged}{
Execute \textbf{Algorithm} \ref{few_cts_pseudo} to access the batch $\{\bm\tau_{t,i}\}_{i=1}^{\mathcal{B}}$ and $\{\mathcal{D}_{\tau_{t,i}}^{S}\cup\mathcal{D}_{\tau_{t,i}}^{Q}\}_{i=1}^{\mathcal{B}}$;

\tcp{\textcolor{blue}{\textbf{Inner Loop to Fast Adapt}}}
\For{$i=1$ {\bfseries to} $K$}{
Compute the task-specific gradient: $\nabla_{\bm\theta}\ell(\mathcal{D}_{\tau_{t,i}}^{S};\bm\theta)$;

Perform gradient updates as fast adaptation:

$\bm\theta_{t}^{i}\leftarrow\bm\theta_{t}^{\text{meta}}-\lambda_{1,1}\nabla_{\bm\theta}\ell(\mathcal{D}_{\tau_i}^{S};\bm\theta)$;
}

\tcp{\textcolor{blue}{\textbf{Outer Loop to Meta-train}}}
Evaluate fast adaptation performance $\{\ell_{t,i}:=\ell(\mathcal{D}_{\tau_{t,i}}^{Q};\bm\theta_{t}^{i})\}_{i=1}^{\mathcal{B}}$;

Return $H_{t}=\{[\bm\tau_{t,i},\ell_{t,i}]\}_{i=1}^{\mathcal{B}}$ as the Input to \textbf{Algorithm} \ref{few_cts_pseudo};

Perform meta initialization updates:

$\bm\theta_{t+1}^{\text{meta}}\leftarrow\bm\theta_{t}^{\text{meta}}-\frac{\lambda_{1,2}}{\mathcal{B}}\sum_{i=1}^{\mathcal{B}}\nabla_{\bm\theta}\ell_{t,i}$;

Update the iteration number: $t\leftarrow t+1$;

}
\caption{MPTS for Model Agnostic Meta Learning (Few-Shot Scenarios)}
\label{maml_pseudo}
\end{algorithm}
\end{minipage}\hfill
\begin{minipage}[t]{0.485\textwidth}
\begin{algorithm}[H]
\SetAlgoLined
\SetKwInOut{Input}{Input}
\SetKwInOut{Output}{Output}
\Input{Task distribution $p(\tau)$;
 Task batch size $\mathcal{B}$;
 Candidate batch size $\hat{\mathcal{B}}$;
 Latest updated $\{\bm\psi,\bm\phi\}$;
 Latest history $H_{t-1}$;
 Iteration number $K$;
 Learning rate $\lambda_{2}$.}
\Output{Task identifier batch $\{\bm\tau_{t,i}\}_{i=1}^{\mathcal{B}}$.}

\tcp{\textcolor{red}{\textbf{Functional Posterior Inference}}}
\For{$i=1$ \KwTo $K$}
{   
Perform gradient updates given $H_{t-1}$:

$\bm\phi\leftarrow\bm\phi+\lambda_{2}\nabla_{\bm\phi}\mathcal{G}_{\text{ELBO}}(\bm\psi,\bm\phi)$ in Eq. (17b);

$\bm\psi\leftarrow\bm\psi+\lambda_{2}\nabla_{\bm\psi}\mathcal{G}_{\text{ELBO}}(\bm\psi,\bm\phi)$ in Eq. (17b);

}

\tcp{\textcolor{red}{\textbf{Simulating Adaptation Results}}}
Randomly sample $\{\bm\hat{\bm\tau}_{t,i}\}_{i=1}^{\hat{\mathcal{B}}}$ from $p(\tau)$;

Run amortized evaluation on candidate tasks $\{\delta_{i}:=\gamma_{0}m(\ell_i)+\gamma_{1}\sigma(\ell_i)\}_{i=1}^{\hat{\mathcal{B}}}$ in Eq. (19);

\tcp{\textcolor{red}{\textbf{Active Subset Selection from Predicted Results}}}
Rank $\{\delta_{i}\}_{i=1}^{\hat{\mathcal{B}}}$ and screen Top-$\mathcal{B}$ values;

Return the screened task batch $\{\bm\tau_{t,i}\}_{i=1}^{\mathcal{B}}$.

\caption{Model Predictive Task Sampling}
\label{few_cts_pseudo}
\end{algorithm}
\end{minipage}
\end{figure}

\begin{figure}[!htbp]
\noindent\begin{minipage}[t]{0.485\textwidth}
\begin{algorithm}[H]
\SetAlgoLined
\SetKwInOut{Input}{Input}
\SetKwInOut{Output}{Output}
\Input{Task distribution $p(\bm x)$;
 Task batch size $\mathcal{B}$;
 Learning rate $\lambda_1$.}
\Output{Fine-tuned machine learner $\bm\theta$.}
Set the initial iteration number $t=1$;

Randomly initialize machine learner $\bm\theta$;

Randomly initialize RPM $\{\bm\psi,\bm\phi\}$;

\While{not converged}{
Execute \textbf{Algorithm} \ref{ft_cts_pseudo} to access the batch $\{\bm\tau_{t,i}\}_{i=1}^{\mathcal{B}}$ and $\{[\bm x_{t,i},\bm y_{t,i}]\}_{i=1}^{\mathcal{B}}$;

\tcp{\textcolor{blue}{\textbf{Eval Adaptation Performance}}}
Compute the instance-specific adaptation risk $\{\ell_{t,i}:=\ell([\bm x_{t,i},\bm y_{t,i}];\bm\theta_{t})\}_{i=1}^{\mathcal{B}}$;

Return $H_{t}=\{[\bm\tau_{t,i},\ell_{t,i}]\}_{i=1}^{\mathcal{B}}$ as the Input to \textbf{Algorithm} \ref{ft_cts_pseudo};

\tcp{\textcolor{blue}{\textbf{Update Machine Learner}}}
Perform batch gradient updates:

$\bm\theta_{t+1}\leftarrow\bm\theta_{t}-\frac{\lambda_1}{\mathcal{B}}\sum_{i=1}^{\mathcal{B}}\nabla_{\bm\theta}\ell_{t,i}$;

Update the iteration number: $t\leftarrow t+1$;

}
\caption{MPTS for Pretrained Model Finetuning}
\label{ft_pseudo}
\end{algorithm}
\end{minipage}\hfill
\begin{minipage}[t]{0.485\textwidth}
\begin{algorithm}[H]
\SetAlgoLined
\SetKwInOut{Input}{Input}
\SetKwInOut{Output}{Output}
\Input{Offline processed $\bm\tau$ dataset;
 Task batch size $\mathcal{B}$;
 Candidate batch size $\hat{\mathcal{B}}$;
 Latest updated $\{\bm\psi,\bm\phi\}$;
 Latest history $H_{t-1}$;
 Iteration number $K$;
 Learning rate $\lambda_{2}$.}
\Output{Task identifier batch $\{\bm\tau_{t,i}\}_{i=1}^{\mathcal{B}}$.}

\tcp{\textcolor{red}{\textbf{Functional Posterior Inference}}}
\For{$i=1$ \KwTo $K$}
{   
Perform gradient updates given $H_{t-1}$:

$\bm\phi\leftarrow\bm\phi+\lambda_{2}\nabla_{\bm\phi}\mathcal{G}_{\text{ELBO}}(\bm\psi,\bm\phi)$ in Eq. (17b);

$\bm\psi\leftarrow\bm\psi+\lambda_{2}\nabla_{\bm\psi}\mathcal{G}_{\text{ELBO}}(\bm\psi,\bm\phi)$ in Eq. (17b);

}

\tcp{\textcolor{red}{\textbf{Simulating Adaptation Results}}}
Randomly sample $\{\bm\hat{\bm\tau}_{t,i}\}_{i=1}^{\hat{\mathcal{B}}}$ from $p(\tau)$;

Run amortized evaluation on candidate tasks $\{\delta_{i}:=\gamma_{0}m(\ell_i)+\gamma_{1}\sigma(\ell_i)\}_{i=1}^{\hat{\mathcal{B}}}$ in Eq. (19);

Rank $\{\delta_{i}\}_{i=1}^{\hat{\mathcal{B}}}$ and screen Top-$\mathcal{B}$ values;

\tcp{\textcolor{red}{\textbf{Exact Evaluation or Active Annotations}}}
Return the screened batch $\{[\bm x_{t,i},\bm y_{t,i}]\}_{i=1}^{\mathcal{B}}$.

\caption{Model Predictive Task Sampling}
\label{ft_cts_pseudo}
\end{algorithm}
\end{minipage}
\end{figure}

\begin{figure}[htbp]
\begin{center}
\centerline{\includegraphics[width=0.65\textwidth]{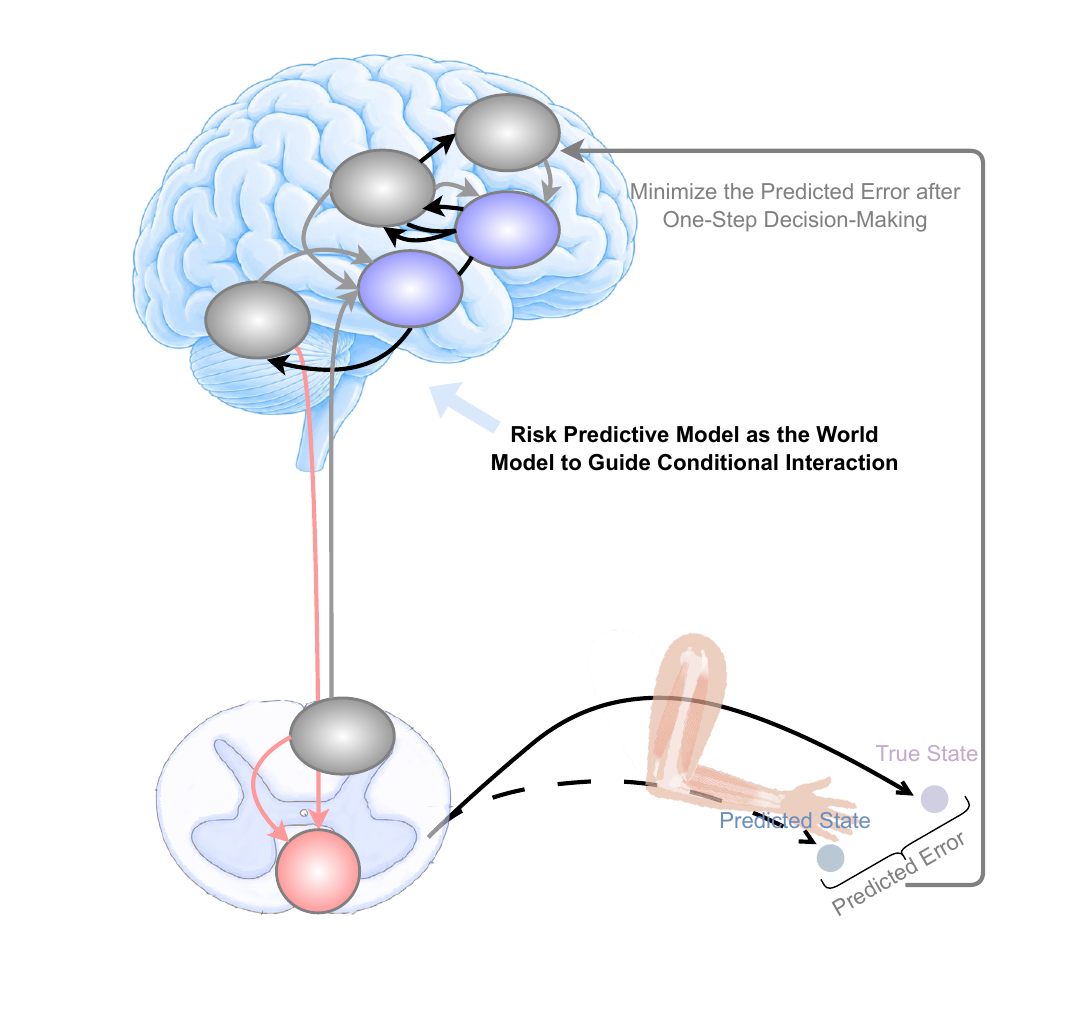}}
\caption{\textbf{Risk Predictive Model as the World Model to Guide Conditional Interaction.}
This shows the explanation of MPTS in the presence of active inference theory.
}
\label{fig_rpm_world_model}
\end{center}
\end{figure}

\begin{figure}[htbp]
\begin{center}
\centerline{\includegraphics[width=0.65\textwidth]{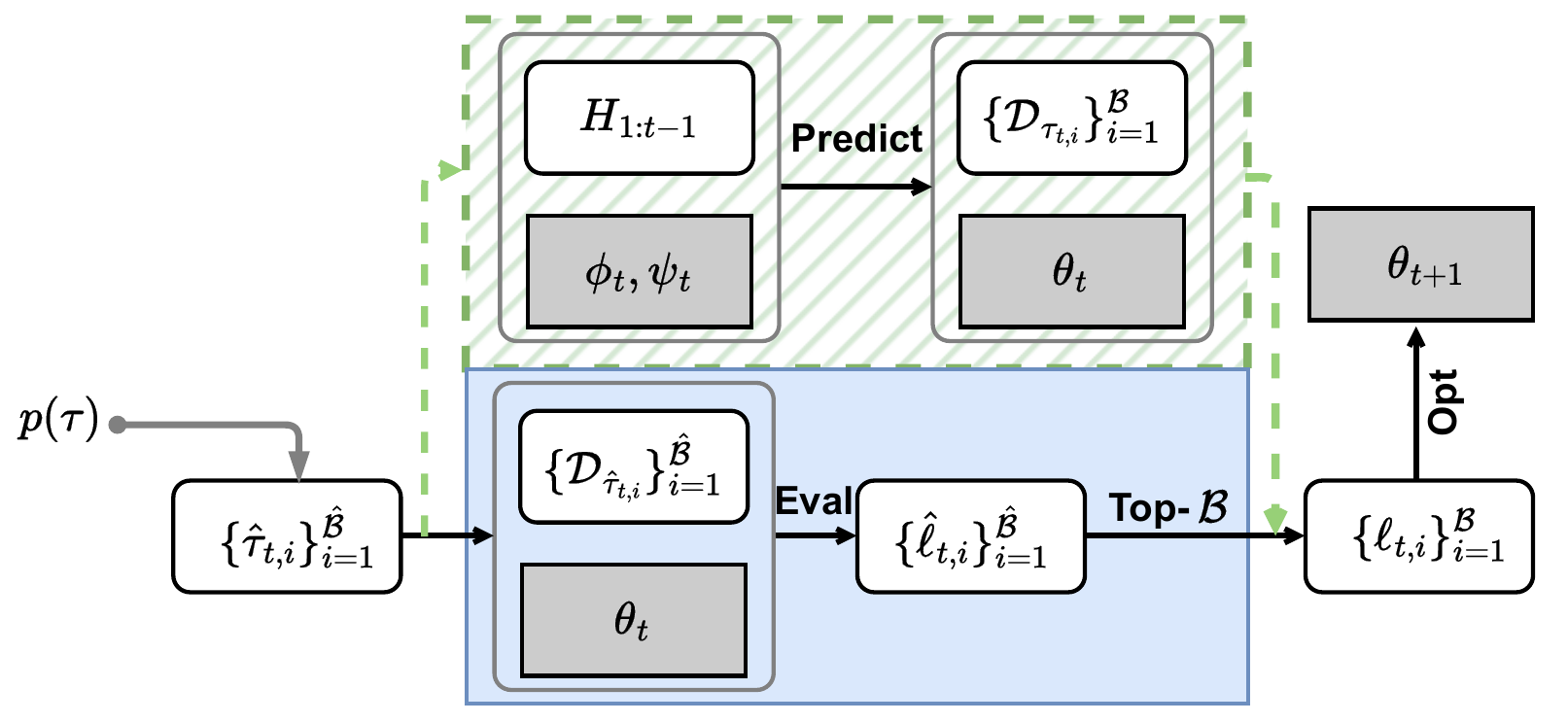}}
\caption{\textbf{Risk Predictive Module in MPTS for Active Subset Selection.}
MPTS adopts a \textit{predict-then-optimize} strategy and uses a predictive module in green to approximately score the task subset difficulty and obtain the preferred task subset. 
While the traditional method in blue exhausts $\hat{\mathcal{B}}$ tasks in construction and evaluation to filter preferred subset.
}
\label{fig_pred_then_opt}
\end{center}
\end{figure}

\begin{figure}[htbp]
\begin{center}
\centerline{\includegraphics[width=0.65\textwidth]{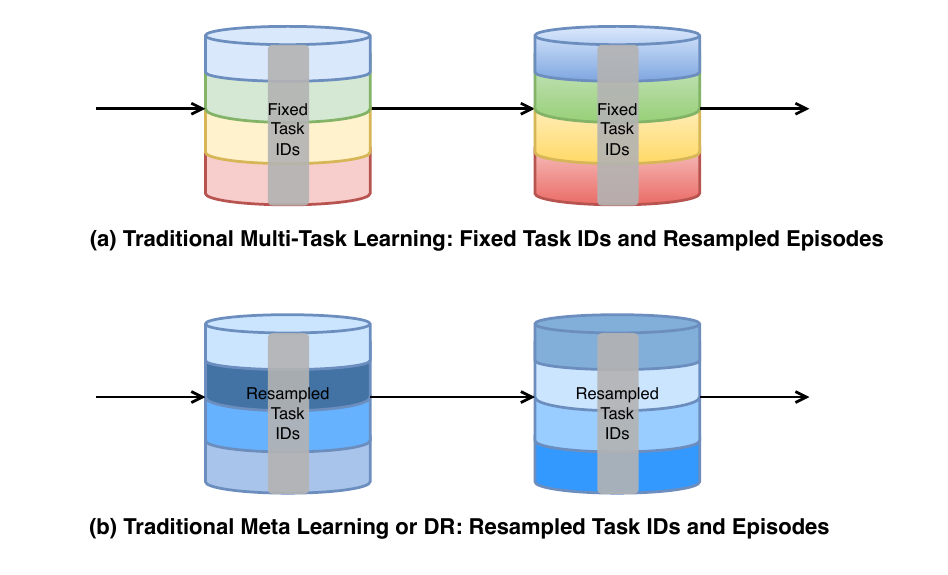}}
\caption{\textbf{Comparison between Traditional Multi-Task Learning and Meta Learning and Domain Randomization.}
Traditional multi-task learning aims to exploit the correlation between heterogeneous tasks, and task IDs in a batch do not change \citep{he2024robust}.
In comparison, meta learning and domain randomization typically consider the homogeneous task family and resample task IDs in a batch to optimize.
The latter is our studied scenario, referred to as task episodic optimization.
[Same types of colors mean the homogeneous tasks with depth to denote the various episode.]
}
\label{fig_mto_vs_meta}
\end{center}
\end{figure}

\paragraph{MPTS to Enable Conditional Interactions.}
This work introduces the concept of conditional interaction to improve either sample or computational efficiency.
As illustrated in Fig. \ref{fig_rpm_world_model}, MPTS aligns conceptually with the Bayesian brain, world model, and active inference perspectives \citep{friston2016active}, in which intelligent systems construct internal generative models of the environment to mentally simulate the consequences of candidate actions before physically interacting with the real world. 
Instead of exhaustively evaluating all tasks through costly environment interactions or forward computations, MPTS employs a lightweight risk predictive model as a task-level world model that forecasts the expected adaptation outcomes of candidate tasks and actively selects those predicted to maximize robustness improvement. 
After actual optimization and evaluation are performed, the discrepancy between predicted and observed adaptation risks is incorporated into posterior updates of the predictive model.
This resembles the prediction-error minimization and belief-updating mechanisms that underlie active inference and predictive coding theories in neuroscience.

\paragraph{Computational Complexity Analysis.}
The involvement of the RPM inevitably brings extra computational overhead in optimization. 
However, the RPM used in this work is lightweight with the model complexity $\mathcal{O}(|\bm\phi|+|\bm\psi|)<<\mathcal{O}(|\bm\theta|)$.
We can roughly estimate these extra computations that arise from the predictive model as $\mathcal{O}\big((|\bm\phi|+|\bm\psi|)T_{\text{MPTS}}\big)$ throughout the training phase.
Moreover, the computational and task evaluation complexities of different methods are estimated in Table~\ref{table_comput_complexity}.
Compared with DRM, MPTS retains more computational and task efficiency when the filtering ratio $\hat{\alpha}$ is high, and the machine learner $\bm\theta$ is largely given similar convergence iteration steps. 
OHTM \citep{kumar2023effect} uses the risk buffer to store historical difficult tasks and mixes it into the instant evaluated ones, hence, there is no additional computational cost.
As for DATS \citep{toloubidokhti2023dats} and TDPS \citep{wang2024towards}, they rely on additional gradient updates to compute the weights and raises more computational cost.

\begingroup
\setlength{\tabcolsep}{5.0pt}
\begin{table}[htbp]
   \begin{center}
    \caption{\textbf{Computational Complexities using Different Methods.}
    Here, we drop out the ranking or reweighting computational complexity as the model complexity of the machine learner considered in this analysis is major, such as the multimodal foundation models.
    $T$ refers to the required iteration steps until the convergence for separate methods.  
    }
    \label{table_comput_complexity}
    \begin{tabular}{|c|c|c|c|c|c|}
      \toprule %
       & ERM & DRM & GDRM & OHTM & MPTS (Ours)\\
      \toprule %
      computation & $\mathcal{O}(|\bm\theta|T_{\text{ERM}})$ & $\mathcal{O}(\frac{1}{1-\hat{\alpha}}|\bm\theta|T_{\text{DRM}})$ & $\mathcal{O}(|\bm\theta|T_{\text{GDRM}})$ & $\mathcal{O}(|\bm\theta|T_{\text{OHTM}})$ &$\mathcal{O}\big((|\bm\phi|+|\bm\psi|+|\bm\theta|)T_{\text{MPTS}}\big)$ \\
      \midrule %
      task eval & $\mathcal{O}(\mathcal{B}T_{\text{ERM}})$ & $\mathcal{O}(\frac{\mathcal{B}}{1-\hat{\alpha}}T_{\text{DRM}})$ & $\mathcal{O}(\mathcal{B}T_{\text{GDRM}})$ &$\mathcal{O}(\mathcal{B}T_{\text{OHTM}})$ &$\mathcal{O}(\mathcal{B}T_{\text{MPTS}})$ \\
      \bottomrule %
    \end{tabular}
  \end{center}
\end{table}
\endgroup

\paragraph{Choice of Surrogate Models.}
Among MPTS's core components, the RPM works to predict the adaptation risk values based on historical information and further serves the calculation of acquisition functions.
Importantly, this work investigates the feasibility and effectiveness of risk predictive strategies and does not impose rigid constraints on the form of the RPM $ p(\ell\vert\bm\tau,H_{1:t};\bm\theta)$ too much in modeling.
The design of this RPM $ p(\ell\vert\bm\tau,H_{1:t};\bm\theta)$ must meet several criteria: it is tractable in optimization, can process historical risk information, and offers uncertainty in prediction.

A series of candidate probabilistic models exist that probably apply to adaptation risk modeling.
One alternative choice can be the Gaussian process \citep{damianou2013deep}, which provides an analytical form of the predictive distribution.
However, its implementation (i) is less scalable in the case of relatively higher dimensional task identifiers, (ii) holds the cubic runtime complexity in obtaining the predictive covariance matrix, (iii) is sensitive to kernel selection, coupled with limited expressiveness of the Gaussian distribution in learned risk functions.
Hence, for simplicity and computational efficiency, we adopt the basic VAE-like model and execute a handful of gradient updates to train the RPM.
We leave more advanced RPM modeling for future exploration.

\paragraph{Bayesian Optimization for Black-box Functions.}
This work relates to active sampling and Bayesian optimization.
The purpose of BO \citep{snoek2012practical} is to sequentially find a global optimum of a black-box function $f(\bm x)$ expensive to evaluate in $\mathcal{S}$, namely $\bm x_*=\arg\max_{\bm x\in\mathcal{S}\subset\mathbb{R}^{d}} f(\bm x)$.

In each iteration $t=1,\dots,T$, the BO method actively queries $\bm x_t$ to evaluate $f(\bm x_t)$, yeilding an output $\ell_t=f(\bm x_t)+\epsilon$ with a white noise $\epsilon\sim\mathcal{N}(0,\sigma^2)$. 
Due to the high cost of function evaluation, the key to BO is constructing a surrogate model to guide the data point to query.
The resulting acquisition function \citep{wilson2018maximizing} works as an active sampling objective to maximize and obtain the candidate $\bm x_t$ based on the previous sequence.
BO requires limited function evaluations as observations and exploits the correlations in queried data points.
These properties make it more theoretically data efficient than random or grid search in seeking the optimal solution \citep{garnett2023bayesian}.
This work differs from standard BO as task episodic learning is not the optimal parameter search problem.

\paragraph{Specific Pseudo Algorithms in Considered Scenarios.}
The main text provides the workflow of MPTS in Algorithm~\ref{alg_mpts}.
For separate scenarios, we attach detailed pseudo algorithms as follows.
These illustrated Algorithms are in the context of supervised learning.
Regarding RL scenarios, such as meta RL and DR, there is a slight modification for MPTS.
As simply picking up worst-case MDPs restricts the task subspace in optimization \citep{greenberg2024train}, we adopt the mixture of the identifier subset from the random sampler and the identifier subset from the MPTS sampler.
For example, in meta RL, with the pseudo batch size $\hat{\mathcal{B}}=1.5\mathcal{B}$, there $1.5\mathcal{B}$ identifier candidates from the random sampler.
We retain $0.5\mathcal{B}$ random ones and execute standard MPTS amortized evaluation and acquisition rule to obtain another $0.5\mathcal{B}$ ones from the rest random $\mathcal{B}$ identifiers, formulating the mixed $\mathcal{B}$ task batch for RL training.
Such an operation makes RL over the MDP distribution stable in optimization.
See the open-source code for more RL details.

\paragraph{Application Scope.}
MPTS considers the task distribution to conduct robust optimization, where tasks are homogeneous in implementations.
The entire optimization pipeline adheres to task-episodic training, which means that a batch of tasks is resampled to train in each iteration.
However, for classical multi-task learning scenarios, when tasks are fixed and heterogeneous, the task identity of the batch is not changed over iterations.
Such a difference can be illustrated in Fig.~\ref{fig_mto_vs_meta}.
Hence, vanilla MPTS might encounter the applicability issue.
We leave developing model predictive strategies to balance the weight of tasks for robust optimization as future work.

\section{Research Background}\label{sec_supp_liter}

\subsection{Adaptation Learning for Cross-Task Generalization}

Learning from zero-shot or few-shot examples has been identified as a crucial adaptation capability of the machine learner nowadays \citep{wang2020generalizing,wang2023large}.
In SFT, this work treats the individual example as each task to meet MPTS setup.
As SFT techniques have been widely discussed in the field \citep{ding2023parameter}, we skip this part in the background introduction.

\paragraph{Zero-Shot Adaptation.}
This assesses the machine learner's generalization capability when directly deploying in unseen scenarios without the help of a support dataset. 
Such a cross-task generalization is commonly studied in computer vision \citep{xu2022attribute}, and the core of the relevant methods is effective semantic representation either from embedding-based methods \citep{liu2018generalized,li2020transferrable,keshari2020generalized} or generative-based methods \citep{xian2018feature,schonfeld2019generalized}. 
In the era of the foundation models, the pretraining mechanism between multimodality also sometimes empowers the machine learner, such as CLIP \citep{radford2021learning}, with zero-shot capability.
When it comes to sequential decision-making, a commonly seen method is DR \citep{tobin2017domain,mehta2020active}, which places a distribution over environments for the agent to interact.

\paragraph{Few-Shot Adaptation.}
This examines the machine learner's capability of resolving unseen tasks from some annotated examples as hints.
Meta-learning, as the typical learning paradigm, has gained popularity over the past decade.
It achieves few-shot adaptation by leveraging past experience and distilling knowledge to unseen but similar scenarios in a few-shot way \citep{hospedales2021meta}.
In brief, we categorize commonly seen methods into context-based, optimization-based, geometric-based, and others. 
(i) Formulated in an encoder-decoder structure, the context-based method resembles variational autoencoders and encodes the few-shot information into latent variables or embeddings.
Typical ones are neural process families \citep{garnelo2018neural,wang2022bridge,gondal2021function,garnelo2018conditional}, which aim to constitute exchangeable deep stochastic processes with neural networks.
(ii) The optimization-based methods, with their versatile nature and ability to enable cross-task skill transfer, have piqued the interest and engagement of researchers in the field.
For example, MAML \citep{finn2017model,finn2018probabilistic,abbas2022sharp,rajeswaran2019meta} reduces meta-learning to a bi-level optimization in the parameter space, and its extensions have been widely investigated in the field.
(iii) The deep metric-based methods \citep{snell2017prototypical,allen2019infinite} attempt to embed tasks into the latent space and are more suitable for few-shot image classification tasks.
Besides, there are other families, such as hyper-networks \citep{ha2016hypernetworks,sendera2023hypershot}, recurrent meta-learning \citep{duan2016rl}, etc.

\subsection{Dataset Curation and Task-Level Robustness}

\paragraph{Task Curation in Robust Adaptation Learning Pipelines.}
Recent works \citep{sorscher2022beyond,evans2023bad} demonstrate the effectiveness of challenging task prioritization over uniform sampling in improving cross-task generalization and adaptation robustness, particularly when the learning dataset is sufficiently large. 
Many methods \citep{rajeswaran2022epopt,sorscher2022beyond,evans2023bad,wang2024simple,dubey2024llama,greenberg2024train,evans2024data} adopt an Evaluate-Rank-Filter step for iterative model updates, introducing a batch filtering ratio $\hat{\alpha}=1-\frac{\mathcal{B}}{\hat{\mathcal{B}}}\in[0,1)$ to quantify the fraction of discarded tasks in a sample batch. 
This prioritization of "difficult" tasks aligns with minimizing $\text{CVaR}_{\alpha}$ \citep{rockafellar2000optimization}, a robustness metric for tail-case performance. 
Alternatively, other methods \citep{sagawa2019distributionally,xie2024doremi,hejnaremix} focus on constructing uncertainty sets and reweighting tasks within the batch to achieve robust adaptation. 
Additionally, coreset methods \citep{mirzasoleiman2020coresets,yang2023towards,joshi2023data} aim to select a small subset of tasks that effectively represent the utility of the full dataset, often through gradient approximation in optimization. 
These approaches address a subproblem of data efficiency, with the acquisition strategy in MPTS serving as an episodic coreset selection mechanism tailored for robustness.

\paragraph{Task Distributional Robustness.}
The $\text{CVaR}_{\alpha}$ or expected shortfall \citep{rockafellar2000optimization} is a statistical measure to assess the proportional worst-case performance of some models at certain levels.
This is widely adopted in risk-averse applications. 
As implied in Definition~\ref{def_cvar}, $\text{CVaR}_{\alpha}$ describes the expected risk under the normalized $(1-\alpha)$ proportional tail risk task distribution, and this work specifies the distribution in the task space.
Meanwhile, the normalized tail task distribution $p_{\alpha}(\tau;\bm\theta)$ can be viewed as a shifted result from the initial task distribution $p(\tau)$; hence, such a measure provides robustness quantification in the presence of task distribution shifts \citep{wang2024simple,lv2024theoretical,greenberg2024train}.

Another indicator to evaluate the machine learner's robustness is the performance in OOD tasks.
This refers to the case when the training and the testing task distributions are different.
Particularly, in DR and prompt-tuning scenarios, we also use the OOD tasks that never appear in the training task distribution to test the trained policy, and this setup corresponds to domain generalization, a type of substantial distribution shift \citep{koh2021wilds}. 

Similar to the setup \citep{wang2024simple}, let $(\Omega_{\tau},\mathcal{F}_{\tau},\mathbb{P}_{\tau})$ be a probability space over tasks, where $\mathcal{F}_{\tau}$ is a $\sigma$-algebra on subsets of $\Omega_{\tau}$.
Consider $(\mathbb{R},\mathbb{B})$ with $\mathbb{B}$ the Borel $\sigma$-algebra, defining a probability measure for the adaptation risk function $\ell(\mathcal{D}_{\tau};\bm\theta)$.
For $\bm\theta\in\bm\Theta$, the adaptive optimization operator is
$\mathcal{M}_{\bm\theta}:\tau\mapsto\ell(\mathcal{D}_{\tau};\bm\theta)$.

Thus $\ell(\cdot)$ acts as a random variable to induce the risk distribution $p(\ell)$. 
The corresponding cumulative distribution is
$F_{\ell}(\ell;\bm\theta)=\mathbb{P}(\{\ell(\mathcal{D}_{\tau};\bm\theta)\leq \ell;\tau\in\Omega_{\tau},\ell\in\mathbb{R}\})$.
Note that $F(\ell;\bm\theta)$ depends on $\bm\theta$ and generally lacks a closed form.
The resulting probability density function and the following explanation can be found in Definition~\ref{def_cvar} of the main text.

\paragraph{Sample Average Approximation of CVaR.}
Note that the vanilla optimization objective of CVaR can be expressed in the form of dual representation, which corresponds to:
\begin{equation}
    \min_{\bm\theta\in\bm\Theta,\zeta\in\mathbb{R}}\text{CVaR}_{\alpha}(\bm\theta):=\zeta+\frac{1}{1-\alpha}\mathbb{E}_{p(\tau)}
        \Big[[\ell(\mathcal{D}_{\tau}^{Q},\mathcal{D}_{\tau}^{S};\bm\theta)-\zeta]^{+}
        \Big],
\end{equation}
where the signed function means $[\ell(\mathcal{D}_{\tau}^{Q},\mathcal{D}_{\tau}^{S};\bm\theta)-\zeta]^{+}=\max\{\ell(\mathcal{D}_{\tau}^{Q},\mathcal{D}_{\tau}^{S};\bm\theta)-\zeta,0\}$.
With the help of sample average approximation in Monte Carlo, this can be further written as:
\begin{equation}\label{eq_cvar_mc}
    \begin{split}
        \min_{\bm\theta\in\bm\Theta,\zeta\in\mathbb{R}}\text{CVaR}_{\alpha}(\bm\theta):=\zeta+\frac{1}{(1-\alpha)\hat{\mathcal{B}}}\sum_{i=1}^{\hat{\mathcal{B}}}[\ell(\mathcal{D}_{\hat{\tau}_{i}}^{Q},\mathcal{D}_{\hat{\tau}_{i}}^{S};\bm\theta)-\zeta]^{+}.
    \end{split}
\end{equation}
Since the optimality for the auxiliary variable holds at the condition $\zeta=\text{VaR}_{\alpha}(\bm\theta)$, selecting the Top-$\mathcal{B}$ element in the set $\{\ell_i\vert\ell_i=\ell(\mathcal{D}_{\hat{\tau}_{i}}^{Q},\mathcal{D}_{\hat{\tau}_{i}}^{S};\bm\theta)\}_{i=1}^{\hat{\mathcal{B}}}$ is an unbiased Monte Carlo estimate of CVaR.
The gradient of Eq. (\ref{eq_cvar_mc}) is the steepest direction in gradient optimization, and executing gradient descent decreases the CVaR value. 
In addition, the adaptation risk in this work $\ell(\mathcal{D}_{\hat{\tau}_{i}}^{Q},\mathcal{D}_{\hat{\tau}_{i}}^{S};\bm\theta)$, e.g., classification accuracies or MSEs, is typically bounded.
With the monotonic improvement and bounded risk function values, optimizing the Monte Carlo CVaR (MC-CVaR) leads to convergence.
Such a robust optimization method corresponds to the baseline DRM \citep{wang2024simple} in this work.

\subsection{Risk Minimization Principles and Prioritized Sampling}\label{sebsec_risk_min_principle}

The risk minimization principles are entangled with task sampling and robust optimization.

\paragraph{Expected/Empirical Risk Minimization (ERM).}
With the fixed $p(\tau)$, the principle follows the statistical learning theory \citep{vapnik1998statistical} and minimizes the expectation of adaptation risk over the task space.
As a result, we can have:
\begin{equation}\label{erm}
\min_{\bm\theta\in\bm\Theta}\mathbb{E}_{p(\tau)}
        \Big[\ell(\mathcal{D}_{\tau}^{Q},\mathcal{D}_{\tau}^{S};\bm\theta)
        \Big].
\end{equation}
It draws batches with a random task sampler to optimize iteratively.

\paragraph{Distributionally Robust Risk Minimization (DRM) \citep{rajeswaran2022epopt,evans2023bad,greenberg2024train,wang2024simple,lv2024theoretical}.}
We retain the notation of task robust work \citep{wang2024simple}, which terms the tail task risk minimization as DRM.
It aims to improve the robustness of adaptation to the tail tasks over iteration.
No explicit form exists as the tail task distribution is $\bm\theta$-dependent.
The optimization objective is derived as the $\text{CVaR}_{\alpha}(\bm\theta)$ \citep{rockafellar2000optimization}:
\begin{equation}\label{tail_rm}
\min_{\bm\theta\in\bm\Theta}\text{CVaR}_{\alpha}(\bm\theta):=\mathbb{E}_{p_{\alpha}(\tau;\bm\theta)}
        \Big[\ell(\mathcal{D}_{\tau}^{Q},\mathcal{D}_{\tau}^{S};\bm\theta)
        \Big],
\end{equation}
where we write $p_{\alpha}(\tau;\bm\theta)$ to express the $(1-\alpha)$ proportional worst case for easier formulation.
In other words, $\mathbb{E}_{p_{\alpha}(\tau;\bm\theta)}\Big[\ell(\mathcal{D}_{\tau}^{Q},\mathcal{D}_{\tau}^{S};\bm\theta)\Big]$ also relates to the task distribution with constraints.
Also note that when $\alpha$ approaches $1$, the problem degenerates to the worst-case risk minimization.

This work retains the setup in work \citep{wang2024simple} and picks up the Top-$\mathcal{B}$ in optimization, which corresponds to sample average Monte Carlo of $\text{CVaR}_{\alpha}$.
This implies that the actual task batch to evaluate is $\frac{\mathcal{B}}{1-\alpha}$.
And for fair comparison with MPTS and light computations, we retain the Monte Carlo estimator for the risk quantile in implementation.
To ensure stable training, in all benchmarks, we keep the actual task batch $\hat{\mathcal{B}}=2\mathcal{B}$ to evaluate and discard the easiest half before the machine learner's optimization.

\paragraph{Group Distributionally Robust Risk Minimization (GDRM) \citep{sagawa2019distributionally}.}
This can be interpreted as a min-max optimization problem.
Such a principle \citep{sagawa2019distributionally} effectively improves robustness in distribution shifts and has shown positive effects on training foundation models \citep{xie2024doremi,hejnaremix}.
It constructs a collection of uncertainty sets over tasks and results in the optimization objective as follows:
\begin{equation}\label{eq_gdro}
    \begin{split}
        \min_{\bm\theta\in\bm\Theta}\sup_{g\in\mathcal{G}}\mathbb{E}_{p_{g}(\tau)}
        \Big[\ell(\mathcal{D}_{\tau}^{Q},\mathcal{D}_{\tau}^{S};\bm\theta)
        \Big],
    \end{split}
\end{equation}
where $\mathcal{G}$ are groups of uncertainty sets, and $p_{g}(\tau)$ indicates the probability measure over the task group.
The operation inside Eq. (\ref{eq_gdro}) prioritizes the worst group to optimize in a soft way.

GDRM increases the machine learner's robustness by assigning more probability mass to worst cases in a reweighted manner.
That means in each iteration with the best selected $p_{\hat{g}}(\tau)$, the optimization problem is reduced to
\begin{equation}
    \begin{split}
        \min_{\bm\theta\in\bm\Theta}\mathbb{E}_{p_{\hat{g}}(\tau)}
        \Big[\ell(\mathcal{D}_{\tau}^{Q},\mathcal{D}_{\tau}^{S};\bm\theta)
        \Big]
        =\mathbb{E}_{p(\tau)}\left[\frac{p_{\hat{g}}(\tau)}{p(\tau)}\ell(\mathcal{D}_{\tau}^{Q},\mathcal{D}_{\tau}^{S};\bm\theta)\right],
    \end{split}
\end{equation}
where we use $\omega(\tau)=\frac{p_{\hat{g}}(\tau)}{p(\tau)}$ to denote the weight.

Given a fixed number of tasks, GDRM will heuristically or dynamically group them into clusters and then perform a reweighting mechanism according to the evaluated risk.
In task episodic learning, there is no task grouping operation as the task batch is reset after each iteration.
And the default computation of task-specific weights is $\omega(\tau_{i})=\frac{\exp(\eta\ell(\mathcal{D}_{\tau_{i}}^{Q},\mathcal{D}_{\tau_{i}}^{S};\bm\theta))}{\sum_{b=1}^{\mathcal{B}}\exp(\eta\ell(\mathcal{D}_{\tau_{b}}^{Q},\mathcal{D}_{\tau_{b}}^{S};\bm\theta))}$, where $\eta$ is the temperature parameter and $\{\bm\tau_{b}\}_{b=1}^{\mathcal{B}}$ is the identifier of the task batch.
The implementation detail can be found in \url{https://github.com/kohpangwei/group_DRO}.

As revealed in works \citep{oren2019distributionally,collins2020task,wang2024simple}, the heuristic operation as the Evaluate-Rank-Filter or reweighting mechanism in GDRM is widely adopted for approximate optimization.
For example, in task robust meta-learning scenarios, the prerequisite step in DR-MAML \citep{wang2024simple} is to execute gradient updates in the inner loop for all candidate tasks and then screen the tail task subset to meta-optimize according to the evaluation results.

\paragraph{Difficulty-Aware Task Sampler (DATS) \citep{toloubidokhti2023dats}.}
Such a baseline is originally developed for few-shot learning problems of physics-informed neural networks (PINNs).
Here, we modify it to satisfy the studied benchmarks in few-shot learning scenarios.
The basic idea of scoring task difficulty is to compute the inner product of the task-specific gradient on the support dataset $\bm g_{t,i}:=\nabla_{\bm\theta}\ell(\mathcal{D}_{\tau_{t,i}}^{S};\bm\theta_{t})$ and the task-average gradient on the query dataset of all tasks $\bar{\bm g_t}:=\frac{1}{\mathcal{B}}\sum_{j=1}^{\mathcal{B}}\nabla_{\bm\theta}\ell(\mathcal{D}_{\tau_{t,j}}^{Q};\bm\theta_{t})$ in the batch.

In implementation, it computes the gradient inner product $s_{t,i}=<\bm g_{t,i},\bar{\bm g_t}>$ and then perform normalization as $\omega(\tau_{i})=\frac{\exp(\eta s_{t,i})}{\sum_{i=1}^{\mathcal{B}}\exp(\eta s_{t,i})}$.
The explanation for such a task weighting mechanism lies in the fact that higher $s_{t,i}$ reveals a more consistent gradient update direction to reduce the validation loss across all tasks, while lower $s_{t,i}$ probably encounter the conflicting gradient issue.
Hence, DATS places higher weight on higher $s_{t,i}$.
And the vanilla implementation involves the coefficient over the exponential term, and the default is $\frac{1}{\mathcal{B}}$ for all tasks, i.e., the uniform-KL regularization.

\paragraph{Online Hard Task Mining Sampler (OHTM) \citep{kumar2023effect}.}
The OHTM strategy selects the most challenging tasks from the set of tasks already encountered. 
To adapt OHTM for meta-learning, \citet{kumar2023effect} implements a hybrid scheme: half of each meta-batch is drawn using the OHTM sampler, while the other half is chosen uniformly at random. 
This design ensures diversity by including a broad range of tasks rather than limiting training to previously observed ones.

\paragraph{Task difficulty Prioritized Sampler (TDPS) \citep{wang2024towards}.} 
The vanilla Adaptive Sampler \citep{wang2024towards} is primarily designed to improve the overall generalization of meta-learning methods, identifying task difficulty, task diversity, and task entropy as key factors for acquiring high-quality tasks. 
It integrates multiple prioritized sampling criteria into a combined metric. 
To align with the objective of robust optimization and achieve fair comparison, we adopt only the difficulty-prioritized module of TDPS as our baseline and use the difficulty-aware module. 
The core idea is to compute gradients on both the support and query sets and quantify their discrepancy, where the gradient inconsistency within $\mathcal{D}_{\tau}$ serves as a proxy for task difficulty in optimization.

\paragraph{Other Representative Sampling Strategies.}
Beyond the baselines adopted in this study, other domain-specific task sampling strategies have been proposed with distinct optimization objectives. 
For instance, \citet{liu2020adaptive} introduce greedy class-pair (GCP) sampling, which emphasizes class-based adaptive sampling for few-shot image classification by actively constructing challenging tasks for meta-learning. 
In contrast, our setting operates at the instance level without altering the task construction pipeline. 
Another representative method, probabilistic active meta-learning (PAML) \citep{kaddour2020probabilistic}, aims to improve data efficiency by inferring task embeddings to quantify informativeness.
However, PAML is not designed for robust optimization and is more suitable for robotic system identification problems.
Adaptive task scheduler \citep{yao2021meta} also adopts the gradient inner product between the query and support dataset to reweigh task losses, and DATS and TDPS share a similar motivation.
Given this, we include the previously mentioned baselines in comparison.

\section{Task Construction and Identifiers}\label{sec_supp_task_identifier}
Here we refer to the variables that sufficiently configure a task as the task identifier $\bm\tau$.
In other literature work, these task identifiers can be viewed as the task representations in a lower dimensional space.
To clarify these concepts, we provide more explanations in specific scenarios.

\subsection{Tasks with Explicit Identifiers}

\paragraph{\texttt{K-shot} Sinusoid Regression.}
In this setup \citep{finn2017model},  meta learners aim at quickly adapting the model to an unseen function $f(x)=a\sin(x-b)$ with the help of $K$ data points randomly sampled from the function.
This case treats the amplitude and phase variables $(a,b)$ as the task identifier to configure the task.
And the task distribution is induced by the uniform distribution over the task identifier.

\paragraph{Meta Reinforcement Learning.}
Here, we take the ReacherPos task as an example.
The goal of the robot arm is to reach an unobserved target location $[x_1,x_2]$.
The end-effector position of the robot arm is initialized randomly, and the step-wise reward corresponds to the feedback to the agent after each move based on its distance to the target location.
As the task distribution is specified by a uniform distribution over the target location, $\bm\tau=[x_1,x_2]$ can be viewed as the task identifier.
Similarly, we vary physics parameters in simulators to generate diverse MDPs.
This constitutes different meta RL benchmarks.

\paragraph{Domain Randomization.}
DR is a promising paradigm to achieve zero-shot adaptation in unseen scenarios, which is widely adopted in robotics \citep{mehta2020active} and computer vision \citep{tobin2017domain}.
The basic idea is to train the machine learner in 
a distribution over environments and then directly apply the learned model to new ones.

\begingroup
\setlength{\tabcolsep}{16.0pt}
\begin{table}[htbp]
   \begin{center}
    \caption{\textbf{Benchmarks with Explicit Task Identifiers.}
    Here, we list the detail information about the task identifier to induce the task distribution.
    }
    \label{table_bench_explicit_identifier}
    \begin{tabular}{|c|c|c|}
      \toprule %
       Benchmarks & Identifier Meaning & Identifier Range\\
      \toprule %
      \texttt{K-shot} sinusoid regression & amplitude and phase $(a,b)$ & $[0.1,5.0]\times[0,\pi]$ \\
      \toprule %
      Meta-RL: HalfCheetahMassVel & mass and velocity $(m,v)$ & $[0.75,1.25]\times[0,2.0]$ \\
      Meta-RL: HalfCheetahVel & velocity $v$ & $[0,2.0]$ \\
      Meta-RL: ReacherPos & goal location $(x_1,x_2)$ & $[-0.2,0.2]\times[-0.2,0.2]$ \\
      Meta-RL: Walker2dMassVel & mass and velocity $(m,v)$ & $[0.75,1.25]\times[0,2.0]$ \\
      Meta-RL: Walker2dVel & velocity $v$ & $[0,2.0]$ \\
      \toprule %
      DR: LunarLander & main engine strength $s$ & $[4,20]$ \\
      DR: ErgoReacher & joint damping $d$ and max torque $t$ ($\times$4 joints) & $[0.1,2.0]\times[2,20]$ \\
      \bottomrule %
    \end{tabular}
  \end{center}
\end{table}
\endgroup

As noted in the main text, we suppose that the task identifier contains semantics that reflects the difficulty of tasks to resolve and the adaptation risk function is smooth with respect to the identifier.
In total, we summarize these bechmarks with explicit task identifiers in Table~\ref{table_bench_explicit_identifier}.

\subsection{Tasks with Implicit Identifiers}

As previously mentioned, we assume the existence of a statistical correlation between task identifiers and adaptation risk values given a specific adaptive machine learner.
This implies that the task identifier preserves precise semantics about the task information.
These provide the basis for establishing the RPM from the coupled dataset $\{[\bm\tau_i,\ell_i]\}_{i=1}^{\mathcal{B}}$.

Nevertheless, in several scenarios, it is intractable to access the explicit task identifier.
For example, in few-shot image classification, the task information is just the coupled support and query dataset $\mathcal{D}_{\tau}=\mathcal{D}_{\tau}^{S}\cup\mathcal{D}_{\tau}^{Q}$.
Similarly, in SFT for LLMs, the task can be in the form of the QA pair $\mathcal{D}_{\tau}=\mathcal{D}_{\tau}^{Q}$.
There is no explicit representation method, such as $\bm\tau$, for these tasks, which brings difficulty in building up the RPM.
Retaining the prior notation, the episodic task batch can be written as $\hat{H}_{t}=\left\{\bm\theta_{t},\left(\bm\tau_{t,i},\mathcal{D}_{\tau_{t,i}},\ell_{t,i}\right)\right\}_{i=1}^{\mathcal{B}}$, where $\bm\tau$ of our interest is unobservable.
Some experiments in this work, such as few-shot image classification and SFT, encounter such circumstance.

\paragraph{Task Representation through Identifier Inference.}
To scale our approach under these circumstances, we propose an alternative candidate schema as the complementary.
The probabilistic relationship between variables is depicted in Fig.~\ref{fig_basic_setup}c.
We consider obtaining the implicit identifier through inference from the task dataset.
To do so, we include additional module $f_{\xi}$ with $\xi\in\Xi$ to embed $\mathcal{D}_{\tau}^{S}$ and $\mathcal{D}_{\tau}^{Q}$ and further induce a vector $\bm\tau=f_{\xi}(\mathcal{D}_{\tau}^{S},\mathcal{D}_{\tau}^{Q})$ as the approximate task identifier.
These operations imply seeking the appropriate inference module directly influences the RPM's performance.

Fortunately, there exist pretrained models that enable the task representation to be generalizable to downstream tasks.
For example, in the \textbf{\texttt{N-way K-shot}} image classification, the task is in the form of support image-label pairs and the query images and the goal is to assign labels to the query images from the support dataset.
With the help of CLIP models \citep{radford2021learning}, for a fixed task in the form of $\mathcal{D}_{\tau}$, we can access a \texttt{N} vectors $\{\bm z_i\}_{i=1}^{N}$ by inputing the set of text-form classes $\{\mathcal{C}_i\}_{i=1}^{N}$ extracted from the support dataset $\mathcal{D}_{\tau}^{S}$, i.e., $\text{CLIP}(\{\mathcal{C}_i\}_{i=1}^{N})=[\text{CLIP}_{\text{text}}(\mathcal{C}_1),\dots,\text{CLIP}_{\text{text}}(\mathcal{C}_K)]:=\bm\tau$.
As a result, we can obtain $H_{t}=\{[\bm\tau_{t,i},\ell_{t,i}]\}_{i=1}^{\mathcal{B}}$ conditioned on $\bm\theta_t$ for feasible task risk functional prosterior inference. 
This helps our approach to circumvent the unavailability of exact task identifiers.
And it is plausible for the RPM to optimize in learning $p(\ell\vert\bm\tau,H_{1:t})$.
It is worth noting that this case still prefers lightweight models for identifier inference, and the text encoder of CLIP well satisfies this requirement and can be used in the \textbf{\texttt{N-way K-shot}} image classification.
Details on specific task identifier inference modules can be found in Section \ref{sec_supp_pseudo_class} and \ref{sec_supp_backbone}.

\subsection{Scalability with Large Reasoning Models}

The rise of large reasoning models (LRMs) this year makes MPTS even more critical in cutting off policy evaluation cost, i.e., expensive agent-environment interactions.

\paragraph{Finite tasks and unobservable variables.}
The optimization of large reasoning models (LRMs) often requires massive rollouts to validate outcomes, such as in mathematical problem-solving, particularly under the reinforcement learning from verified reward (RLVR) paradigm. 
In this setting, the prompt in RLVR corresponds to the task in our framework, yet the task dataset is typically finite and lacks explicit identifiers during RL-based finetuning. 
A common workflow involves sampling a batch of prompts, generating multiple rollouts per prompt, and extracting informative signals for optimization. 
For instance, in verifiable mathematical problem-solving, the average success rate estimated from multiple rollouts constitutes an unobservable variable for MPTS to predict, as it reflects task difficulty.
The model predictive prompt selection (MoPPS) \citep{qu2025can}, a variant of MPTS tailored for LRMs, casts each prompt as a bandit problem. 
It leverages the 0–1 signals from rollout histories to perform online Bayesian inference over the success rate of each prompt. 
The inferred distribution then serves as a predictive prior for active prompt selection, avoiding direct interactions with the underlying LRM.

\paragraph{Compatibility with curriculum task selection.}
 In the context of RL finetuning for LRMs, MoPPS and related approaches demonstrate that curriculum task selection can be more effective than naïvely prioritizing task difficulty. 
Incorporating curriculum criteria accelerates finetuning and highlights the versatility of MPTS, which can be seamlessly integrated with loss-oriented task samplers.

\section{Auto-Encoding Adaptation Risk through Streaming VI}\label{sec_supp_formulation}

Note that the basis of MPTS is to establish the bridge between the task identifier and the adaptation risk value over the course of the machine learner's optimization.
In other words, we are seeking a lightweight stochastic risk function in  Definition S1 to approximate the posterior $p(\ell\vert\bm\tau,H_{1:t})$ in the task space.

\textit{Definition S1}
(Stochastic Risk Function)
Let $\mathfrak{X}$ denote the index set's Cartesian product with the task identifier's dimension $\bm\tau\in\mathbb{N}^{d}$.
For any $k\in\mathbb{N}$ and finite index sequence $\bm\tau_1,\dots,\bm\tau_k\in\mathfrak{X}$, we write some probability measure over $\mathbb{R}^{k}$ as $\nu_{(\bm\tau_1,\dots,\bm\tau_k)}$.
By introducing the probability space $(\Omega_{\tau},\mathcal{F}_{\bm\theta},\mathcal{P})$ and $\forall\theta\in\Theta$, we can induce a stochastic function $\mathcal{F}_{\bm\theta}:\mathcal{T}\times\Omega_{\tau}\mapsto\mathbb{R}^{k}$, so that $\nu_{(\bm\tau_1,\dots,\bm\tau_k)}(C_{1}\times\dots\times C_{k})=\mathcal{P}(\mathcal{F}_{\bm\theta}(\bm\tau_1)\in C_{1},\dots,\mathcal{F}_{\bm\theta}(\bm\tau_k)\in C_{k})$ $\forall\bm\tau_i\in\mathfrak{X}$ and $C_{i}\in\mathbb{R}$.

This section details steps in auto-encoding historical task risk information, parameterizing variational distributions, deriving the approximate optimization objective, and estimating the stochastic gradients of parameters.

\subsection{Neural Modules to Parameterize Distributions}

Here, we detail the neural modules to parameterize the distributions of interest.
For the approximate posterior $q_{\bm\phi}(\bm z_{t}\vert H_{t})$ and conditional prior $p(\bm z_{t}\vert H_{1:t-1})$, the inputs of the module are a set of task risk pairs.
The neural module requires the permutation invariance \textit{w.r.t.} the order of the data points in the set $H_t$ or $H_{1:t-1}$ in Definition~\ref{def_pif}.
Hence, we adopt the DeepSet style neural network \citep{zaheer2017deep} to process the collected $H_t$ or $H_{1:t-1}$.

\begin{figure}[htbp]
\begin{center}
\centerline{\includegraphics[width=0.85\textwidth]{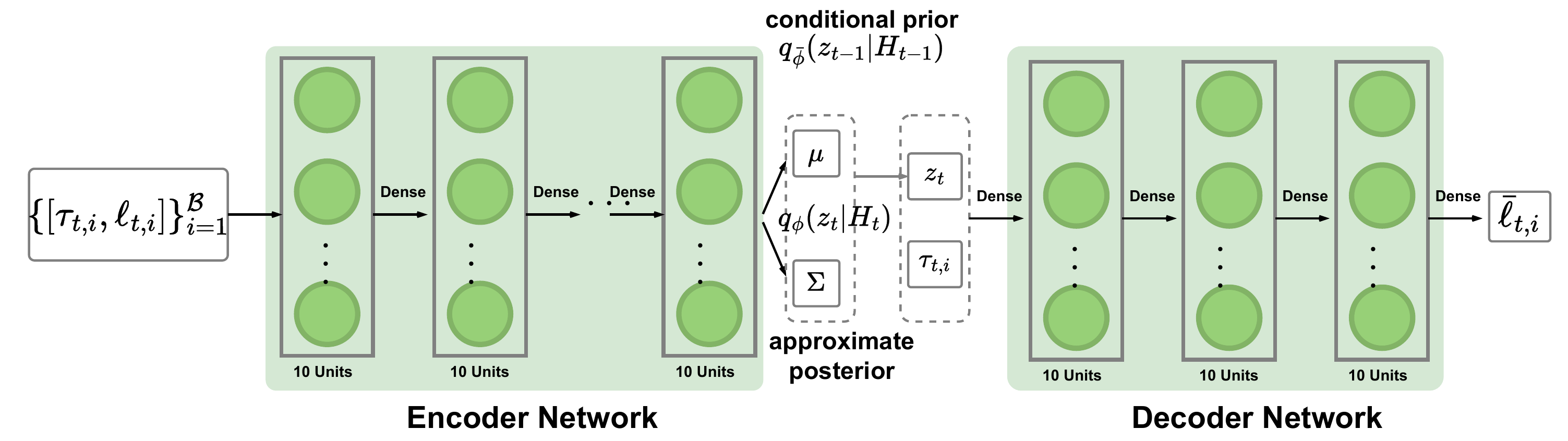}}
\caption{\textbf{The Encoder-Decoder Neural Network to Paramterize the RPM.}
}
\label{fig_risk_nn}
\end{center}
\end{figure}

For example, we denote the neural network parameters by $\bm\phi=\{\bm\phi_1,\bm\phi_{2,1},\bm\phi_{2,2}\}$ together with a mean pooling operator $\oplus$, we can have:
\begin{equation}\label{append_eq_neural_module}
    \begin{split}
        \bm r_{i}=h_{\bm\phi_{1}}(\bm\tau_{k,i},\ell_{k,i})
        \
        \forall i\in\{1,\dots,\mathcal{B}\},
        \quad
        \bar{\bm r}=\oplus_{i=1}^{\mathcal{B}}\bm r_{i},
        \quad
        \bm\mu_{\bm\phi}=h_{\bm\phi_{2,1}}(\bar{\bm r})
        \
        \text{and}
        \
        \bm\Sigma_{\bm\phi}=h_{\bm\phi_{2,2}}(\bar{\bm r}),
    \end{split}
\end{equation}
where the output corresponds to $q_{\bm\phi}(\bm z_{t}\vert H_{t})=\mathcal{N}(\bm\mu_{\bm\phi},\bm\Sigma_{\bm\phi})$ (see Fig.~\ref{fig_risk_nn} for details).

Regarding the task risk functional posterior inference module, this work has a close connection with the NP family \citep{garnelo2018conditional,garnelo2018neural,gordon2019convolutional,kim2019attentive,foong2020meta,wang2022learning,wang2022bridge}.
Both handle the set data points in probabilistic inference.

\subsection{Formulation of ELBO \& Stochastic Gradient Estimates}

Unlike previous risk minimization principles in task episodic learning, ours include an additional risk predictive module, which guides the task batch sampling.
Importantly, we use the latent variable to summarize the historical information information and quantify uncertainty in predicting task-specific adaptation risk.
The following details the steps.
\begin{subequations}
    \begin{align}
        \mathcal{L}_{\text{ML}}(\bm\psi):=\ln p_{\bm\psi}(H_{t}\vert H_{1:t-1})
        =\ln\Big[\int p_{\bm\psi}(H_{t}\vert \bm z_{t})p(\bm z_{t}\vert H_{1:t-1})d\bm z_{t}\Big]
        \\
        =\ln\Big[\int q_{\bm\phi}(\bm z_{t}\vert H_{t})\frac{p(\bm z_{t}\vert H_{1:t-1})}{q_{\bm\phi}(\bm z_{t}\vert H_{t})} p_{\bm\psi}(H_{t}\vert \bm z_{t})d\bm z_{t}\Big]
        \\
        \geq
        \mathbb{E}_{q_{\bm\phi}(\bm z_{t}\vert H_{t})}\Big[\ln p_{\bm\psi}(H_{t}\vert \bm z_{t})\Big]
        -D_{KL}\Big[q_{\bm\phi}(\bm z_{t}\vert H_{t})\parallel p(\bm z_{t}\vert H_{1:t-1})\Big]
        :=\mathcal{G}_{\text{ELBO}}(\bm\psi,\bm\phi)
    \end{align}
\end{subequations}

Then, we can rewrite the ELBO with the help of reparameterization trick \citep{kingma2013auto} in Eq. (\ref{append_eq_rep_elbo}).

\begin{subequations}\label{append_eq_rep_elbo}
    \begin{align}
        \hat{\mathcal{G}}_{\text{ELBO}}(\bm\psi,\bm\phi)
        =\mathbb{E}_{q_{\bm\phi}(\bm z_{t}\vert H_{t})}\Big[\ln p_{\bm\psi}(H_{t}\vert \bm z_{t})\Big]
        -D_{KL}\Big[q_{\bm\phi}(\bm z_{t}\vert H_{t})\parallel p(\bm z_{t}\vert H_{1:t-1})\Big]
        \\
        =\mathbb{E}_{p(\bm\epsilon)}\Big[\ln p_{\bm\psi}(H_{t}\vert g_{\bm\phi}(\bm\epsilon,H_{t}))\Big]
        -D_{KL}\Big[q_{\bm\phi}(\bm z_{t}\vert H_{t})\parallel p(\bm z_{t}\vert H_{1:t-1})\Big]
        \\
        \approx\ln p_{\bm\psi}(H_{t}\vert g_{\bm\phi}(\bm\epsilon,H_{t}))-D_{KL}\Big[q_{\bm\phi}(\bm z_{t}\vert H_{t})\parallel p(\bm z_{t}\vert H_{1:t-1})\Big],
        \quad
        \text{with}
        \quad
        \bm\epsilon\sim\mathcal{N}(0,\bm I_{d})
    \end{align}
\end{subequations}

Moreover, we estimate the stochastic gradients \textit{w.r.t.} all model parameters based on the reparameterized latent variable distribution.

\begin{subequations}\label{append_eq:sg_estimate}
    \begin{align}
        \nabla_{\bm\phi}\mathcal{G}_{\text{ELBO}}(\bm\psi,\bm\phi)
        \approx
        \nabla_{\bm\phi}\ln p_{\bm\psi}(H_{t}\vert g_{\bm\phi}(\bm\epsilon,H_{t}))
        -\frac{1}{2}\nabla_{\bm\phi}\Big(\text{Tr}(\hat{\bm\Sigma}^{-1}\bm\Sigma_{\bm\phi})+(\hat{\bm\mu}-\bm\mu_{\bm\phi})^{T}\hat{\bm\Sigma}(\hat{\bm\mu}-\bm\mu_{\bm\phi})-\ln(\det\bm\Sigma_{\bm\phi})\Big)
        \\
        \text{with}
        \quad
        q_{\bm\phi}(\bm z_{t}\vert H_{t})=\mathcal{N}(\bm\mu_{\bm\phi},\bm\Sigma_{\bm\phi})
        \ 
        \text{and}
        \
        p(\bm z_{t}\vert H_{1:t-1})=\mathcal{N}(\hat{\bm\mu},\hat{\bm\Sigma})
        \\
        \nabla_{\bm\psi}\mathcal{G}_{\text{ELBO}}(\bm\psi,\bm\phi)
        \approx
        \nabla_{\bm\psi}\ln p_{\bm\psi}(H_{t}\vert g_{\bm\phi}(\bm\epsilon,H_{t}))
    \end{align}
\end{subequations}

As illustrated in Eq. (\ref{append_eq:sg_estimate}), one stochastic forward pass is required for gradient estimates in the training process.
For flexible implementation, we adopt a $\beta$-VAE strategy to turn Eq. (\ref{append_eq_rep_elbo}) into 
\begin{equation}
    \begin{split}
        \max_{\bm\psi\in\bm\Psi,\bm\phi\in\bm\Phi}\mathcal{G}_{\text{ELBO}}(\bm\psi,\bm\phi):=\mathbb{E}_{q_{\bm\phi}(\bm z_{t}\vert H_{t})}\left[\sum_{i=1}^{\mathcal{B}}\ln p_{\bm\psi}(\ell_{t,i}\vert\bm\tau_{t,i},\bm z_{t})\right]-\beta D_{KL}\Big[q_{\bm\phi}(\bm z_{t}\vert H_{t})\parallel q_{\bar{\bm\phi}}(\bm z_{t}\vert H_{t-1})\Big]
    \end{split}
\end{equation}

\subsection{Theoretical Guarantee for Task Difficulties' Scoring with Posterior Inference}

\textit{Assumption 1} {(Lipschitz Continuity)}
{We assume the adaptation risk function $\ell(\cdot;\bm\theta)$ reserves the Lipschitz continuity w.r.t. $\bm\theta$ and $\bm\tau$, i.e.,
    \begin{equation}\label{eq_lipschitz_l_supp}
        \begin{split}
            |\ell(\mathcal{D}_{\tau}^{Q},\mathcal{D}_{\tau}^{S};\bm\theta)-\ell(\mathcal{D}_{\tau}^{Q},\mathcal{D}_{\tau}^{S};\bm\theta^{\prime})|\leq\beta_{1}||\bm\theta-\bm\theta^{\prime}||\quad\text{and}\quad
            |\ell(\mathcal{D}_{\tau}^{Q},\mathcal{D}_{\tau}^{S};\bm\theta)-\ell(\mathcal{D}_{\tau^{\prime}}^{Q},\mathcal{D}_{\tau^{\prime}}^{S};\bm\theta)|\leq\beta_{2}||\bm\tau-\bm\tau^{\prime}||,
        \end{split}
    \end{equation}
    where $\forall\{\bm\theta,\bm\theta^{\prime}\}\in\bm\Theta$ and $\forall\{\bm\tau,\bm\tau^{\prime}\}\in\mathcal{T}$ with Lipschitz constants $\beta_1$ and $\beta_2$.}

\textit{Assumption 2} {(Bounded Sample Gradient)}
{We assume the norm of the adaptation risk function's gradient $\nabla\ell(\cdot;\bm\theta_{t})$ is bounded:
    \begin{equation}
        \sup_{\tau\in\mathcal{T}}||\nabla_{\bm\theta}\ell(\mathcal{D}_{\tau}^{Q},\mathcal{D}_{\tau}^{S};\bm\theta_{t})||_{2}<G_{t}\
        \text{and}\
        \sup_{\tau\in\mathcal{T},t\in\mathbb{N}_{+}}||\nabla_{\bm\theta}\ell(\mathcal{D}_{\tau}^{Q},\mathcal{D}_{\tau}^{S};\bm\theta_{t})||_{2}<G,
    \end{equation}
    where $G_t$ is a positive constant and $G$ is a overall bound.}

\textit{Assumption 3} {(Sub-Gaussian Stochastic Gradient)}
{The stochastic gradient $\Tilde{\bm g}:=\bm g+\bm\epsilon$ for the machine learner's adaptation at $t$-th iteration is $\sigma$-sub-Gaussian, which means:
    \begin{equation}
        \begin{split}
            \mathbb{E}\left[\exp{(\eta\bm v^{T}\bm\epsilon)}\right]\leq\exp{\left(\frac{\eta^{2}\sigma^{2}||\bm v||_{2}^{2}}{2}\right)}
            \quad
            \forall
            \eta\in\mathbb{R}\
            \text{and}\
            \bm v\in\mathbb{R}^{d},
        \end{split}
    \end{equation}
where $\mathbb{E}[\Tilde{\bm g}]=\bm g$, $\mathbb{E}[||\Tilde{\bm g}-\bm g||_{2}^{2}]\leq\sigma^2$ and $\sigma\in\mathbb{R}_{+}$.}

Given the Assumption 3 and the Chernoff bound \citep{rigollet2023high}, we can have the concentration inequality as:
\begin{equation}
    \begin{split}
        \mathbb{P}(||\Tilde{\bm g}-\bm g||_2\geq t)\leq 2\exp{(-\frac{t^2}{2\sigma^2})}\quad
        \forall t\in\mathbb{R}.
    \end{split}
\end{equation}

\textit{Theorem~\ref{theorem_prob_score}} {(Provably Approximately Invariant Task Difficulties)}
{Given arbitrary $K$ data points $\{(\bm\tau_i,\ell(\mathcal{D}_{\tau_i}^{Q},\mathcal{D}_{\tau_i}^{S};\bm\theta_t)\}_{i=1}^{K}$, the adaptation gradient $\nabla_{\bm\theta}\mathcal{\bm L}(\bm\theta_t)$ as a $\sigma$-sub-Gaussian random variable and $\bm\theta_{t+1}=\bm\theta_{t}-\eta\nabla_{\bm\theta}\mathcal{\bm L}(\bm\theta_t)$, we denote the relative difficulty via the difference $\Delta_{ij}(\bm\theta_{t+1})=\ell(\mathcal{D}_{\tau_i}^{Q},\mathcal{D}_{\tau_i}^{S};\bm\theta_{t+1})-\ell(\mathcal{D}_{\tau_j}^{Q},\mathcal{D}_{\tau_j}^{S};\bm\theta_{t+1})$ and $\Delta_{ij}(\bm\theta_{t})=\ell(\mathcal{D}_{\tau_i}^{Q},\mathcal{D}_{\tau_i}^{S};\bm\theta_{t})-\ell(\mathcal{D}_{\tau_j}^{Q},\mathcal{D}_{\tau_j}^{S};\bm\theta_{t})$ between $t$-th and $(t+1)$-th iterations, and the gradient difference as $\bm v_{ij}:=\nabla_{\bm\theta}\ell(\mathcal{D}_{\tau_i}^{Q},\mathcal{D}_{\tau_i}^{S};\bm\theta_{t})-\nabla_{\bm\theta}\ell(\mathcal{D}_{\tau_j}^{Q},\mathcal{D}_{\tau_j}^{S};\bm\theta_{t})$.}

{Under Assumption 1/2/3, the set of rank-preserving variable $E_{ij}:=\mathds{1}\left[\text{sign}(\Delta_{ij}(\bm\theta_{t+1}))=\text{sign}(\Delta_{ij}(\bm\theta_{t}))\right]$ satisfies the probability inequality:
$$\mathbb{P}(\bigcap_{i<j}E_{ij})\geq 1-\xi,$$
when $\eta_{t}\leq\frac{\delta_t}{2G_{t}M_{t}+\sqrt{8\sigma^{2}G_{t}^{2}\ln\left(\frac{K(K-1)}{2\xi}\right)}}$ with $G_t$ in Assumption 2, $\delta_{t}:=\min_{i\neq j}|\ell(\mathcal{D}_{\tau_i}^{Q},\mathcal{D}_{\tau_i}^{S};\bm\theta_t)-\ell(\mathcal{D}_{\tau_j}^{Q},\mathcal{D}_{\tau_j}^{S};\bm\theta_t)|\in\mathbb{R}_{+}$, the stochastic gradient norm $M_{t}:=||\nabla_{\bm\theta}\mathcal{\bm L}(\bm\theta_t)||_{2}$.}

The purpose of this part is to uncover the mechanism of the RPM in amortized evaluation of adaptation risk values and scoring the difficulty of tasks.
The function of the RPM relies on Assumptions 1/2/3 and the posterior inference $p(\ell\vert\tau,H_{1:t};\bm\theta_t)$ from the historical risk information $H_{1:t}$.
The foundation of predicting the outcome of optimization in a rough granularity lies in the Theorem~\ref{theorem_prob_score}, and we detail the proof of such a theorem as below.

\textcolor{red}{\textbf{\textcircled{1}. Any-Shot Adaptation After One-step Gradient Descent.}}

Here, we consider a set of data points for the RPM $\{(\bm\tau_i,\ell(\mathcal{D}_{\tau_i}^{Q},\mathcal{D}_{\tau_i}^{S};\bm\theta_t)\}_{i=1}^{K}$ under an arbitrary fixed machine learner $\bm\theta_t$, where tasks in the set $\{\tau_i\}_{i=1}^{K}$ are randomly sampled from $p(\tau)$.
Without loss of generality, we can assume that the adaptation risk values satisfy a rank ordering:
\begin{equation}
    \ell(\mathcal{D}_{\tau_1}^{Q},\mathcal{D}_{\tau_1}^{S};\bm\theta_t)\geq\ell(\mathcal{D}_{\tau_2}^{Q},\mathcal{D}_{\tau_2}^{S};\bm\theta_t)\geq\cdots\geq\ell(\mathcal{D}_{\tau_K}^{Q},\mathcal{D}_{\tau_K}^{S};\bm\theta_t).
\end{equation}

The gradient descent as fast adaptation is denoted by:
\begin{equation}
    \begin{split}
        \bm\theta_{t+1}=\bm\theta_{t}-\eta_{t}\nabla_{\bm\theta}\mathcal{\bm L}(\bm\theta_t).
    \end{split}
\end{equation}
After the above operator, we can obtain another set of data points for the updated RPM $\{(\bm\tau_i,\ell(\mathcal{D}_{\tau_i}^{Q},\mathcal{D}_{\tau_i}^{S};\bm\theta_{t+1})\}_{i=1}^{K}$.

\textcolor{red}{\textbf{\textcircled{2}. Changes of Adaptation Risk Values and Pairwise Ranks.}}

Based on the Assumption 1, we can perform local approximation over $\ell(\mathcal{D}_{\tau_i}^{Q},\mathcal{D}_{\tau_i}^{S};\bm\theta)$ with the help of first-order Talor expansion \textit{w.r.t.} the $\bm\theta_{t}$:
\begin{subequations}
    \begin{align}
        \ell(\mathcal{D}_{\tau_i}^{Q},\mathcal{D}_{\tau_i}^{S};\bm\theta_{t+1})
    =\ell(\mathcal{D}_{\tau_i}^{Q},\mathcal{D}_{\tau_i}^{S};\bm\theta_{t})-\eta_{t}\nabla_{\bm\theta}\ell(\mathcal{D}_{\tau_i}^{Q},\mathcal{D}_{\tau_i}^{S};\bm\theta_{t})^{T}\mathcal{\bm L}(\bm\theta_t)+\mathcal{O}(\eta_{t}^{2}||\nabla_{\bm\theta}\mathcal{\bm L}(\bm\theta_t)||_{2}^2)\\
     \ell(\mathcal{D}_{\tau_i}^{Q},\mathcal{D}_{\tau_i}^{S};\bm\theta_{t+1})
    \approx\ell(\mathcal{D}_{\tau_i}^{Q},\mathcal{D}_{\tau_i}^{S};\bm\theta_{t})-\nabla_{\bm\theta}\ell(\mathcal{D}_{\tau_i}^{Q},\mathcal{D}_{\tau_i}^{S};\bm\theta_{t})^{T}\mathcal{\bm L}(\bm\theta_t)
    \quad\forall i\in \{1,2,\dots,K\}.
    \end{align}
\end{subequations}

One straightforward way to assess the task difficulty is to compare arbitrary paired tasks $\{\bm\tau_i,\bm\tau_j\}$'s adaptation risk values $\{\ell(\mathcal{D}_{\tau_i}^{Q},\mathcal{D}_{\tau_i}^{S};\bm\theta_t),\ell(\mathcal{D}_{\tau_j}^{Q},\mathcal{D}_{\tau_j}^{S};\bm\theta_t)\}$ with $i<j$.
Then, we can estimate the relative difficulty via the difference as:
\begin{equation}
    \begin{split}
        \Delta_{ij}(\bm\theta_{t+1})
        \approx\Delta_{ij}(\bm\theta_{t})-\eta_{t}\left(\nabla_{\bm\theta}\ell(\mathcal{D}_{\tau_i}^{Q},\mathcal{D}_{\tau_i}^{S};\bm\theta_{t})-\nabla_{\bm\theta}\ell(\mathcal{D}_{\tau_j}^{Q},\mathcal{D}_{\tau_j}^{S};\bm\theta_{t})\right)^{T}\nabla_{\bm\theta}\mathcal{\bm L}(\bm\theta_t),
    \end{split}
\end{equation}
where we denote the relative difficulty via the difference as $\Delta_{ij}(\bm\theta_{t+1})=\ell(\mathcal{D}_{\tau_i}^{Q},\mathcal{D}_{\tau_i}^{S};\bm\theta_{t+1})-\ell(\mathcal{D}_{\tau_j}^{Q},\mathcal{D}_{\tau_j}^{S};\bm\theta_{t+1})$ and $\Delta_{ij}(\bm\theta_{t})=\ell(\mathcal{D}_{\tau_i}^{Q},\mathcal{D}_{\tau_i}^{S};\bm\theta_{t})-\ell(\mathcal{D}_{\tau_j}^{Q},\mathcal{D}_{\tau_j}^{S};\bm\theta_{t})$ between $t$-th and $(t+1)$-th iterations.
As $\Delta_{ij}(\bm\theta_{t})$ is positive, one feasible condition for $\Delta_{ij}(\bm\theta_{t+1})\in\mathbb{R}_{+}$ is:
\begin{subequations}\label{eq_diff_difer}
    \begin{align}
         \Delta_{ij}(\bm\theta_{t+1})
        \approx\Delta_{ij}(\bm\theta_{t})-\eta_{t}\left(\nabla_{\bm\theta}\ell(\mathcal{D}_{\tau_i}^{Q},\mathcal{D}_{\tau_i}^{S};\bm\theta_{t})-\nabla_{\bm\theta}\ell(\mathcal{D}_{\tau_j}^{Q},\mathcal{D}_{\tau_j}^{S};\bm\theta_{t})\right)^{T}\nabla_{\bm\theta}\mathcal{\bm L}(\bm\theta_t)\\
        \geq\Delta_{ij}(\bm\theta_{t})-2\eta_{t} G_{t}M>0,
        \quad\Rightarrow
        \eta_{t}<\frac{\Delta_{ij}(\bm\theta_{t})}{2G_{t}M_{t}}
        \quad
        \text{with}\quad
        M_{t}:=||\nabla_{\bm\theta}\mathcal{\bm L}(\bm\theta_t)||_{2}.
    \end{align}
\end{subequations}
The above implies that when the learning rate $\eta_{t}$ in gradient step is smaller enough, the relative difficulty between the task $i$ and $j$ can be preserved after the machine learner's update with the Assumption 2.

\textcolor{red}{\textbf{\textcircled{3}. Probabilistic Inequality with a Nearly Invariant Ranking Guarantee.}}

In practice, the stochastic gradient descent is performed, which means the gradient is a random variable $\nabla_{\bm\theta}\mathcal{\bm L}(\bm\theta_t)=\bm g_{t}+\bm\epsilon$ with $\mathbb{E}[\bm\epsilon]=0,\mathbb{E}[||\bm\epsilon||_2^2]<\sigma^2$ and $\bm g_{t}=\mathbb{E}\left[\nabla_{\bm\theta}\mathcal{\bm L}(\bm\theta_t)\right]$.
Meanwhile, we denote the gradient difference by $\bm v_{ij}:=\nabla_{\bm\theta}\ell(\mathcal{D}_{\tau_i}^{Q},\mathcal{D}_{\tau_i}^{S};\bm\theta_{t})-\nabla_{\bm\theta}\ell(\mathcal{D}_{\tau_j}^{Q},\mathcal{D}_{\tau_j}^{S};\bm\theta_{t})$, which leads to:
\begin{equation}
    \begin{split}
        ||\bm v_{ij}||_2\leq 2\sup_{\tau\in\mathcal{T}}||\nabla_{\bm\theta}\ell(\mathcal{D}_{\tau}^{Q},\mathcal{D}_{\tau}^{S};\bm\theta)||_{2}<2G_{t},
    \end{split}
\end{equation}
according to the Assumption 2.
Another variable is introduced as the minimum separation between arbitrary paired adaptation risk values:
\begin{equation}
    \begin{split}
        \delta_{t}:=\min_{i\neq j}|\ell(\mathcal{D}_{\tau_i}^{Q},\mathcal{D}_{\tau_i}^{S};\bm\theta_t)-\ell(\mathcal{D}_{\tau_j}^{Q},\mathcal{D}_{\tau_j}^{S};\bm\theta_t)|\in\mathbb{R}_{+}.
    \end{split}
\end{equation}
Still, to make sure the invariant rank, one necessary condition can be:
\begin{equation}
    \begin{split}
        \Delta_{ij}(\bm\theta_{t})-\eta_{t}\bm v_{ij}^{T}\bm g_{t}\geq 0.
    \end{split}
\end{equation}
And the above inequality reasonably holds when $\eta_{t}\bm v_{ij}^{T}\bm g_{t}<\delta_t$.
Here, we define the random event $E_{ij}:=\mathds{1}\left[\text{sign}(\Delta_{ij}(\bm\theta_{t+1}))=\text{sign}(\Delta_{ij}(\bm\theta_{t}))\right]$ from the task pair together with $E_{ij}^{c}:=\mathds{1}\left[\text{sign}(\Delta_{ij}(\bm\theta_{t+1}))\neq\text{sign}(\Delta_{ij}(\bm\theta_{t}))\right]$.
With the help of $\sigma$-sub-Gaussain property in Assumption 3, we can bound the case of the rank flipping as (note some critical conditions that $\bm v_{ij}^{T}\bm g_{t}\in\mathbb{R}_{+}$ and $\eta_{t}\bm v_{ij}^{T}\bm g_{t}<\delta_{t}$ as the learning rate $\eta_{t}$ can be typically smaller enough):
\begin{subequations}
    \begin{align}
    \mathbb{P}(E_{ij}^{c})=\mathbb{P}(\eta_{t}\bm v_{ij}^{T}\nabla_{\bm\theta}\mathcal{\bm L}(\bm\theta_t)\geq\delta_t)\leq \exp{\left(-\frac{(\delta_{t}-\eta_{t}\bm v_{ij}^{T}\bm g_{t})^{2}}{2\eta_{t}^{2}\sigma^{2}||\bm v_{ij}||_{2}^{2}}\right)}
    < \exp{\left(-\frac{(\delta_{t}-2\eta_{t} G_{t}M_{t})^{2}}{8\eta_{t}^{2}\sigma^{2}G_{t}^{2}}\right)}\\
    \mathbb{P}(\bigcup_{i<j}E_{ij}^{c})\leq\sum_{i<j}\mathbb{P}(E_{ij}^{c})
    \leq\frac{K(K-1)}{2}\exp{\left(-\frac{(\delta_{t}-2\eta_{t} G_{t}M_{t})^{2}}{8\eta_{t}^{2}\sigma^{2}G_{t}^{2}}\right)}\\
    \mathbb{P}(\bigcap_{i<j}E_{ij})=1-\mathbb{P}(\bigcup_{i<j}E_{ij}^{c})\geq 1-\frac{K(K-1)}{2}\exp{\left(-\frac{(\delta_{t}-2\eta_{t} G_{t}M_{t})^{2}}{8\eta_{t}^{2}\sigma^{2}G_{t}^{2}}\right)}\geq 1-\xi.
    \end{align}
\end{subequations}

The condition for the above inequality holds is $\eta_{t}\leq\frac{\delta_t}{2G_{t}M_{t}+\sqrt{8\sigma^{2}G_{t}^{2}\ln\left(\frac{K(K-1)}{2\xi}\right)}}$.
With the above \textcolor{red}{\textbf{steps \textcircled{1}-\textcircled{3}}} and corresponding conditions, we complete the proof.

\section{Convergence Analysis}\label{sec_supp_convergence}

\paragraph{Predictive Foundation of MPTS.}
In brief, the foundation of MPTS is to use the risk history and generalization results under $\bm\theta_t$ to approximately rank the difficulty of task samples in $(t+1)$-th iteration.
Note that precise sample average Monte Carlo for CVaR relies on the exact evaluation and access to the $(t+1)$-th adaptation risk values in $\hat{\mathcal{B}}$ tasks to pick up the Top-$\mathcal{B}$ ones.
MPTS tries to circumvent the complete evaluation and uses the $t$-th adaptation risk batch $\{\ell_{t,i}\}_{i=1}^{\mathcal{B}}$ to train a generative model with $H_{1:t-1}$ to extract the prior to forecast the $\hat{\mathcal{B}}$ tasks' difficulty order for $(t+1)$-th subset selection. (Importantly, such a setup necessitates MPTS when accessing expensive evaluation of $\{\ell_{t+1,i}\}_{i=1}^{\hat{\mathcal{B}}}$ is prohibited.)

The theoretical analysis focuses on the rank-flipping that occurs during the model's update $\bm\theta_{t}\to\bm\theta_{t+1}$, and we will demonstrate that, under certain conditions, this does not affect the convergence of the adopted strategy to surrogate MC-CVaR in a biased manner.
The candidate task pool is $\mathcal{T}_{t+1}^{\hat{\mathcal{B}}}$ used for the subset selection and robust adaptation at the $(t+1)$-th iteration.
The Top-$\mathcal{B}$ task batch under $\bm\theta_t$'s risk evaluation is denoted by $\bar{\mathcal{T}}_{t}^{\mathcal{B}}\subset\mathcal{T}_{t+1}^{\hat{\mathcal{B}}}$, and its complementary set is $\mathcal{T}_{t}^{C}:=\mathcal{T}_{t+1}^{\hat{\mathcal{B}}}\setminus\bar{\mathcal{T}}_{t}^{\mathcal{B}}$.
The ground truth Top-$\mathcal{B}$ task batch under $\bm\theta_{t+1}$'s risk evaluation is $\mathcal{T}_{t+1}^{\mathcal{B}}$.

\textit{Assumption 4}
{(Margin Anti-Concentration)} 
{We assume $\Delta_{ij}(\bm\theta_{t})=\ell(\mathcal{D}_{\tau_i}^{Q},\mathcal{D}_{\tau_i}^{S};\bm\theta_{t})-\ell(\mathcal{D}_{\tau_j}^{Q},\mathcal{D}_{\tau_j}^{S};\bm\theta_{t})$ has a density uniformly bounded by $\rho$ in a neighborhood of zero.
That is, $\forall \epsilon>0$, the probability inequality holds:
\begin{equation}
        \mathbb{P}(|\Delta_{ij}(\bm\theta_{t})|\leq\epsilon)\leq\rho\epsilon.
\end{equation}}

The above assumption is reasonable in online robust optimization scenarios.
If a large mass of task pairs has nearly identical loss, this suggests the Top-$\mathcal{B}$ set is unstable, making it difficult for any difficulty prioritized sampling algorithm to approximate the CVaR subset stably. Anti-concentration ensures that near-ties are probabilistically rare, hence rank changes are infrequent over iterations.

\textit{Lemma~\ref{lemma_mis_rank_quant}} (Misranked Subset Quantity)
{In the presence of adaptation risk value, let $N_{f}$ be the number of order flipped cross-pairs between the ground-truth Top-$\mathcal{B}$ subset $\bar{\mathcal{T}}_{t}^{\mathcal{B}}$ under $\bm\theta_{t}$ and the remainder of the candidate tasks $\mathcal{T}_{t}^{C}:=\mathcal{T}^{\hat{\mathcal{B}}}_{t+1}\setminus\bar{\mathcal{T}}_{t}^{\mathcal{B}}$ when the machine learner's parameter changes from $\bm\theta_t$ to $\bm\theta_{t+1}$.
We denote the number of tasks that change Top-$\mathcal{B}$ membership by $m_{t+1}=|\bar{\mathcal{T}}_{t}^{\mathcal{B}}\bigtriangleup\mathcal{T}_{t+1}^{\mathcal{B}}|$, where $\mathcal{T}_{t+1}^{\mathcal{B}}$ is the ground-truth Top-$\mathcal{B}$ subset under $\bm\theta_{t+1}$.
Then the inequality holds: $m_{t+1}\leq 2N_{f}$.}

\begin{figure}[htbp]
\begin{center}
\centerline{\includegraphics[width=0.7\textwidth]{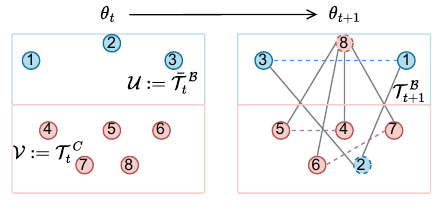}}
\caption{\textbf{Rank Flipping after One-Step Update.}
We number tasks from the sampled pseudo task batch $\mathcal{T}_{t+1}^{\hat{\mathcal{B}}}$ for the $(t+1)$-th subset selection.
The left figure partitions the set into $\mathcal{U}\cup\mathcal{V}$ as illustrated in the Proof of Lemma~\ref{lemma_mis_rank_quant}, with the blue ones, i.e., 1, 2, 3 as Top-$\mathcal{B}$ tasks evaluated under $\bm\theta_t$ and the red ones as the complementary set.
Note that tasks 2 and 8 are swapped in relative Top-$\mathcal{B}$ ranking after one-step model update; hence, the right figure depicts the Top-$\mathcal{B}$ tasks as 1,3, and 8.
The gray lines indicate possible order-flipped cross-pairs between $\mathcal{U}$ and $\mathcal{V}$.
The blue and the red lines are probably flipped cross-pairs in separate partitioned sets' induced graphs.
}
\label{fig_rank_flip}
\end{center}
\end{figure}

\textit{Proof.}
{Now we can construct a bipartite graph $\mathcal{G}=(\mathcal{U},\mathcal{V},\mathcal{E})$, where $\mathcal{U}$ denotes the ground-truth Top-$\mathcal{B}$ subset $\bar{\mathcal{T}}_{t}^{\mathcal{B}}$ while $\mathcal{V}$ is the remainder of the candidate tasks $\mathcal{T}_{t}^{C}:=\mathcal{T}^{\hat{\mathcal{B}}}_{t+1}\setminus\bar{\mathcal{T}}_{t}^{\mathcal{B}}$.
Each task in the set corresponds to a vertex in the graph.
Also note that there are exactly $\mathcal{B}(\hat{\mathcal{B}}-\mathcal{B})$ ordered task cross-pairs.}

{For each ordered cross pair $(i\in \mathcal{U}; j\in \mathcal{V})$, which flips, i.e. the sign of $\Delta_{ij}(\bm\theta_t)$ changes between $\bm\theta_{t}$ and $\bm\theta_{t+1}$, we place an undirected edge $e_{ij}$ between vertices $i$ and $j$, and this constitutes the edge set $\mathcal{E}$.
Further, the subset of vertices, where the order flipping occurs, can be defined as $\mathcal{V}_{E}=\{\tau\in\mathcal{T}^{\hat{\mathcal{B}}}_{t+1}\vert\exists e\in\mathcal{E}\ \text{with}\ \tau\in e\}$, which describes the set of tasks that participate in at least one flipped cross-pair.
As a result, we can have $\bar{\mathcal{T}}_{t}^{\mathcal{B}}\bigtriangleup\mathcal{T}_{t+1}^{\mathcal{B}}\subseteq\mathcal{V}_{E}$.
Obviously, $m_{t+1}\leq|\mathcal{V}_{E}|$.}

{The total number of flipping pairs is $N_f$.
As illustrated in Fig.~\ref{fig_rank_flip}, the flipped ordered cross-pairs comprises three disjoint cases or edge partition in $\mathcal{G}$: (i) order flipping between $\bar{\mathcal{T}}_{t}^{\mathcal{B}}$, (ii) order flipping between $\mathcal{T}_{t}^{C}$, and (iii) order flipping between $\bar{\mathcal{T}}_{t}^{\mathcal{B}}$ and $\mathcal{T}_{t}^{C}$.
This indicates that $|\mathcal{E}|\leq N_{f}$.}

{Putting the above together, we can have $m_{t+1}\leq|\mathcal{V}_{E}|\leq 2|\mathcal{E}|\leq 2N_{f}$.
And this establishes the connection between the relative task difficulty ranking and the direct task difficulty ranking.
And this completes the proof of \textbf{Lemma} 1.
}

\textit{Lemma~\ref{lemma_rank_preserve_bound}} (Rank-Preserving Bound in Expectation)
{With the risk difference notation $\Delta_{ij}(\bm\theta_{t})=\ell(\mathcal{D}_{\tau_i}^{Q},\mathcal{D}_{\tau_i}^{S};\bm\theta_{t})-\ell(\mathcal{D}_{\tau_j}^{Q},\mathcal{D}_{\tau_j}^{S};\bm\theta_{t})$ and $\Delta_{ij}(\bm\theta_{t+1})=\ell(\mathcal{D}_{\tau_i}^{Q},\mathcal{D}_{\tau_i}^{S};\bm\theta_{t+1})-\ell(\mathcal{D}_{\tau_j}^{Q},\mathcal{D}_{\tau_j}^{S};\bm\theta_{t+1})$, when the $(i,j)$ cross-pair flips its order from $\bm\theta_t$ to $\bm\theta_{t+1}$, we conclude that $|\Delta_{ij}(\bm\theta_t)|\leq|\Delta_{ij}(\bm\theta_{t+1})-\Delta_{ij}(\bm\theta_{t})|$.}

\textit{Proof.}
{Now let us consider limited scenarios for the occurrence of the order flipping event $\bar{E}_{ij}:=\mathds{1}\left[\text{sign}(\Delta_{ij}(\bm\theta_{t+1}))\neq\text{sign}(\Delta_{ij}(\bm\theta_{t}))\right]$:}

{In scenario (1), where $\Delta_{ij}(\bm\theta_t)<0$ and $\Delta_{ij}(\bm\theta_{t+1})\geq 0$ hold, this implies that:}
\begin{equation}
    \underbrace{\Delta_{ij}(\bm\theta_{t+1})-\Delta_{ij}(\bm\theta_t)}_{\text{Positive Value}}\geq \underbrace{-\Delta_{ij}(\bm\theta_t)}_{\text{Positive Value}}.
\end{equation}

\textit{In scenario (2), where $\Delta_{ij}(\bm\theta_t)\geq 0$ and $\Delta_{ij}(\bm\theta_{t+1})< 0$ hold, this implies that:}
\begin{equation}
    \underbrace{\Delta_{ij}(\bm\theta_{t+1})-\Delta_{ij}(\bm\theta_t)}_{\text{Non-Positive Value}}\leq \underbrace{0-\Delta_{ij}(\bm\theta_t)}_{\text{Non-Positive Value}}.
\end{equation}

{Putting the above together, we can derive that $|\Delta_{ij}(\bm\theta_t)|\leq|\Delta_{ij}(\bm\theta_{t+1})-\Delta_{ij}(\bm\theta_{t})|$.
And this completes the proof of \textbf{Lemma} 2.}

Next, we analyze the term $|\Delta_{ij}(\bm\theta_{t+1})-\Delta_{ij}(\bm\theta_{t})|$ and associate it with $|\Delta_{ij}(\bm\theta_t)|$:
\begin{subequations}\label{eq_delta_bound}
    \begin{align}
        |\Delta_{ij}(\bm\theta_{t+1})-\Delta_{ij}(\bm\theta_{t})|
        =|\ell(\mathcal{D}_{\tau_i}^{Q},\mathcal{D}_{\tau_i}^{S};\bm\theta_{t+1})-\ell(\mathcal{D}_{\tau_i}^{Q},\mathcal{D}_{\tau_i}^{S};\bm\theta_{t})-\ell(\mathcal{D}_{\tau_j}^{Q},\mathcal{D}_{\tau_j}^{S};\bm\theta_{t+1})+\ell(\mathcal{D}_{\tau_j}^{Q},\mathcal{D}_{\tau_j}^{S};\bm\theta_{t})|\\
        \leq|\ell(\mathcal{D}_{\tau_i}^{Q},\mathcal{D}_{\tau_i}^{S};\bm\theta_{t+1})-\ell(\mathcal{D}_{\tau_i}^{Q},\mathcal{D}_{\tau_i}^{S};\bm\theta_{t})|+|\ell(\mathcal{D}_{\tau_j}^{Q},\mathcal{D}_{\tau_j}^{S};\bm\theta_{t+1})-\ell(\mathcal{D}_{\tau_j}^{Q},\mathcal{D}_{\tau_j}^{S};\bm\theta_{t})|\\
        \leq 2G_{t}||\bm\theta_{t+1}-\bm\theta_{t}||_{2}\\
        \Rightarrow
        |\Delta_{ij}(\bm\theta_t)|\leq 2G_{t}||\bm\theta_{t+1}-\bm\theta_{t}||_{2}.
    \end{align}
\end{subequations}

\paragraph{Connection between Rank Flipping and Gradient Difference.}
From \textbf{Lemma} 2 together with the anti-concentration Assumption 4, we can bound the probability of the order flipping event $\bar{E}_{ij}$ with the one-step iteration:
\begin{equation}\label{eq_bounded_eij}
    \mathbb{P}(\bar{E}_{ij}\vert\bm\theta_{t},\bm\theta_{t+1})\leq
    \mathbb{P}\left(|\Delta_{ij}(\bm\theta_t)|\leq 2G_{t}||\bm\theta_{t+1}-\bm\theta_{t}||_{2}\right)
    \leq 2\rho G_{t}||\bm\theta_{t+1}-\bm\theta_{t}||_{2}.
\end{equation}

Note that the graph only cares about pairwise relation between the Top-$\mathcal{B}$ tasks and other $\hat{\mathcal{B}}-\mathcal{B}$ tasks.
Now in a population level, the scale of plausible pairs in the task set $\mathcal{T}^{\hat{\mathcal{B}}}_{t+1}$ is $\mathcal{B}(\hat{\mathcal{B}}-\mathcal{B})$ and $|\mathcal{V}_{E}|=\sum_{i,j}\bar{E}_{ij}$.
Also, with the linearity of expectation over $\bar{E}_{ij}$, \textbf{Lemma} 1 and Eq. (\ref{eq_delta_bound})/(\ref{eq_bounded_eij}), we can bound the flipping count expectation:
\begin{subequations}\label{eq_bound_misrank_ratio}
    \begin{align}
        \mathbb{E}[m_{t+1}]\leq\mathbb{E}[|\mathcal{V}_{E}|]\leq 2\rho\mathcal{B}(\hat{\mathcal{B}}-\mathcal{B})G_{t}\mathbb{E}[||\bm\theta_{t+1}-\bm\theta_{t}||_{2}]\\
        r_{t+1}:=\frac{\mathbb{E}[m_{t+1}]}{\mathcal{B}}\leq 2\rho(\hat{\mathcal{B}}-\mathcal{B})G_{t}\mathbb{E}[||\bm\theta_{t+1}-\bm\theta_{t}||_{2}]\leq 2\rho(\hat{\mathcal{B}}-\mathcal{B})G_{t}^{2}\mathcal{O}(\eta_t).
    \end{align}
\end{subequations}

Next, we need to show that the selection bias from rank flipping vanishes quickly and the gradient norm declines to ensure the convergence.
Recall that the expectation form of CVaR can be written as $\mathbb{E}_{p_{\alpha}(\tau;\bm\theta_{t+1})}\left[\ell(\mathcal{D}_{\tau}^{Q},\mathcal{D}_{\tau}^{S};\bm\theta)\right]$, and MC-CVaR is its unbiased estimate from the subset $\mathcal{T}_{t+1}^{\mathcal{B}}\subset\mathcal{T}_{t+1}^{\hat{\mathcal{B}}}$.
Accordingly, the expectation form of perturbed CVaR can be written as $\mathcal{L}(\bm\theta)=\mathbb{E}_{q(\tau;\bm\theta_{t+1})}\left[\ell(\mathcal{D}_{\tau}^{Q},\mathcal{D}_{\tau}^{S};\bm\theta)\right]$ and $q({\tau};\bm\theta_{t+1})$ denotes the distribution of biased subset due to miskranked cases, which produced the biased estimate of expected CVaR from $\bar{\mathcal{T}}_{t}^{\mathcal{B}}\subset\mathcal{T}_{t+1}^{\hat{\mathcal{B}}}$.

Given the latest machine learner parameter $\bm\theta_{t+1}$, we can derive the equation after one-step gradient update as:
\begin{subequations}\label{eq_grad_decomp}
    \begin{align}
        \text{One-Step Gradient Update:}\quad\bm\theta_{t+2}=\bm\theta_{t+1}-\eta_{t+1}\underbrace{\nabla_{\bm\theta}\mathcal{L}(\bm\theta_{t+1})}_{\text{Perturbed Gradient After Rank-Flipping}}\\
        \text{with Gradient Decomposition}\quad
        \nabla_{\bm\theta}\mathcal{L}(\bm\theta_{t+1})=\underbrace{\bar{\bm g}_{t+1}}_{\text{Unbiased Average Gradient}}+\underbrace{\Delta\bm g_{t+1}}_{\text{Gradient Difference}},
    \end{align}
\end{subequations}
where we refer to the average task gradient after rank-flipping as the \textit{perturbed gradient}.
The unbiased average gradient are estimated from $\mathcal{T}_{t+1}^{\mathcal{B}}$, the ground-truth Top-$\mathcal{B}$ subset under $\bm\theta_{t+1}$.
And term 
\[\Delta\bm g_{t+1}=\frac{2}{m_{t+1}}\left(\sum_{\tau\in\bar{\mathcal{T}}_{t}^{\mathcal{B}}\setminus\mathcal{T}_{t+1}^{\mathcal{B}}}\ell(\mathcal{D}_{\tau}^{Q},\mathcal{D}_{\tau}^{S};\bm\theta_{t+1})-\sum_{\tau\in\mathcal{T}_{t+1}^{\mathcal{B}}\setminus\bar{\mathcal{T}}_{t}^{\mathcal{B}}}\ell(\mathcal{D}_{\tau}^{Q},\mathcal{D}_{\tau}^{S};\bm\theta_{t+1})\right)\]
 denotes the gradient difference between rank-flipping tasks after model update.
Eq. (\ref{eq_grad_decomp}) will work for the convergence analysis $\forall t\in\mathbb{N}_{+}$ in the following contents.

\textit{Lemma~\ref{lemma_misrank_acute_angle}} (Misranking Acute-Angle)
{Let $\bar{\bm g}_{t}$ and $\Delta\bm g_{t}$ respectively denote the unbiased Monte Carlo estimate of CVaR and the gradient difference between it and the biased gradient $\nabla_{\bm\theta}\mathcal{L}(\bm\theta_t)$.
For a given $c\in(0,1)$ and arbitrary small $\epsilon_*>0$, there exists $T$ such that $\forall t\geq T$, the following dichotomy holds: either (i) $||\bar{\bm g}_{t}||_{2}\leq\epsilon_*$, or (ii) $||\Delta\bm g_{t}||_{2}\leq c||\bar{\bm g}_{t}||_{2}$.}

\textit{Proof.}
{In case (i) if $||\bar{\bm g}_{t}||_{2}\leq\epsilon_*$, with arbitrary small $\epsilon_*$, we can see that $\lim_{t\to\infty}||\bar{\bm g}_{t}||_{2}=0$.
Also, recall the truth in Eq. (\ref{eq_bound_misrank_ratio}), we can find that $\lim_{t\to\infty}r_{t+1}=0$ when $\lim_{t\to\infty}\eta_t=0$.
Then we can have $\lim_{t\to\infty}||\nabla_{\bm\theta}\mathcal{L}(\bm\theta_t)||_{2}\leq\lim_{t\to\infty}||\bar{\bm g}_{t}||_{2}+||\Delta\bm g_{t}||_{2}\leq\lim_{t\to\infty}||\bar{\bm g}_{t}||_{2}+r_{t+1}G$, which implies the stationarity of $||\nabla_{\bm\theta}\mathcal{L}(\bm\theta_t)||_{2}$ and already guarantees the convergence of algorithms with diminishing rank flipping.}

{In case (ii) if $||\bar{\bm g}_{t}||_{2}>\epsilon_*$, we perform analysis as:
recall the previous finding about the misranking rate $r_{t+1}:=\frac{\mathbb{E}[m_{t+1}]}{\mathcal{B}}\leq 2\rho(\hat{\mathcal{B}}-\mathcal{B})G_{t}^{2}\mathcal{O}(\eta_t)$.
When $\lim_{t\to\infty}\eta_{t}=0$, it is also trivial to see that $\lim_{t\to\infty}r_{t+1}=0$.
Hence, choosing a large enough $T$ such that for all $t>T$, we can have $r_{t+1}\leq\frac{c}{G}\epsilon_*$.
Then the inequality holds:
\begin{equation}\label{eq_lemma_ineq}
    ||\Delta\bm g_{t}||_{2}\leq r_{t+1}G_{t}\leq r_{t+1}G\leq c\epsilon_*.
\end{equation}
Then, together with the case condition $||\bar{\bm g}_{t}||_{2}>\epsilon_*$, we can have:
\begin{equation}
    ||\Delta\bm g_{t}||_{2}\leq c\epsilon_*\leq c||\bar{\bm g}_{t}||_{2}\Rightarrow
    ||\Delta\bm g_{t}||_{2}\leq c||\bar{\bm g}_{t}||_{2}.
\end{equation}}

{Finally, the dichotomy holds for the studied two cases.
And this completes the proof of \textbf{Lemma} 3.}

\textit{Theorem~\ref{theorem_converge}} (Convergence with Diminishing Rank Flipping)
    {Suppose there exists $c\in(0,1)$ and $T\geq 0$ such that $\forall t\geq T$, the inequality holds: $||\Delta\bm g_{t}||_{2}\leq c||\bar{\bm g}_{t}||_{2}$.
    Given the Assumptions 1/2/3/4, and the appropriate construction of the step sizes $\{\eta_t\}$ from Theorem~\ref{theorem_prob_score}, we conclude that:
    $\lim_{t\to\infty}\mathbb{E}\left[||\nabla_{\bm\theta}\mathcal{L}(\bm\theta_t)||_2\right]=0$.
    Hence, the iteration converges to first-order stationary points in expectation.}

\textit{Proof.}
{The \textbf{Theorem} 1 indicates that the probability inequality holds when the learning rate follows that:
$\xi_t\geq\frac{K(K-1)}{2}\exp{\left(-\frac{(\frac{\delta_t}{\eta_t}-2G_{t}M_{t})^2}{8\sigma^{2}G_{t}^{2}}\right)}$.
Then for a given $\eta_t$, we can guarantee the probability of rank-preserving for all pairs as $1-\xi_{t}^{*}:=1-\frac{K(K-1)}{2}\exp{\left(-\frac{(\frac{\delta_t}{\eta_t}-2G_{t}M_{t})^2}{8\sigma^{2}G_{t}^{2}}\right)}$.}

{Since $G_t$ and $M_t$ are bounded, the behavior of the exponent is determined by the coefficient $\frac{\delta_t}{\eta_t}$.
According to the Assumption 4, for a sufficiently small constant $\delta_{\min}>0$, there holds the probability inequality $\mathbb{P}(\delta_t>\delta_{\min})>1-\rho\delta_{\min}$, which 
means $\delta_t>0$ stands with a sufficiently large probability with the sampled tasks well scattered in random sampling.
Hence, the limit of $\frac{\delta_t}{\eta_t}>\frac{\delta_{\min}}{\eta_t}$ goes to the infinity when $\lim_{t\to\infty}\eta_t=0$ in a probabilistic sense.
This suggests the tightening bound of rank preserving with the diminishing $\xi_{t}^{*}$.}

{In practice, we can construct the example of the plausible learning rate scheduling strategy by setting $\eta_t=t^{-b}$ with $b\in(\frac{1}{2},1]$, which naturally makes Robbins-Monro conditions hold:
$\sum_{t}\eta_{t}=\infty\quad\text{and}\quad\sum_{t}\eta_{t}^{2}<\infty$.
This decays the learning rate $\lim_{t\to\infty}\eta_t=0$ over iteration, which probabilistically reduces the chance of rank flipping cases and tightens the inequality bound in \textbf{Theorem} 1.
Note that in this theorem, we leave out the case of approaching stationarity, i.e., case (i) when the gradient norm limit approaches zero from \textbf{Lemma} 3.}

{The next step is to perform the Taylor expansion of the MC-CVaR risk function $\mathcal{L}(\bm\theta)$ at the parameter $\bm\theta_t$, which obtains $\mathcal{L}(\bm\theta)=\mathcal{L}(\bm\theta_t)+\bar{\bm g}_{t}^{T}(\bm\theta-\bm\theta_t)+\mathcal{O}(||\bm\theta-\bm\theta_t||)$.
Then, we input the biased gradient $\nabla_{\bm\theta}\mathcal{L}(\bm\theta_t)$ and the updated parameter to derive the equation as:
\begin{subequations}
    \begin{align}
        \mathcal{L}(\bm\theta_{t+1})
        =\mathcal{L}(\bm\theta_t)-\eta_{t}\bar{\bm g}_{t}^{T}\nabla_{\bm\theta}\mathcal{L}(\bm\theta_t)+\mathcal{O}(||\bm\theta_{t+1}-\bm\theta_{t}||)\\
        \Rightarrow
        \mathcal{L}(\bm\theta_{t+1})
        \approx\mathcal{L}(\bm\theta_t)-\eta_{t}||\bar{\bm g}_{t}||_2^2-\eta_{t}\bar{\bm g}_{t}^{T}\Delta\bm g_{t}\\
        \leq\mathcal{L}(\bm\theta_t)-\eta_{t}||\bar{\bm g}_{t}||_2^2+\eta_{t}||\bar{\bm g}_{t}||_{2}||\Delta\bm g_{t}||_{2}\\
        \leq\mathcal{L}(\bm\theta_t)-\eta_{t}||\bar{\bm g}_{t}||_2^2+c\eta_{t}||\bar{\bm g}_{t}||_{2}^{2}
        =\mathcal{L}(\bm\theta_t)-(1-c)\eta_{t}||\bar{\bm g}_{t}||_2^2\\
        \Rightarrow
        (1-c)\eta_{t}||\bar{\bm g}_{t}||_2^2\leq\mathcal{L}(\bm\theta_{t})-\mathcal{L}(\bm\theta_{t+1})\\
        \Rightarrow
        \sum_{t=T}^{T+N}\eta_{t}(1-c)\mathbb{E}\left[||\bar{\bm g}_{t}||_2^2\right]\leq\mathbb{E}\left[\mathcal{L}(\bm\theta_{T})-\mathcal{L}(\bm\theta_{t+N+1})\right]<\infty\\
        \lim_{N\to\infty}\sum_{t=T}^{T+N}\eta_{t}(1-c)\mathbb{E}\left[||\bar{\bm g}_{t}||_2^2\right]<\infty.
    \end{align}
\end{subequations}
Together with the condition $\sum_{t}\eta_t=\infty$ and $r_{t+1}=\mathcal{O}(\eta_t)$, we can validate the convergence of the optimization process, i.e., $\lim_{t\to\infty}\mathbb{E}\left[||\bar{\bm g}_{t}||_2\right]=0$.
This further indicates the $\lim_{t\to\infty}\mathbb{E}\left[||\nabla_{\bm\theta}\mathcal{L}(\bm\theta_t)||_2\right]\leq\lim_{t\to\infty}\mathbb{E}\left[||\bar{\bm g}_{t}||_2\right]+\mathbb{E}\left[\Delta\bm g_{t}\right]\leq\lim_{t\to\infty}\mathbb{E}\left[||\bar{\bm g}_{t}||_2\right]+r_{t+1}G=0$.
And this completes the proof of \textbf{Theorem} 2.}

\paragraph{Diminishing Temporal Bias in the RPM.}
The above few sections have clarified the predictive foundation of MPTS and illustrated the theoretical convergence in stochastic optimization.
Intuitively, the diminishing rank flipping can be viewed as the noise noise injected to the gradient of the MC-CVaR.
In implementation, MPTS uses the function approximation capability of neural networks and introduces the RPM to utilize the optimization history $H_{1:t}$, generalizing from previously sampled and evaluated tasks to a new candidate pool $\mathcal{T}_{t+1}^{\hat{\mathcal{B}}}$.
That is the rank flipping can also be attributed to the error of the function approximation in the RPM.
However, this can be well alleviated if seeking an optimal RPM.
For better understanding, we analyze the temporal bias on arbitrary task $\tau$ through the adaptation risk approximation lens:
\begin{subequations}
    \begin{align}
    \text{Temporal Bias:}\quad
    \left|\mathbb{E}_{p(\ell\vert\bm\tau, H_{1:t})}[\ell] - \ell(\mathcal{D}_{\tau}^{Q},\mathcal{D}_{\tau}^{S}; \bm\theta_{t+1})\right|\\
    \leq \left|\mathbb{E}_{p(\ell\vert\bm\tau, H_{1:t})}[\ell] - \ell(\mathcal{D}_{\tau}^{Q},\mathcal{D}_{\tau}^{S}; \bm\theta_t)\right| + \left|\ell(\mathcal{D}_{\tau}^{Q},\mathcal{D}_{\tau}^{S}; \bm\theta_t) - \ell(\mathcal{D}_{\tau}^{Q},\mathcal{D}_{\tau}^{S}; \bm\theta_{t+1})\right| \\
    \leq \epsilon_t + \beta_1 \|\bm\theta_{t+1} - \bm\theta_t\|_{2} 
    = \epsilon_t^{\text{approx error}} + \beta_1 \eta_t \|\nabla_{\bm\theta}\mathcal{L}(\bm\theta_t)\|_{2}.
    \end{align}
\end{subequations}
Here $\epsilon_t^{\text{approx error}}$ denotes the RPM's approximation error due to the function approximation of the RPM, while $\beta_1$ is the Lipschitz from the Assumption 1.
Note that if the RPM is expressive enough, the generalization error from function approximation can be theoretically well decreased with the accumulation of the optimization history based on the statistical learning theory.
Hence, the limit of the right side term shrinks to extremely tiny values over iteration.

As specific analysis from multiple approximation bias depends on the design of the RPM.
This work provides the VAE-like model in implementation, and we believe there exist more optimal candidates for the choice of RPMs.
And this can be a promising research direction in the future.

To conclude, MPTS is computationally efficient and requires minimal annotation, such as agent-environment interactions; however, it also introduces approximation bias from the RPM and irreducible bias due to rank flipping when the model parameters change from $\bm\theta_t$ to $\bm\theta_{t+1}$. (Please see this point in Fig.~\ref{fig_rank_flip})

\section{Backbone Methods \& Experimental Details in Any-Shot Learning}\label{sec_supp_backbone}

\subsection{MAML, Reptile and PEARL}

In sinusoid regression and Meta-RL, MAML is used as the backbone algorithm.
As previously discussed, MAML is widely applied in solving few-shot learning tasks.
In mathematics, its optimization objective can be characterized as:
\begin{equation}
    \begin{split}
        \min_{\bm\theta\in\bm\Theta}\mathbb{E}_{p(\tau)}\left[\ell(\mathcal{D}_{\tau}^{Q};\bm\theta-\lambda\nabla_{\bm\theta}\ell(\mathcal{D}_{\tau}^{S};\bm\theta))\right],
    \end{split}
\end{equation}
where the term inside the bracket specifies the adaptation risk $\ell(\mathcal{D}_{\tau}^{Q},\mathcal{D}_{\tau}^{S};\bm\theta)$, and $\bm\theta-\lambda\nabla_{\bm\theta}\ell(\mathcal{D}_{\tau}^{S};\bm\theta)$ denotes the gradient update with the learning rate $\lambda$ as fast adaptation to the task $\bm\tau$.
After meta-training, we can access the meta initialization $\bm\theta$ that generalizes across the task space.

When it comes to reinforcement learning scenarios, $\mathcal{D}_{\tau}$ corresponds to episodic returns collected from MDPs with either the meta policy or the fast adapted policy.
To ensure enough coverage of task space, we adopt a mixture strategy of MPTS and random sampling as an empirical regularizer in all RL scenarios, which is similar to work \citep{greenberg2024train}.

Similar to MAML, Reptile \citep{nichol2018first} is a simple yet effective meta-learning algorithm to learn an initialization of model parameters for rapid adaptation to new tasks with just a few gradient updates. 
Unlike MAML, which explicitly differentiates through the adaptation process, Reptile approximates the meta-objective by repeatedly sampling tasks, performing a few steps of stochastic gradient descent on each, and then moving the initialization toward the adapted weights. 

PEARL \citep{rakelly2019efficient} is an off-policy SoTA Meta-RL algorithm that enables efficient adaptation to new tasks by learning a latent probabilistic context variable representing task-specific information. 
Instead of relying on on-policy trajectories, PEARL uses off-policy data and amortized inference to infer this latent variable $q_{\bm\phi}(\bm z \vert \mathcal{C})$, where $\mathcal{C}$ is a small context set of experience from the target task. 
The policy $\pi_{\bm\theta}(\bm a \vert\bm s, \bm z)$ and value functions are conditioned on $\bm z$, allowing fast adaptation through posterior inference rather than gradient updates. 
This design benefits from sample efficiency of off-policy learning and rapid task adaptation.

\subsection{Domain Randomization}

Robotic DR refers to the setup that trains the agent in a collection of environments to obtain a generalizable policy.
The diversity of environments tends to increase the robustness of policies in deployment.
Such a setup does not require few-shot episodes in unseen but similar environments.
In mathematics, we can express the optimization objective as:
\begin{equation}\label{eq_dr}
    \begin{split}
        \max_{\bm\theta\in\bm\Theta}\mathcal{J}(\bm\theta):=\mathbb{E}_{\pi_{\bm\theta}}\mathbb{E}_{p(\tau)}\left[\sum_{t=0}^{H}\gamma^{t}r_{t}\right]
    \end{split}
\end{equation}
where $p(\tau)$ defines the distribution over MDPs, and $\{r_{t}\}_{t=0}^{H}$ is the episodic stepwise reward after interacting with a specific MDP with $H$ as the horizon.
Once finishing the optimization of Eq. (\ref{eq_dr}), we can access the policy $\pi_{\bm\theta}$ as the zero-shot decision-maker in new environments.
In this case, the adaptation risk can be in the form $\ell(\mathcal{D}_{\tau}^{Q},\mathcal{D}_{\tau}^{S};\bm\theta)=-\sum_{t=0}^{H}\gamma^{t}r_{t}$.

Remember that MDP distribution $p(\tau)$ is mostly induced by physical parameters, e.g., mass, gravity, friction, etc., or the reward functions.
In each training iteration, the machine learner resamples a batch of MDPs and gets the shared policy to interact with them to collect episodes.
Consequently, the query dataset contains the episodes collected with no support dataset.
Overall, policy optimization follows the standard TD3 algorithm \citep{fujimoto2018addressing} due to its sample efficiency and stability.

\subsection{Multi-Modal Prompt Learning}

Multi-modal prompt learning is based on the backbone of the prompt tuning method MaPLe \citep{khattak2023maple}, which we use on both few-shot and SFT for image classifications.

\paragraph{Few-shot classification.}
The few-shot prompt learning refers to the common few-shot classification setting \citep{finn2017model, snell2017prototypical}.
We integrate the MaPLe backbone with the ProtoNet \citep{snell2017prototypical} to fully utilize the support sets in few-shot learning.
As illustrated in Section \ref{sec_supp_pseudo_class}, we generate the model prediction using both the textual classifiers from the CLIP textual encoder and the visual classifiers from the support set. 
By freezing the CLIP model parameters, only the prompts are optimized during meta-training.
In mathematics, the optimization objective can be formulated as:

\begin{equation}
\max_{\bm u} \mathbb{E}_{p(\tau)}\mathbb{E}_{\bm x \sim \mathcal{D}_{\tau}^{Q}}\left[\log p_{\bm \theta} (y|\bm x, \bm u)\right],
\end{equation}
where $\bm u$ denotes the learnable multi-modal prompts. $\bm \theta = (\bm \theta_t, \bm \theta_i)$ contains the parameters of the CLIP textual and visual encoders, which are frozen during training. 
The prediction of each query image $\bm x$ from task $\tau$ is calculated by Eq. (\ref{eq: fewshot_maple}).
Loglikelihood maximization is implemented by minimizing the classification cross-entropy loss.

\paragraph{SFT for image classification.}
The prompt learning setting refers to the 16-shot classification task proposed in work \citep{zhou2022learning, zhou2022conditional}. 
Based on the MaPLe backbone \citep{khattak2023maple}, we again freeze the CLIP model parameters and tune multi-modal prompts. 
The prompts are optimized on a selected ImageNet subset, with 16 samples from each category.

Note that in the SFT setting, we do not have pre-defined \textbf{\texttt{N-way K-shot}} tasks, either the splits of support and query sets in each task.
Therefore, we replace the ``tasks'' in the meta-learning setting with training samples. 
Model predictive task sampling is then achieved through data sampling.
In mathematics, the objective can be formulated as:
\begin{equation}
\max_{\bm u} \mathbb{E}_{\bm x \sim \mathcal{D}}\left[\log p_{\bm \theta} (y|\bm x, \bm u)\right],
\end{equation}
where $\bm u$ and $\bm \theta$ denote the prompts and frozen CLIP parameters as in the few-shot prompt learning setting, respectively. 
$\bm x$ are training samples from the entire training set $\mathcal{D}$.
The prediction of each image $p_{\bm \theta} (y|\bm x, \bm u)$ is calculated by $\frac{\exp(-d(f_{\bm\theta_i}(\bm x, \bm u),\bm t_{k}))}{\sum_{k^{\prime}}\exp(-d(f_{\bm\theta_i}(\bm x, \bm u),\bm t_{k^{\prime}}))}$ with the textual classifiers $\bm t$, obtained similarly to Eq. (\ref{eq: cliptext}).

\section{Experimental Setups \& Implementation Details}\label{sec_supp_exp}

\paragraph{Practical Learning Efficiency and Robustness.}
Widely recognized in reinforcement learning is the high sample complexity in policy evaluation, which demands massive interactions with environments, while policy optimization over the MDP distribution makes this even more severe.
In \texttt{N-way K-shot} image classification, we can create \texttt{K-shot} classification task from an arbitrary combination of $N$ classes; then the task space complexity $\mathcal{O}\Big(C_{M}^{N}\Big)$ grows with the number of categories $M$ in image datasets.
Meanwhile, challenges arise when gradient updates of foundation models consume substantial computational power and memory with a large batch size.
Similar circumstance also occurs in robust finetuning foundation models.

\begingroup
\setlength{\tabcolsep}{20.0pt}
\begin{table}[htbp]
   \begin{center}
    \caption{\textbf{A Summary of the Considered Benchmarks.}
    Here, we list the primary expensive part in task episodic learning for each scenario together with backbone methods.
    Also note that \texttt{N-way K-shot} image classification and SFT requires implicit task identifiers while others can directly access explicit task identifiers as the lower dimensional task representation.
    }
    \label{table_bench_summary}
    \begin{tabular}{|c|c|c|c|}
      \toprule %
      Benchmarks & Adaptation & Backbone & Expensive Part\\
      \toprule %
      \texttt{K-shot} sinusoid regression & few-shot & MAML/Reptile & computations\\
      \texttt{N-way K-shot} image classification & few-shot & MaPLe & computation/memory\\
      Meta-RL & few-shot & MAML/PEARL& interactions\\
      \toprule %
      Robotic DR & zero-shot & TD3 & interactions\\
      \midrule %
      SFT & many-shot & MaPLe & computation/memory\\
      \bottomrule %
    \end{tabular}
  \end{center}
\end{table}
\endgroup

\paragraph{Neural architecture of the RPM.}
As mentioned in the main text, the RPM is in an encoder-decoder structure.
For generality sake, we keep the neural architecture same for all benchmarks, including regression, classification and reinforcement learning.
The encoder includes an embedding network with 4 hidden layers of size 128 (for Image Classification and Prompt-Tuning) or 10 (for Sinusoid Regression, Meta-RL, and Robotic DR) with the Rectified Linear Unit (ReLU) nonlinear activation units to encode $\{[\bm\tau_{t,i},\ell_{t,i}]\}$ batch for mean pooling and then maps to $[\bm\mu,\bm\Sigma]$ with an output layer.
The decoder is a network with 3 hidden layers with nonlinear activation units to map $[\bm z,\bm\tau]$ to $\ell\in\mathbb{R}$.
For further details, please refer to our code.

\paragraph{Visualized Results during Training Phases.}
Note that the active selection and the random sampler with different batches affect the reflection of the machine learner's performance.
Hence, learning curves in sinusoid regression (Fig.~\ref{fig_meta_sin}a), Meta-RL (Fig.~\ref{fig_merged_metarl}a-b), and DR (Fig.~\ref{fig_merged_dr}a-b) are actually evaluated in a uniformly sampled validation task dataset for fair comparison.
Details on these validation task dataset are attached in the opensourced code.

\paragraph{Considerations in Experimental Design.}
To comprehensively evaluate MPTS, we consider benchmarks that span typicality and various levels of complexity and computational demands. 
For trait illustration, we adopt sinusoid regression, a classical and widely used benchmark in meta-learning research, as it provides an accessible and interpretable testbed for verifying the core traits of MPTS. 
To examine computational and memory efficiency in pattern recognition with foundation models, we employ few-shot learning and multimodal prompt-tuning tasks in image classification, which are standard benchmarks in the community. 
These tasks illuminate the advantage of MPTS in alleviating the intensive computational and memory costs typically associated with large backbone models under CVaR-based strategies. 
Finally, to investigate annotation and computational efficiency in sequential decision-making, we include Meta-RL and DR benchmarks. 
As standard environments for adaptive agent training, they are notoriously expensive due to extensive agent–environment interactions and the heavy task evaluations required by MC-CVaR strategies. 
In contrast, MPTS is capable of reducing such costs by partially surrogating agent–environment interactions and directly identifying a subset of MDPs from which to collect trajectories.

\subsection{Sinusoid Regression}

\paragraph{Task setup.}
For sinusoid regression, we retain the setup in MAML\citep{finn2017model}, where the few-shot machine learner tries to complete a wave function with the support dataset.
In specific, sampling the amplitude $a$ and the phase $b$ configures the wave function, and $10$ data points are uniformly sampled from the interval $[-5.0,5.0]$ coupled with $y=a\sin(x-b)$ to obtain the support dataset. 
This formulates the $10-\texttt{shot}$ sinusoid regression task.

\paragraph{Meta training process and neural architectures for MAML.}
The machine learner is a neural network with 2 hidden layers of size $40$ with two nonlinear activation units.
The task batch for ERM, GDRM, OHTM, DATS and TDPS is $16$, and that for DRM is $32$ as default.
DRM selects the $16$ hardest tasks for optimization based on their real evaluation losses.
OHTM selects the $8$ hardest tasks from the memory and randomly samples the remaining ones from the current batch, following \citet{kumar2023effect}.
The temperature parameter is set to $\eta=0.001$ in GDRM, $\eta=0.02$ in DATS, and $\eta=0.005$ in PDTS.
The learning rates for the inner loop and the outer loop are $0.001$.
For MPTS, the batch size of the candidate task identifier in training is $32$, the Lagrange multiplier is set as $1$, and the balance parameters $\{\gamma_0, \gamma_1\}$ is set to $\{1, 3\}$.
We use the Adam optimizer with the learning rate $3e-4$ to update the RPM for 20000 step.
In sinusoid regression and Meta-RL, we use the standard repository in MAML \citep{finn2017model}.

\paragraph{Meta training process and neural architectures for Reptile.}
Reptile is implemented following \citet{nichol2018first}, with $8$ inner iterations, an inner learning rate of $0.01$, and an outer learning rate of $0.5$.
The neural network architecture for the machine learner and the hyperparameters for all baselines are kept consistent with MAML, except that the temperature parameter in DATS is set to $\eta=0.2$.

\subsection{\texttt{N-way K-shot} Image Classification}

\paragraph{Task Setup.}
This is a commonly seen benchmark in few-shot learning.
It learns a model that can classify images from $N$ distinct classes with support of $K$ labeled examples for each class.
The support dataset as reference is in the form $\mathcal{D}_{\tau}^{S}=\{\{[\bm x_{i,k},y_{i,k}=i]\}_{k=1}^{K}\}_{i=1}^{N}$.
And the query dataset corresponds to the image information for the model to classify.
Hence, for a large image dataset with $M$ classes, the complexity of the task space is $\mathcal{O}\Big(C_{M}^{N}\Big)$. 
Here, we include ImageNet-CG \citep{hendrycks2019benchmarking}, ImageNet-CI \citep{hendrycks2019benchmarking}, ImageNet-CS \citep{hendrycks2019benchmarking}, ImageNet-A \citep{hendrycks2021natural}, ImageNet-S \citep{wang2019learning} and ImageNet-R \citep{hendrycks2021many} as the dataset in evaluation.

\paragraph{Meta training process and neural architectures.}
Explicit $\bm\tau$ are unavailable to specify the task; however, it can be approximately resolved by describing the identifier through a small reference model. 
Specifically, we leverage CLIP's text encoder to obtain $\bm\tau\approx[\text{CLIP}_{\text{text}}(\mathcal{C}_1),\dots,\text{CLIP}_{\text{text}}(\mathcal{C}_K)]$ with the tokenizations of $K$ class texts $\mathcal{C}_{1:K}$ from $\mathcal{D}_{\tau}^{S}$.
The machine learner utilizes a prompt learning backbone following MaPLe, with the frozen CLIP model.
The task batch for ERM and GDRM is 4, and that for DRM is 8 as default. 
The temperature parameter in GDRM is 0.001.
The learning rate for the outer loop is 0.01.
The learning rate for the inner loop follows that in MaPLe \citep{khattak2023maple}.
The following is about extra optimization details or setups in MPTS.
The task identifier is generated by the frozen CLIP text encoder using the input class names, with a dimensionality of 512.
The batch size of the identifier in training is 8, the Lagrange multiplier is set as $\beta$, and we use the Adam optimizer with the learning rate 0.01 to update the RPM.

\subsection{Prompt-based Few-shot Image Classification}\label{sec_supp_pseudo_class}

We adopt the standard few-shot image classification setting \citep{finn2017model, snell2017prototypical}, where tasks are constructed using the \textbf{\texttt{N-way K-shot}} paradigm for both meta-training and meta-testing. 
Each task comprises support and query sets. 
The support set contains \textbf{\texttt{K}} examples for each of the \textbf{\texttt{N}} classes, while the query set includes \textbf{\texttt{15}} examples per class.
During meta-training, labels for both support and query data are accessible to the adaptive machine learner. 
During meta-testing, the query dataset's labels are to be predicted given the labeled support dataset.
The class categories of task datasets in the meta-training and meta-testing do not overlap.
In experiments, we specifically consider a \textbf{\texttt{5-way 1-shot}} image classification configuration.
During meta-training, we set the task batch for ERM and DRM as $\mathcal{B}=4$ (For implementation simplicity, the data loader samples $8$ tasks and then randomly keeps half without ranking to optimize).
The task batch for DRM is $\mathcal{B}=8$ before the filtering operation; DRM filters half to optimization.
Similarly, that for MPTS is $\mathcal{B}=8$ and only $4$ tasks are screened to optimize.
For DATS and TDPS, we set the task batch as $\mathcal{B}=4$ and compute the gradients of each task. The optimization of the 4 tasks are then weighted by gradient-derived diffculty scores (TDPS) or similarity scores (DATS).
In OHTM, we introduce a hard task buffer as size ten to store the most difficult tasks during meta-training. We set $\mathcal{B}=8$ and randomly sample 4 tasks from the buffer in each iteration.

To enable few-shot learning by prompt-tuning, we integrate the multimodal prompt learning methods MaPLe \citep{khattak2023maple} and prototypical network (ProtoNet) \citep{snell2017prototypical}.
MaPLe operates on the CLIP model \citep{radford2021learning}, capturing multimodal prompts to refine both visual and textual feature representations with frozen CLIP's parameters. 
These refined textual features serve as classifiers for predicting refined image features.
In parallel, ProtoNet is utilized to derive class-specific visual embeddings from the support set, which assist in distinguishing query samples.

To utilize both MaPLe and ProtoNet, we construct classifiers based on both textual features and visual embeddings. 
Predictions for query samples are generated by combining the classifiers through a weighted sum. 
This combination strategy is employed during meta-training to optimize the multimodal prompts. 
Meanwhile, the CLIP model's parameters remain frozen throughout optimization.
During meta-testing, these trained prompts are adopted to create textual and visual classifiers and process query image features. 
Final predictions for each query sample are made using the same weighted combination approach as in meta-training.

In mathematics, we can characterize the mentioned pipeline as:
\begin{equation}
\label{eq: cliptext}
    \text{Textual Classifier from the Textual Features:} \quad
    \bm t_{k}=f_{\bm\theta_t}(\bm l_k, \bm u),
\end{equation}
\begin{equation}
    \text{Prototypical Classifier from the Support Image:} \quad
    \bm c_{k}=\frac{1}{|S_k|}\sum_{(\bm x_i,y_i)\in S_k}f_{\bm\theta_i}(\bm x_i, \bm u),
\end{equation}
\begin{equation}
\label{eq: fewshot_maple}
\begin{aligned}
    \text{Classification Likelihood from the Query Dataset:} & \\ 
    \quad
    p_{\bm\theta}(y=k\vert \bm x, \bm u)
        = \lambda_1 \frac{\exp(-d(f_{\bm\theta_i}(\bm x, \bm u),\bm c_{k}))}{\sum_{k^{\prime}}\exp(-d(f_{\bm\theta_i}(\bm x, \bm u),\bm c_{k^{\prime}}))} 
        & + \lambda_2 \frac{\exp(-d(f_{\bm\theta_i}(\bm x, \bm u),\bm t_{k}))}{\sum_{k^{\prime}}\exp(-d(f_{\bm\theta_i}(\bm x, \bm u),\bm t_{k^{\prime}}))},
\end{aligned}
\end{equation}
where $\bm \theta_t$ and $\bm \theta_i$ denote the textual and visual encoders of the CLIP model, respectively.
$\bm l_k$ and $\bm u$ respectively denote the textual descriptions of category \textbf{\texttt{k}} and the multimodal prompts.
$(\bm x_i,y_i)\in S_k$ denotes the images and labels of category \textbf{\texttt{k}} in the support dataset $\mathcal{D}_{\tau}^{S}$. 
Once the textual classifier $\bm t$ and support visual classifier $\bm c$ are obtained, we predict the query sample $\bm x$ by the classifiers with hyperparameters $\lambda_1=0.25$ and $\lambda_2=1.0$.

\subsection{Meta-RL}

\paragraph{Task Setup.}
We construct MDP distributions based on Mujoco physics engines \citep{todorov2012mujoco}.
These include HalfCheetahVel, HalfCheetahMassVel, Walker2dVel, Walker2dMassVel, and ReacherPos.
The HalfCheetahVel and Walker2dVel tasks involve training the cheetah or walker robot to achieve a target velocity. 
These tasks define the reward function as the negative absolute difference between the robot's current velocity and the target velocity, supplemented by a control penalty and an alive bonus to facilitate the learning process.
The goals and rewards of HalfCheetahMassVel and Walker2dMassVel are the same as those of the corresponding velocity-related tasks, with the additional identifier of varying mass for the cheetah or walker robot.
The ReacherPos task tries to move a two-jointed robot arm's end effector close to a target position, and its reward function is defined as the negative $L$-1 distance between the robot arm's position and the target position, supplemented by a control cost to ensure robustness.

\paragraph{Meta training process and neural architectures for MAML.}
The machine learner is a neural network with 2 hidden layers of size 64 with the Rectified Linear Unit (ReLU) nonlinear activation units.
The task batch for ERM, GDRM, OHTM, DATS and TDPS is 20, and that for DRM is 40 as default. 
DRM selects the $20$ hardest tasks for optimization based on their real evaluation losses.
OHTM selects the $10$ hardest tasks from the memory and randomly samples the remaining ones from the current batch, following \citet{kumar2023effect}.
The temperature parameter in GDRM is $0.001$.
For DATS, it is set to $0.05$ for HalfCheetahMassVel, $0.1$ for ReacherPos, HalfCheetahVel, and Walker2dVel, and $0.2$ for Walker2dMassVel.
For TDPS, it is set to $0.05$ for HalfCheetahMassVel, Walker2dMassVel, and HalfCheetahVel, $0.1$ for Walker2dVel, and $0.2$ for ReacherPos.
The learning rates for the inner loop and the outer loop are 0.1.
For MPTS, the batch size of the candidate task identifier in training is 30, the Lagrange multiplier is set as $\beta=0.0001$, and the balance parameters $\{\gamma_0, \gamma_1\}$ is set to $\{1, 5\}$. We use the Adam optimizer with the learning rate 0.005 to update the RPM.

\paragraph{Meta training process and neural architectures for PEARL.}
PEARL \citep{rakelly2019efficient} and RoML \citep{greenberg2024train} are implemented based on the official RoML codebase available at \url{https://github.com/ido90/RoML-pearl}
.
Two tasks, HalfCheetahBody and HalfCheetahMass, are adopted from RoML.
For MPTS, the batch size of the candidate task identifier during training is set to $5\times$ that of the real training batch size ($30$) for HalfCheetahBody, and $2\times$ for HalfCheetahMass.
All other hyperparameters follow those used in MAML.

\subsection{Robotic DR}

\paragraph{Task Setup.}
We conduct experiments on LunarLander-v2 and ErgoReacher-v0 environments \citep{mehta2020active}. 
LunarLander is a 2 degrees of freedom (DoF) environment in which the agent has to land a spacecraft on a designated landing pad without crashing, implemented using Box2D \citep{box2d}. 
The reward function of LunarLander awards positive rewards for successful landings, negative rewards for crashes, and additional penalties for fuel consumption and deviation from the landing pad, encouraging efficient and controlled landings. 
ErgoReacher is a 4 DoF arm environment from \citep{golemo2018sim} in which the arm has to touch a goal with its end effector, implemented in the Bullet Physics Engine \citep{coumans2015bullet}. 
The reward function of ErgoReacher includes the negative distance between the end effector's position and the target, along with other control costs to promote efficient and safe movements. 
In LunarLander, we randomize the engine strength, while in ErgoReacher, we randomize the joint damping and maximum torque for each of the 4 joints, resulting in a total of 8 parameters. 
The detailed ranges of the randomized parameters for each environment are provided in Table~\ref{table_bench_explicit_identifier}.

\paragraph{DR training process and neural architectures.}
Both the actor and critic are neural networks with $2$ hidden layers of size $256$ with the Rectified Linear Unit (ReLU) nonlinear activation units.
The task batch for ERM, GDRM, and OHTM is $10$, and that for DRM is $20$ as default. 
DRM selects the $10$ hardest tasks for optimization based on their real evaluation losses.
OHTM selects the $5$ hardest tasks from the memory and randomly samples the remaining ones from the current batch, following \citet{kumar2023effect}.
The temperature parameter in GDRM is $0.01$.
The learning rates for actor and critic are $3e-4$.
The following is about extra optimization details or setups in MPTS.
The batch size of the candidate task identifier in training is $25$ for LunarLander and $250$ for ErgoReacher. The Lagrange multiplier is set as $\beta=1.0$, and the balance parameters $\{\gamma_0, \gamma_1\}$ is set to $\{1, 5\}$. We use the Adam optimizer with the learning rate $0.005$ to update the RPM.

\subsection{Prompt-Tuning Multimodal Foundation Models}

\paragraph{Task Setup of Prompt-tuning.}
We refer the reader to MaPLe's implementation in \url{https://github.com/muzairkhattak/multimodal-prompt-learning}.
For all baselines, we retain the MaPLe's task construction in prompt-tuning.

\paragraph{Prompt-tuning process and neural architectures.}

The machine learner follows the prompt learning method MaPLe \cite{khattak2023maple} based on the frozen CLIP model (ViT/B-16).
The task batch for ERM and GDRM is 4, and that for DRM is 8 as default. 
The temperature parameter in GDRM is 0.001.
The learning rate for the outer loop is 0.005.
The learning rate for the inner loop follows that in MaPLe \cite{khattak2023maple}.
The following is about extra optimization details or setups in MPTS.
As for the neural architecture of the RPM, the encoder is a neural network with 5 hidden layers with 4 ReLU nonlinear activation units, and the decoder is a neural network with 4 hidden layers with 3 nonlinear activation units. 
The task identifier's dimension is 512 with the latent embedding from CLIP encoders.
The batch size of the identifier in training is 8, the Lagrange multiplier is set as $\beta$, and we use the Adam optimizer with the learning rate 0.005 to update the RPM for 8000 steps.
During prompt-tuning, we set the task batch for ERM and DRM as $\mathcal{B}=4$ (For implementation simplicity, the data loader samples $8$ tasks and then randomly keeps half without ranking to optimize).
The task batch for DRM is $\mathcal{B}=8$ before the filtering operation; DRM filters half to optimization.
Similarly, that for MPTS is $\mathcal{B}=8$ and only $4$ tasks are screened to optimize.

\section{Computational Tools \& Platforms \& Model Design}

In this research project, we use the Pytorch as the package to implement all methods to run all deep learning experiments.

From Definition~\ref{definition_mpts}, we can see that the goal is to learn some low-dimensional signals online when the task identifier is low-dimensional and the loss is one-dimensional.
According to the Occam's Razor theorem, the best practice is to adopt the generative model that captures historical priors while exhibiting less model complexity.
In our implementation, we typically use the same architecture of shallow networks from the Open-source code.
However, we also believe it is a promising direction to develop more advanced RPMs to online tell task difficulty and achieve the purpose of Definition~\ref{definition_mpts}.

\end{document}